\documentclass[10pt,twocolumn,letterpaper]{article}

\usepackage{iccv}
\usepackage{times}
\usepackage{epsfig}
\usepackage{graphicx}
\usepackage{amsmath}
\usepackage{amssymb}
\usepackage{multirow}
\usepackage[table,xcdraw]{xcolor}

\usepackage[toc,page]{appendix}

\usepackage{algorithm}
\usepackage{algorithmic}
\usepackage{caption}
\usepackage{subcaption}
\usepackage[flushleft]{threeparttable}
\usepackage{enumitem}

\usepackage{graphicx}
\usepackage{booktabs}
\usepackage{multirow}

\usepackage{array}
\newcommand{\PreserveBackslash}[1]{\let\temp=\\#1\let\\=\temp}
\newcolumntype{C}[1]{>{\PreserveBackslash\centering}p{#1}}
\newcolumntype{R}[1]{>{\PreserveBackslash\raggedleft}p{#1}}
\newcolumntype{L}[1]{>{\PreserveBackslash\raggedright}p{#1}}

\DeclareMathOperator*{\argmin}{\arg\min}

\usepackage{soul}
\usepackage{color}

\usepackage{caption}
\usepackage{subcaption}
\usepackage[para]{footmisc}

\usepackage[pagebackref=true,breaklinks=true,letterpaper=true,colorlinks,bookmarks=false]{hyperref}

\iccvfinalcopy % *** Uncomment this line for the final submission

\def\iccvPaperID{10422} % *** Enter the ICCV Paper ID here

\begin{document}

%%%%%%%%% TITLE
\title{ACTIVE: Towards Highly Transferable 3D Physical Camouflage \\ for Universal and Robust Vehicle Evasion}

\author{Naufal Suryanto$^{1, \ddag}$, Yongsu Kim$^{1,2, \ddag}$, Harashta Tatimma Larasati$^{1}$, Hyoeun Kang$^{1,2}$, Thi-Thu-Huong Le$^{1}$\\
Yoonyoung Hong$^{1}$, Hunmin Yang$^{3}$, Se-Yoon Oh$^{3}$, Howon Kim$^{1,2, \textasteriskcentered}$\\
{\small $^{1}$Pusan National University, South Korea; $^{2}$SmartM2M, South Korea; $^{3}$Agency for Defense Development (ADD), South Korea} \\ 
% {\small $^{3}$Agency for Defense Development (ADD), South Korea;}\\ 
% {\small $^{4}$Institut Teknologi Bandung, Indonesia.}\\
{\small \url{https://islab-ai.github.io/active-iccv2023/}}
}
\maketitle
\ificcvfinal\thispagestyle{empty}\fi

%%%%%%%%% ABSTRACT
\begin{abstract}
Adversarial camouflage has garnered attention for its ability to attack object detectors from any viewpoint by covering the entire object's surface. However, universality and robustness in existing methods often fall short as the transferability aspect is often overlooked, thus restricting their application only to a specific target %(i.e., vehicle, model, environment) 
with limited performance. To address these challenges, we present Adversarial Camouflage for Transferable and Intensive Vehicle Evasion (ACTIVE), a state-of-the-art physical camouflage attack framework designed to generate universal and robust adversarial camouflage capable of concealing any 3D vehicle from detectors. Our framework incorporates innovative techniques to enhance universality and robustness, including a refined texture rendering that enables common texture application to different vehicles without being constrained to a specific texture map, a novel stealth loss that renders the vehicle undetectable, and a smooth and camouflage loss to enhance the naturalness of the adversarial camouflage. Our extensive experiments on 15 different models show that ACTIVE consistently outperforms existing works on various public detectors, including the latest YOLOv7. 
Notably, our universality evaluations reveal promising transferability to other vehicle classes, tasks (segmentation models), and the real world, not just other vehicles.
% Notably, our universality evaluations reveal promising transferability beyond vehicles, but also to other vehicle classes, tasks (i.e., segmentation models), and the real world.
% Adversarial camouflage methods have garnered increasing attention due to their ability to attack object detectors from any viewpoint by covering the entire object's surface. In this paper, we present Adversarial Camouflage for Transferable and Intensive Vehicle Evasion (ACTIVE), a state-of-the-art physical camouflage attack framework designed to generate universal and robust adversarial camouflage capable of concealing any 3D vehicle from detectors. Our framework incorporates several innovative techniques to enhance universality and robustness. First, we employ a sophisticated texture rendering method that enables the application of common textures to different vehicles, without being constrained to a specific texture map. Subsequently, we introduce a novel stealth loss function that renders the vehicle undetectable, and utilize a smooth and a camouflage loss that enhance the naturalness of the adversarial camouflage. Our extensive experiments on 15 different models show that ACTIVE consistently outperforms other adversarial camouflage methods on various publicly available detectors, including the latest YOLOv7. Notably, our universality evaluations reveal promising transferability to other vehicle classes, tasks (i.e., segmentation models), and to the real world, not just to other vehicles.
\footnotetext[3]{Equal contribution} \footnotetext[1]{Corresponding author} \end{abstract}
\vspace{-0.2cm}%emphasize comprehensive/total number of exp
% \footnote{Demos are available at \url{https://activeattack.github.io/ACTIVE/}}
% \footnotetext[3]{Equal contribution} \footnotetext[1]{Corresponding author}

%%%%%%%%% BODY TEXT
\section{Introduction}

\begin{figure}
\begin{minipage}[t]{.328\columnwidth}
\begin{subfigure}{\columnwidth}
\includegraphics[width=\columnwidth]{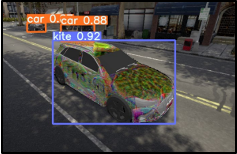}
\subcaption{FCA}
\end{subfigure}
\end{minipage}
\begin{minipage}[t]{.325\columnwidth}
\begin{subfigure}{\columnwidth}
\includegraphics[width=\columnwidth]{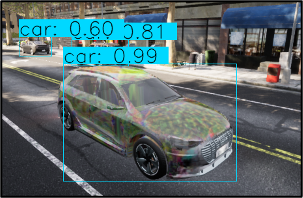}
\subcaption{FCA on UE4}
\end{subfigure}
\end{minipage}
\begin{minipage}[t]{.327\columnwidth}
\begin{subfigure}{\columnwidth}
\includegraphics[width=\columnwidth]{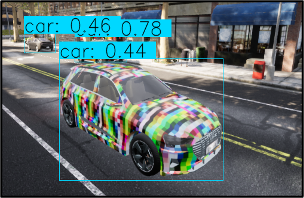}
\subcaption{DTA}
\end{subfigure}
\end{minipage}
\hfill
\begin{minipage}[t]{.475\columnwidth} % .48\columnwidth
\begin{subfigure}{\columnwidth}
\includegraphics[width=\columnwidth]{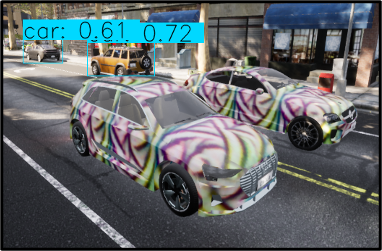}
\subcaption{ACTIVE (Ours)}
\end{subfigure}
\end{minipage}
\hfill
\begin{minipage}[t]{.496\columnwidth} %.495\columnwidth
\begin{subfigure}{\columnwidth}
\includegraphics[width=\columnwidth]{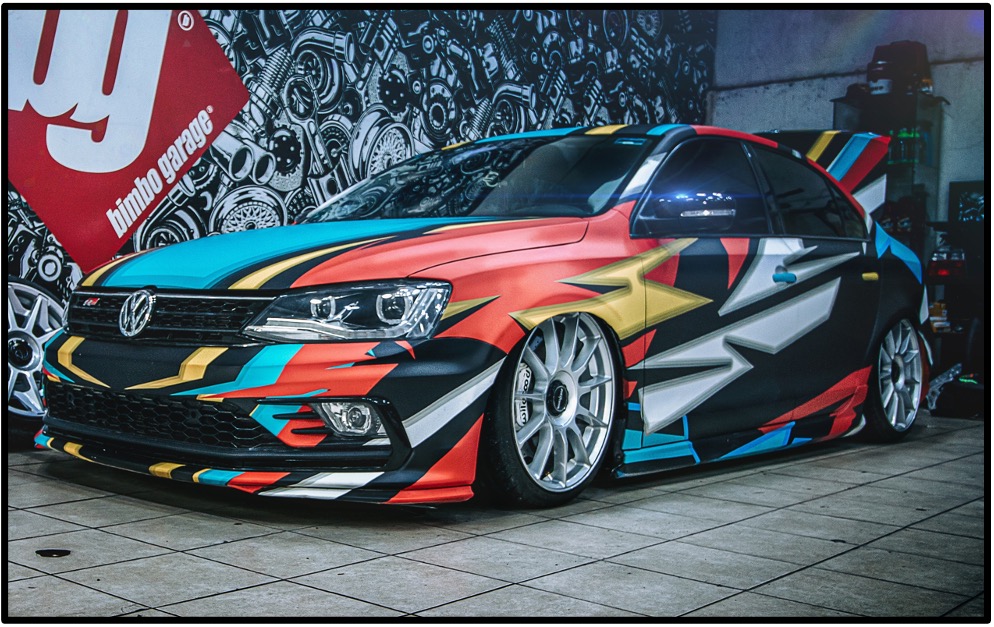}
% \hfill
\subcaption{Car Graffiti}
\end{subfigure}
\end{minipage}
\caption{Current state-of-the-art adversarial camouflage, Full-coverage Camouflage Attack (FCA) \cite{fcaattack}; (a) is undetected as car in their original result, but (b) is normally detected when fully transferred to Unreal Engine 4 (UE4). (c) The result of the Differentiable Transformation Attack (DTA) \cite{Suryanto_2022_CVPR}, with stronger result than FCA, but limited to unnatural mosaic pattern. (d) The result of ACTIVE, which has better attack performance (i.e., not detected as a car at all), is universal (i.e., implementable to other car), and has a natural pattern comparable to (e) a real-world graffiti.}

\label{fig:first_image}
\end{figure}

Deep neural networks (DNNs) have achieved tremendous outcomes in a wide range of research fields, especially in computer vision, such as facial recognition and self-driving cars \cite{junyi2021, wang2022yolov7, detr}. Despite their remarkable performance, DNNs, including object detection models, are vulnerable to adversarial attacks \cite{akhtar2018}. Generally, adversarial attacks can be classified into digital attacks and physical attacks \cite{wang2022}. Digital attacks are primarily carried out by adding small perturbations to pixels of input images. However, digital attacks have limitations in real-world scenarios because they have to manipulate the digital systems that may be configured with security schemes. To alleviate these limitations, physical attacks have been proposed that modify the object in the physical space rather than in the digital. 

Nonetheless, physical attacks are more challenging due to the inherently complex physical constraints (camera pose, lighting, occlusion, etc.). Therefore, most physical attack methods exclude the perturbation constraint (i.e., the result can be suspicious). There are mainly two types of methods in physical attacks: adversarial patch and adversarial camouflage. The former \cite{Brown_2017_CoRR, Fooling} is the method of physical attack by attaching a small, localized patch to an object. It only covers the planar part of the object’s surface and can fail to attack the detector depending on the viewing angles. 

Adversarial camouflage methods \cite{Zhang_2019_ICLR, Wu_CoRR_2020, Wang_2021_CVPR, fcaattack, Suryanto_2022_CVPR} have been proposed to overcome the limitations of the adversarial patch. This approach covers the whole surface of the object by manipulating the texture of the object, which leads to better attack performance regardless of the viewing angles. Most of these methods use \textit{vehicles} as the target objects due to their crucial role in real-world applications, such as surveillance systems and autonomous driving.

Due to the non-planarity of 3D vehicles, it is more challenging to generate optimal adversarial camouflage than in its 2D counterpart. 
Since the general 3D rendering is non-differentiable, early research \cite{Zhang_2019_ICLR, Wu_CoRR_2020} employed a black-box approach to generating adversarial camouflage, inevitably yielding a lower attack performance than the white-box one. More recent research (e.g., DAS \cite{Wang_2021_CVPR}, FCA \cite{fcaattack}, and DTA \cite{Suryanto_2022_CVPR}) utilize a neural renderer to acquire the advantage of the white-box approach, which offers differentiability. In particular, DTA proposes its own neural renderer capable of expressing various physical and realistic characteristics. Additionally, it employs its so-called Repeated Texture Projection function to apply the same attack texture to other vehicle types, thus improving universality. However, DTA relies on a simple texture projection, which may result in an inaccurate texture for non-planar shapes. 
For improving the accuracy and robustness of adversarial camouflage methods, it would be essential to bring forth a more sophisticated texture mapping approach. 
% \hl{This limitation suggests that a more sophisticated texture mapping approach may be necessary further to improve the accuracy and robustness of adversarial camouflage methods.} %maybe need a stronger sentence. Ex: This can cause...., 
%which calls for a more sophisticated texture mapping....

%from rebuttal
% \hl{Our main idea is to achieve a highly transferable 3D adversarial camouflage by enhancing \textit{universality} and \textit{robustness} that lacks in existing methods, addressed primarily by our proposed (1) \textit{texture rendering method} and novel (2) \textit{stealth loss function}. The term "sophisticated" was to underscore our TPM ability to generate \textit{object-independent} textures through \textit{accurate} surface-based projection, combined with stealth loss that minimizes detection score \textit{agnostically}, yields highly-enhanced universality and robustness.}

%--> we can use this only? 
%... that generates \textit{object-independent} textures through \textit{accurate} surface-based projection, combined with stealth loss that minimizes detection score \textit{agnostically} to yield highly-enhanced universality and robustness. --> but then we need to mention naturalness (again)? hmm

In this paper, we propose \textit{Adversarial Camouflage for Transferable and Intensive Vehicle Evasion} (ACTIVE), a state-of-the-art adversarial camouflage framework that greatly enhances robustness, universality, and naturalness compared to previous methods, as shown in Fig. \ref{fig:first_image}. 
%\hl{want to add more info later if we have space}, ex: ACTIVE employs....., which [add advantage]
% In summary, our contributions can be summarized as follows:
Our contributions can be summarized as follows:

\begin{itemize}[itemsep = 0.15em] %[noitemsep]
\item We utilize \textit{Triplanar Mapping}, a sophisticated texture mapping approach available through the neural renderer, to generate adversarial textures with improved robustness and universality. To the best of our knowledge, the use of this method in generating adversarial camouflage has never been found in literature. %,+ "or ... in general."
\item We employ \textit{Stealth Loss}, 
our novel attack loss function that minimizes the detection score from all valid classes, resulting in the target vehicle being not only misclassified but also undetectable.
\item We improve the naturalness of the adversarial camouflage by utilizing larger texture resolutions than previous works \cite{Zhang_2019_ICLR, Wu_CoRR_2020, Suryanto_2022_CVPR} and applying a smooth loss. Furthermore, we introduce a \textit{Camouflage Loss} that can enhance the camouflage of the vehicle against the background.
\item Our extensive experiments demonstrate that ACTIVE consistently outperforms previous works, exhibiting improved universality from multiple perspectives: instance-agnostic (available on various vehicle types), class-agnostic (even across different classes such as truck and bus), model-agnostic (performing on various vehicle detectors), task-agnostic (performing on segmentation models), and domain-agnostic (performing in real-world scenarios).
\end{itemize}

%------------------------------------------------------------------------
\section{Related Works}
\label{sec:related_works}

The Expectation over Transformation (EoT) \cite{EoT} has emerged as a leading approach in generating robust adversarial examples under various transformations, including variations in viewing distance, angle, and lighting conditions. Consequently, many adversarial camouflage methods \cite{Zhang_2019_ICLR,Wu_CoRR_2020, Wang_2021_CVPR,fcaattack,Suryanto_2022_CVPR} incorporate EoT-based algorithms to enhance their attack robustness in the physical scenarios.

Regarding the texture rendering process, differentiability is crucial to enable white-box attacks to obtain optimal adversarial camouflage, whereas non-differentiability of general texture rendering led to the initial proposal of the black-box approach. Zhang et al. \cite{Zhang_2019_ICLR} proposed CAMOU, utilizing the clone network that imitates the texture rendering and detection process, while Wu et al. \cite{Wu_CoRR_2020} suggested finding optimal adversarial texture based on genetic algorithm.

Huang et al. \cite{Huang_2020_CVPR} proposed the Universal Physical Camouflage Attack (UPC) as an alternative method for crafting universal adversarial camouflage, which differs from existing black-box approaches. To make UPC effective for non-rigid or non-planar objects, they introduced a set of transformations that can mimic deformable properties. However, subsequent studies \cite{Wang_2021_CVPR, fcaattack, Suryanto_2022_CVPR} have found that the transferability of UPC is limited when it comes to various viewing angles, other models, and different environments due to its inherent limitations of the patch-based method.

More recent research used a neural renderer, which provides a differentiable texture rendering, to improve attack performance. Wang et al. \cite{Wang_2021_CVPR} proposed the Dual Attention Suppression (DAS) attack, which suppresses both model and human attention. Meanwhile, Wang et al. \cite{fcaattack} proposed the Full-coverage Camouflage Attack (FCA), which is more robust under complex views.

However, Suryanto et al. \cite{Suryanto_2022_CVPR} pointed out that the DAS and FCA used a legacy renderer, which could not reflect various real-world characteristics and complex scenes, such as shadows and light reflections. They proposed the Differentiable Transformation Attack (DTA) that uses their own neural renderer, which provides rendering similar to a photo-realistic renderer. Despite its success, DTA also comes with several limitations. Notably, the method relies on a simple texture projection, which can lead to inaccurate texture mapping. %if possible, give a little bit info/example.
Furthermore, the generated camouflage features a colorful mosaic pattern, which appears unnatural to human observers. Tab. \ref{tab:compare_physical_attack} compares how well existing approaches stack up against the suggested method under various criteria. As shown, our proposal satisfies and achieves all good values under each condition setting compared to prior methods.

\begin{table}[h]
\centering
\caption{Comparison of proposed and existing physical camouflage attack methods.}
\label{tab:compare_physical_attack}
\vspace{-0.2cm}
\scalebox{0.95}{
\begin{tabular}{l|c|c|c|c|c|c|c|c}
\hline
Attack & 3D & WB & FC & (1) & (2) & (3)  & (4) & (5) \\ 
 &  &  &  & U & A & N & D & P \\ 
\hline 
AP \cite{Brown_2017_CoRR} & $\times$ & \checkmark  & $\times$ & $\star\star$ & $\star\star$ & - & $\star\star$ & - \\ \hline
UPC \cite{Huang_2020_CVPR}& $\times $& \checkmark & $\times$ & $\star\star$ & $\star\star$ & $\star\star$ & $\star\star$& $\star$ \\ \hline 
CM \cite{Zhang_2019_ICLR} & \checkmark & $\times$ & \checkmark & $\star$ & $\star\star$ & - & - & $\star\star$ \\ \hline 
ER \cite{Wu_CoRR_2020} & \checkmark & $\times$ & \checkmark & $\star$ & $\star\star$ & - & - & $\star\star$ \\ \hline 
DAS \cite{Wang_2021_CVPR} & \checkmark & \checkmark & $\times $& - & $\star$ & $\star\star$ & - & $\star$ \\ \hline 
FCA \cite{fcaattack} & \checkmark & \checkmark & \checkmark & - & $\star$ & $\star$ & - & $\star$ \\ \hline
DTA \cite{Suryanto_2022_CVPR} & \checkmark & \checkmark & \checkmark & $\star$& $\star\star$ & - & $\star$ & $\star\star$ \\ \hline 
Ours & \checkmark & \checkmark & \checkmark & $\star\star$ & $\star\star$ & $\star\star$ & $\star\star$ & $\star\star$ \\ \hline
\end{tabular}
% \vspace{-0.2cm}
}
    \begin{tablenotes}
      \scriptsize
      \item Notes: 
      \item 3D $|$ White Box (WB) $|$ Full Covering (FC)
      \item (1) Universality (U): $\star$ Might be universal, but only optimized on a single instance  $|$
      \item $\star$$\star$ Universal, the adversarial is optimized on multiple instances on the same category
      \item (2) Applicability (A): $\star$ Require exact position to place the adversarial $|$ \item $\star$$\star$ Can be placed anywhere which satisfies on the target object
      \item (3) Naturalness (N): $\star$ Consider naturalness such as using smooth texture $|$ \item $\star$$\star$ Have a more constrained setting
      \item (4) Digital Transformation (D): $\star$ Affine transformation only $|$ 
      \item $\star$$\star$ Both affine transformation and brightness, contrast 
      \item (5) Physical Transformation (P): $\star$ Camera position only $|$
      \item $ \star$$\star$ Both camera position and physical phenomena 
    \end{tablenotes}
\end{table}
% \vspace{-0.3mm}

%------------------------------------------------------------------------
\section{Methodology}
\label{sec:proposed_method}
% In this section, we first define the problem and then introduce our proposed framework. %not really necessary, can be deleted if required

\subsection{Problem Definition}
\label{subsec:problem_definition}
Assume $f$ is a neural renderer that generates optimal adversarial camouflage through white-box attacks based on differentiable rendering. $f$ learns texture rendering by solving Eq. \ref{eq:EOTnet_eq_fix},
\begin{equation}
f(x_{ref}, \eta) = x_{ren}
\label{eq:EOTnet_eq_fix}
\end{equation}
where $x_{ref}$ refers to the reference image, which includes the target vehicle, $\eta$ is the texture variable, and $x_{ren}$ is the rendered image with $\eta$. Previously, existing works have utilized neural renderer to generate adversarial camouflage by minimizing the loss function as shown in Eq. \ref{eq:loss_e_adv},
\begin{equation}
\argmin_{\eta_{adv}}L(h(f(x_{ref}, \eta_{adv})), y)
\label{eq:loss_e_adv}
\end{equation}
where $h$ is the hypothesis function for the vehicle detector, $y$ is the corresponding detection label output, and $L(h(x), y)$ %%this correct? not really same as eq
is the loss function that represents the confidence score of $h$ regarding class $y$. Thus, solving Eq. \ref{eq:loss_e_adv} involves generating $\eta_{adv}$ to attack $h$ by minimizing its confidence score. 

We have noticed that there are rooms for improvement in at least two aspects. 
First, most of the neural renderers employed in existing adversarial camouflage methods use object-dependent texture mapping, such as UV mapping. Therefore, if the vehicle type changes, the previously generated $\eta_{adv}$ cannot be used. Second, most existing methods only minimize the target class confidence score for $y$. While this may cause misclassification as a different class than $y$, the object detection itself may remain. Meanwhile, our proposal employs a method that applies an advanced and object-independent texture mapping approach (i.e., triplanar mapping), which later proves to solve the universality issues. %in 3D objects.
Furthermore, to address both limitations, we introduce Eq. \ref{eq:loss_final}, %these two or the second?
\begin{equation}
\argmin_{\eta_{adv}}\mathbb{E}_{v\in V,y\in Y}L(h(f(x_v, \eta_{adv})), y)
\label{eq:loss_final}
\end{equation}
where $V$ denotes the available vehicle types while $Y$ signifies the available classes of the detector. Solving Eq. \ref{eq:loss_final} aims to: (1) improve universality by generating an attack pattern that is applicable to various vehicles simultaneously, and (2) enhance robustness by minimizing confidence scores for all valid classes to avoid detection as objects themselves.

\subsection{ACTIVE Framework}
\label{subsec:active_framework}

\begin{figure*}[!t]
\begin{center}

% \centerline{\includegraphics[width=0.85\textwidth]{latex/images/dtn_architecture_fix_ver3.png}}
\centerline{\includegraphics[width=0.9\textwidth]{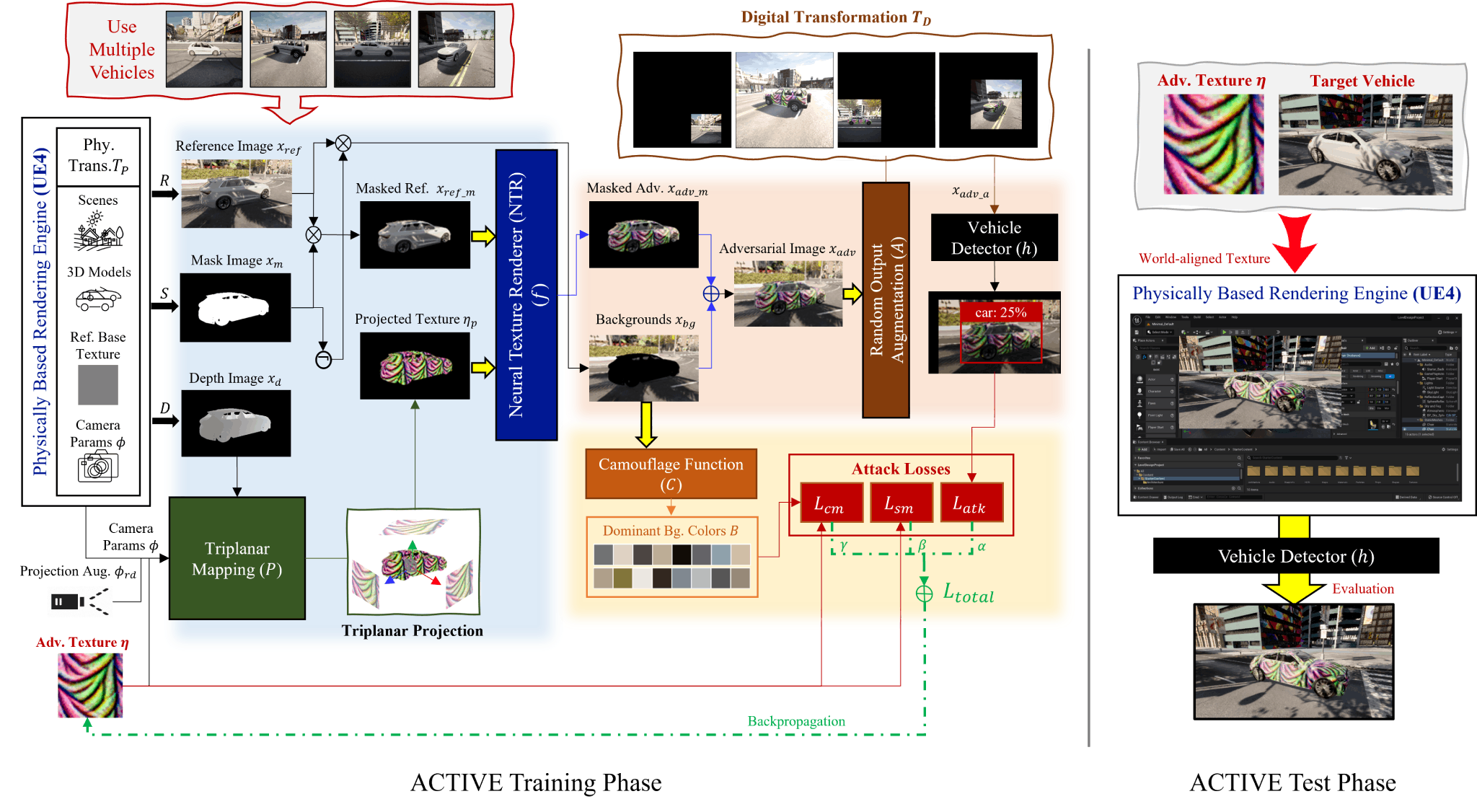}}
\vspace{-0.3cm}
\caption{ACTIVE framework for generating universal and robust adversarial camouflage.}
\label{fig:DTAv2_framework}
\vspace{-1cm}
\end{center}
\end{figure*}

To generate universal and robust adversarial camouflage, we propose the ACTIVE framework, which employs an object-independent texture mapping with a neural renderer and a new attack loss function to cause the vehicle undetectable. The overall framework is as illustrated in Fig. \ref{fig:DTAv2_framework}.

\textbf{Triplanar Mapping (TPM).}
We introduce the use of \textit{triplanar mapping} \cite{nicholson2008gpu}, a texture mapping technique that applies textures to objects by projecting them from three directions using their surface coordinate and surface normal that can be extracted from the depth image. We find this method particularly beneficial for generating object-independent adversarial textures since it does not rely on object-specific texture maps. Thus, we can optimize a common adversarial camouflage for multiple vehicles, making our attack instance-agnostic. Further, we introduce the Neural Texture Renderer (NTR), our improvement of DTN method by Suryanto et al. \cite{Suryanto_2022_CVPR}, which works well with our triplanar mapping while effectively preserving various physical characteristics. To our knowledge, we are the first to refine triplanar-mapped texture as the input to the neural renderer for enabling adversarial texture optimization across 3D instances. Our NTR enhances the efficiency of DTN by removing unnecessary elements of DTN, which the detail can be found in the Supplementary Material.

\textbf{Stealth Loss.}
We propose \textit{stealth loss}, a novel attack loss function for improving robustness. It considers two representative scores used in object detection models: the class confidence score and the objectness score. First, we minimize the objectness score, which determines the presence of an object in the detector, such as in YOLO families. Moreover, instead of minimizing the maximum confidence score for a specific class, we minimize the maximum confidence score across all classes, making our attack class-agnostic. This approach does not only mislead the model to misclassification, but also considers the possibility of the box being empty of objects. The attack loss $L_{atk}$ is written as Eq. \ref{eq:attack_loss},
\begin{equation*}
    h_d(x) = \begin{cases}
    h_c(x) \times h_o(x), &\text{if $IoU(h_b(x), gt)>t$} \\
    0, &\text{otherwise}
    \end{cases}
\end{equation*}
\begin{equation}
    L_{atk}(x) = f_{log}(\max(h_d(x)))
    \label{eq:attack_loss}
\end{equation}
where $x$ is an input image of a vehicle detector $h$, $h_c(x)$ is the confidence score of $h$ for $x$, $h_o(x)$ is the objectness score of $h$ for $x$, $h_b(x)$ is the detection result of $h$ for $x$ in the form of the bounding box, $IoU(h_b(x), gt)$ is the Intersection over Union (IoU) between the bounding box $h_b(x)$ and the ground-truth, $gt$, and $t$ is a custom IoU threshold. We define $h_d(x)$ as a detection score, which is the product of the confidence score and the objectness score. $f_{log}(n)=-log(1-n)$ is a log loss when the ground truth is zero. Then, minimizing $L_{atk}$ has the effect of minimizing both the confidence and objectness scores. Note that we only assign a value to the detection score for valid boxes with IoU greater than $t$, otherwise set it as zero. Thereby, the valid detection performance of the object detection model can be effectively lowered because the loss is applied only to the box that detects the object closely, and otherwise is excluded.

\textbf{Random Output Augmentation (ROA).}
We propose an \textit{ROA module} to enhance the texture robustness by various digital transformations \cite{EoT}. Specifically, the module takes the output of the NTR, i.e., the adversarial example, and further augments it by applying random transformations such as scaling, translation, brightness, and contrast. While the NTR already provides robustness against \textit{various physical} transformations, ROA allows for attaining an additional level of robustness with \textit{digital} transformations to simulate changes happening in the real world to a certain extent, which to the best of our knowledge, have not been considered in most adversarial camouflage methods.

\textbf{Smooth Loss.}
We utilize a smooth loss (i.e., Total Variation (TV) loss \cite{Mahendran_2015_CVPR}) to improve the smoothness of the generated camouflage, which we define as $L_{sm}$ in Eq. \ref{eq:smooth_loss},
\begin{equation}
    L_{sm}(\eta) = \frac{1}{N_{sm}}\sum_{i,j}f_{log}(|\eta_{i,j}-\eta_{i+1,j}|)+f_{log}(|\eta_{i,j}-\eta_{i,j+1}|)
    \label{eq:smooth_loss}
\end{equation}
where $\eta_{i,j}$ is a pixel in a texture, $\eta$, at coordinate $(i,j)$ and $N_{sm}=(H-1)\cdot(W-1)$ is a scale factor with texture image height, $H$, and texture image width, $W$. We slightly modify the loss for scale adjustment and normalization compared to the original TV loss. That is, $L_{sm}$ is low when the values of adjacent pixels are close to each other. Thus, minimizing $L_{sm}$ improve the smoothness of the adversarial camouflage.

\textbf{Camouflage Loss.} We propose a camouflage loss to improve camouflage for human vision as well as computer vision. There are several ways to computationally measure camouflage effectiveness, the most common method being to measure the target-background similarity \cite{toet2020}. We use a method to extract the most dominant background colors and force the object texture color to be similar. First, we employ a camouflage function based on K-means clustering to extract the most dominant colors from all background images \cite{yang2014, jia2019, xiao2020}. Next, we utilize Non-Printability Score (NPS) loss \cite{sharif2016} to regulate the object texture color set. While it was originally proposed to craft a color set that is comprised mostly of colors reproducible by the printer, we replace the printable color set used in the original NPS loss with the most dominant background color set. The camouflage loss, $L_{cm}$, can be expressed as Eq. \ref{eq:camouflage_loss},
\begin{equation}
    L_{cm}(\eta, B) = \frac{1}{N_{cm}}\sum_{i,j}f_{log}(\min_{b \in B}|b-\eta_{i,j}|)
    \label{eq:camouflage_loss}
\end{equation}
where $N_{cm}=H \cdot W$ is a scale factor of camouflage loss, and $B$ is the most dominant background color set. We can acquire the adversarial camouflage, which has a similar color to the background, by minimizing $L_{cm}$. Finally, our total loss, $L_{total}$, is constructed as Eq. \ref{eq:total_loss},
\begin{equation}
    L_{total} = \alpha L_{atk} + \beta L_{sm} + \gamma L_{cm}
    \label{eq:total_loss}
\end{equation}
where $\alpha$, $\beta$, and $\gamma$ are the weights to control the contribution of each loss function. The full pipeline of the ACTIVE framework for generating adversarial camouflage by minimizing $L_{total}$ is illustrated in Fig. \ref{fig:DTAv2_framework} and Alg. \ref{alg:adv_dtav2}.

\begin{algorithm}[]
    \caption{ACTIVE adversarial camouflage generation}
    \label{alg:adv_dtav2}
\begin{algorithmic}
    \STATE {\bfseries Input:} Physical transformation set $T_P$, Digital transformation set $T_D$, Rendering function $R$, Segmentation function $S$, Depth function $D$, Triplanar mapping $P$, NTR $f$, ROA module $A$, 
    \STATE {\bfseries Output:} Adversarial camouflage $\eta$ \\ 
    (1) Export $x_{ref}$, $x_m$, $x_d$, $x_{ref\_m}$, and $x_{bg}$ from the rendering engine
    \STATE $x_{ref} \gets R(T_P), x_m \gets S(T_P), x_d \gets D(T_P)$
    \STATE $x_{ref\_m} \gets x_{ref} \times x_m, x_{bg} \gets x_{ref} \times \neg x_m$ \\
    (2) Export $B$ by Camouflage function $C$
    \STATE $B \gets C(x_{bg})$ \\
    (3) Generate adversarial camouflage $\eta$
    \STATE Initialize $\eta$ with random values
    \FOR {number of training iterations}
    \STATE Select the minibatch sample from $x_{ref\_m}, x_m, x_{bg}$
    \STATE Derive $\phi$, $\phi_{rd}$ corresponding to each $x_{ref\_m}$ from $T_P$
    \STATE $\eta_p \gets P(\eta, x_d, \phi+\phi_{rd})$
    \STATE $x_{adv\_m} \gets f(x_{ref\_m}, \eta_p)$
    \STATE $x_{adv} \gets x_{adv\_m} + x_{bg}$
    \STATE $x_{adv\_a} \gets A(x_{adv}, T_D)$
    \STATE Calculate $L_{atk}(x_{adv\_a}), L_{sm}(\eta), L_{cm}(\eta, B)$ by Eq. \ref{eq:attack_loss}, \ref{eq:smooth_loss}, \ref{eq:camouflage_loss}
    \STATE Set $L_{total}$ by Eq. \ref{eq:total_loss}
    \STATE Update $\eta$ for minimizing $L_{total}$ via backpropagation
    \ENDFOR
\end{algorithmic}
\end{algorithm}

%------------------------------------------------------------------------
\section{Experiments} %%HARAS START HBS MAKAN
\label{sec:experiments}
We perform comprehensive experiments to investigate the performance of our proposed method on multiple aspects, including robustness and universality. 
% % %%%WTD: Ask: can it be removed?: 
% All experiments consider the physical parameters either by physically-based simulation or real-world evaluation for a more physically-realizable and practical proof. 
Each experiment and its comparison with previous works are designed with specific criteria. However, due to limited space, we only present the essential information and leave the detail to the Supplementary Materials.

% Due to limited space, we only present the essential technical information and main results, whereas the detail can be found in the Supplementary Materials.

\subsection{Implementation Details}
%%%%NEW COMPACT VERSION FOR DATASET
\textbf{Environment and Datasets.} We implement our attacking framework using TensorFlow 2 \cite{tensorflow} and utilize CARLA \cite{carla} on Unreal Engine 4 (UE4) \cite{unrealengine} as a physically-based simulator for training and evaluation, following \cite{Suryanto_2022_CVPR}. We synthesize our own dataset for training and evaluation with various photo-realistic settings. We select five types of cars on CARLA for attack texture generation and robustness evaluation, and another different five for universality evaluation, where they are excluded from texture optimization.
For NTR model training, a total of 50,625 and 150,000 photo-realistic images are used for training and testing, respectively. Regarding attack pattern generation, another 15,000 images for reference are employed. As for attack evaluation, we render the generated attack pattern on multiple cars using world-aligned texture in Unreal Engine to produce a repeated pattern while ignoring the texture UV Map, with 14,400 images for a single texture evaluation. 

\textbf{NTR.} We build an NTR with four encoder-decoder layers using DenseNet architecture \cite{huang2017densely} (following \cite{Suryanto_2022_CVPR}) and train using 20 epochs. Our experiment verifies that the network trained with only nine selected colors are able to generalize 50 random colors on the test, achieving 0.985 SSIM (comparable to 0.986 SSIM in the original DTN setting), but with 82\% less data. Details in Supplementary Material.

\textbf{Attack Parameters.} Our attack texture is optimized using Adam \cite{kingma2014adam} with 30 epochs. For ROA, we use $0.25$ random brightness, $[0.75, 1.5]$ random contrast, and $[0.25, 1.0]$ random scale. For projection augmentation on triplanar mapping, we use $[-0.5, 0.5]$ random shift and $[-0.25,  0.25]$ random scale. For loss hyperparameters, we use $\alpha=1.0$ with IoU threshold $t=0.5$, $\beta=0.25$, and $\gamma=0.25$ as default. Also, we set the base texture resolution to $64 \times 64$.

\subsection{Robustness Evaluations}
\textbf{From Digital to Physical Simulation.} 
% \hl{Goal: Evaluate the effectiveness of rendering methods for each approach when transferred to UE4.}
First, we perform a comparative experiment to investigate the effectiveness of existing rendering methods, including ours, by evaluating the performance of attack textures in the original pipeline compared to physical simulation (UE4). For a fair comparison with prior works, we follow DAS \cite{Wang_2021_CVPR} and FCA \cite{fcaattack} by selecting the same simulated town and Audi E-Tron car in CARLA, then optimizing the textures targeting YOLOv3 \cite{redmon2018yolov3}. 
In detail, DAS and FCA utilize Neural Mesh Renderer (NMR) \cite{kato2018neural} to optimize the 3D car texture and attach it to the simulated town as background, while DTA \cite{Suryanto_2022_CVPR} utilizes Repeated Texture Projection (RTP) and DTN to simply project repeated texture and render it to the reference image. Meanwhile, ours utilizes TPM and NTR for more accurate repeated texture mapping and rendering. 

\begin{figure}
\centerline{\includegraphics[width=\columnwidth]{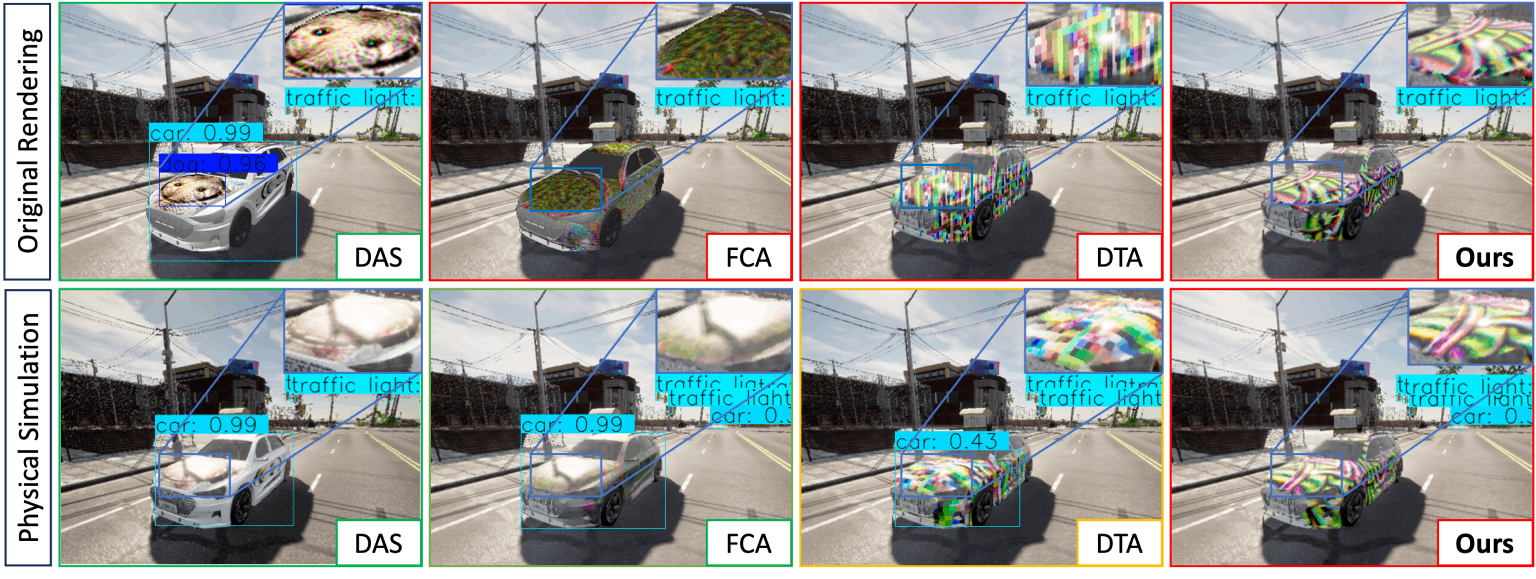}}
\vspace{-0.2cm}
\caption{Rendering comparison of white-box 3D adv. camouflage methods: Adv. example from original pipeline vs. Fully transferred to physical simulation (UE4). Zoom in.}
\label{fig:digital-to-physical-sim}
\end{figure}

As shown in Tab. \ref{tab:digital-to-physical-sim} and Fig. \ref{fig:digital-to-physical-sim}, adversarial example produced by our optimization pipeline results in a very high attack performance compared to related works, even after fully transferred to physical simulation. DAS exhibits poor performance due to its partial texture coverage which is aligned with \cite{fcaattack, Suryanto_2022_CVPR}. FCA can successfully evade the detection in the original pipeline but is fully detected in the physical simulation with a high score. We observe DAS and FCA textures are mostly obscured by light reflections, which cannot be represented by their rendering method. DTA, which considers physical transformations but inaccurate texture mapping, causes misdetection in the original pipeline but is still detected in the physical simulation with a low score. In contrast, ours can successfully evade detection almost perfectly, thanks to our accurate texture mapping.

\begin{table}[]
\caption{Digital-to-physical simulation comparison. Values are Average Precision@0.5 (\%) of car in YOLOv3.}
\vspace{-0.2cm}
\resizebox{\columnwidth}{!}{
\label{tab:digital-to-physical-sim}
\begin{tabular}{l|l|c|c}
\hline
Methods                       & Orig. Rendering & \multicolumn{1}{l|}{Adv. Exm.} & \multicolumn{1}{l}{Phy. Sim.} \\ \hline
Normal                        & UE4            & -                              & 99.57                         \\ \hline
DAS \cite{Wang_2021_CVPR}     & + NMR          & 88.55                          & 96.09                         \\ \hline
FCA \cite{fcaattack}          & + NMR          & 52.05                         & 92.28                         \\ \hline
DTA \cite{Suryanto_2022_CVPR} & + RTP \& DTN   & 16.91                          & 41.95                         \\ \hline
Ours                          & + TPM \& NTR   & \textbf{1.28}                 & \textbf{7.29}                 \\ \hline
\end{tabular}
}
\end{table}

% This section describes the attack comparison of our method with the previous works. 
\textbf{Attack Comparison on Physically-Based Simulation.}
% For a fair comparison with prior works, we re-optimize all textures on YOLOv3 \cite{redmon2018yolov3}, following \cite{fcaattack}. We also evaluate all methods on SSD \cite{liu2016ssd}, Faster R-CNN (FrRCNN) \cite{ren2015faster}, and Mask R-CNN (MkRCNN) \cite{he2017mask} as the black-box model. 
% Note that this part only evaluates AP@0.5 on Audi Etron, because it is where the original FCA texture is optimized (and, hence, can only be applied there). Since we only assess the performance on a single instance, the number of samples per transformation is increased to 50, yielding 7,200 images per evaluation. 
We run a more extensive attack comparison by using diverse camera poses and evaluated models. Specifically, we follow FCA to evaluate all methods on SSD \cite{liu2016ssd}, Faster R-CNN (FrRCNN) \cite{ren2015faster}, and Mask R-CNN (MkRCNN) \cite{he2017mask} as the black-box model while keeping YOLOv3 as the target.

\begin{table}[]
\caption{Attack comparison on physically-based simulation. Values are Average Precision@0.5 (\%) of car.}
\label{tab:attack-comparison-evals}
\vspace{-0.2cm}
\resizebox{\columnwidth}{!}{
\begin{tabular}{l|cc|cc}
\hline
\multicolumn{1}{c|}{\multirow{2}{*}{Methods}} & \multicolumn{2}{c|}{Single-Stage Detector} & \multicolumn{2}{c}{Two-Stage Detector} \\ \cline{2-5} 
\multicolumn{1}{c|}{}         & YOLOv3 & SSD   & FrRCNN  & MkRCNN \\ \hline
Normal                        & 90.67  & 92.20 & 87.84   & 94.30  \\ \hline
Random                        & 70.01  & 79.01 & 74.41   & 69.09  \\ \hline
Naive Cam.                    & 60.26  & 59.48 & 58.71   & 67.21  \\ \hline
DAS \cite{Wang_2021_CVPR}     & 88.53  & 84.28 & 84.31   & 88.48  \\ \hline
FCA \cite{fcaattack}          & 76.92  & 76.35 & 74.09   & 80.27  \\ \hline
CAMOU \cite{Zhang_2019_ICLR}  & 59.20  & 68.02 & 67.84   & 62.31  \\ \hline
ER \cite{Wu_CoRR_2020}        & 58.02  & 70.77 & 62.45   & 61.30  \\ \hline
DTA \cite{Suryanto_2022_CVPR} & 33.33  & 47.80 & 47.44   & 49.85  \\ \hline
Ours                          & \textbf{19.52}      & \textbf{33.56}      & \textbf{41.70}     & \textbf{45.08}    \\ \hline
\end{tabular}
}
\end{table}

\begin{figure}
\centerline{\includegraphics[width=\columnwidth]{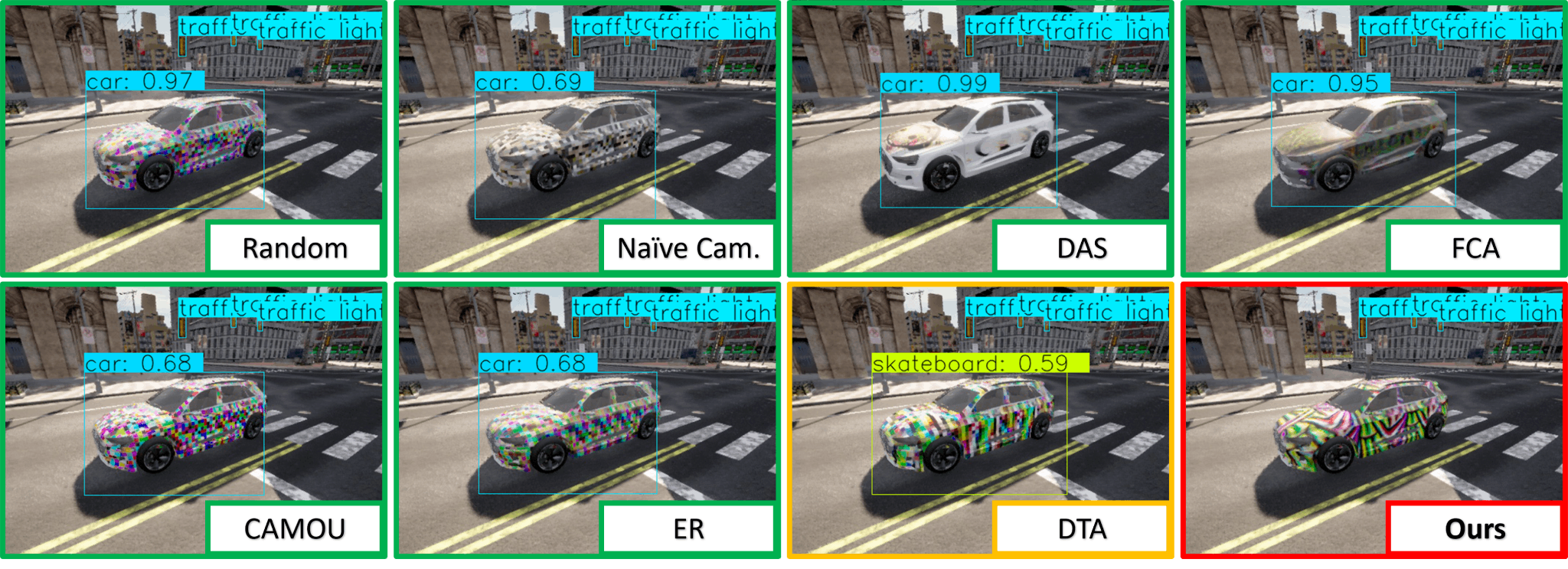}}
\vspace{-0.2cm}
\caption{Attack comparison samples. Zoom for detail.}
\label{fig:attak_comparison_samples}
\end{figure}
% \vspace{-0.3cm}

The attack comparison results are depicted in Tab. \ref{tab:attack-comparison-evals}, from which we can infer that our method consistently has the best attack performance on all models, both on a single-stage and two-stage detector. Note that we include a random and naive camouflage pattern to show the model's robustness against arbitrary textures.
Again, we can see that DAS and FCA, whose limitation prevents them from considering physical parameters during optimization, yield much lower performance.
% , \hl{and sometimes does not even work when transferred to physically-based simulation compared to other method}. 
Also, while CAMOU \cite{Zhang_2019_ICLR} and ER \cite{Wu_CoRR_2020} consider physical parameters from the simulator during optimization, such black-box attacks do not guarantee an optimum result. On the other hand, DTA %we don't need to praise DTA, so i changed the tone
\cite{Suryanto_2022_CVPR} yields much better results since it accounts for both physical parameters and white-box attacks. Nevertheless, it consistently underperforms compared to ours. 

We also provide the model prediction sample of the compared methods in Fig. \ref{fig:attak_comparison_samples}, showing the model can still correctly predict Random, DAS, and FCA textured cars with high detection scores. 
% We highlight how the light reflection by the car's metallic material hides most of the FCA texture, which is not considered during their optimization.
% From the FCA example, we can learn the importance of considering physical parameters since the light reflection by the car's metallic material can hide most of the texture.
FCA failure illustrates the importance of considering physical parameters, as the car's metallic material may cause light reflections that can hide the texture.
Naive, CAMOU, and ER decrease the detection score, but still insufficiently to result in misclassification. Alternatively, DTA texture \textit{does} misclassify the object, but nowhere near our method, which renders the car undetected. %More detailed evaluation and samples are available in Supplementary Material.

\begin{figure}
% \centerline{\includegraphics[width=0.85\textwidth]{latex/images/dtn_architecture_fix_ver3.png}}
\centerline{\includegraphics[width=\columnwidth]{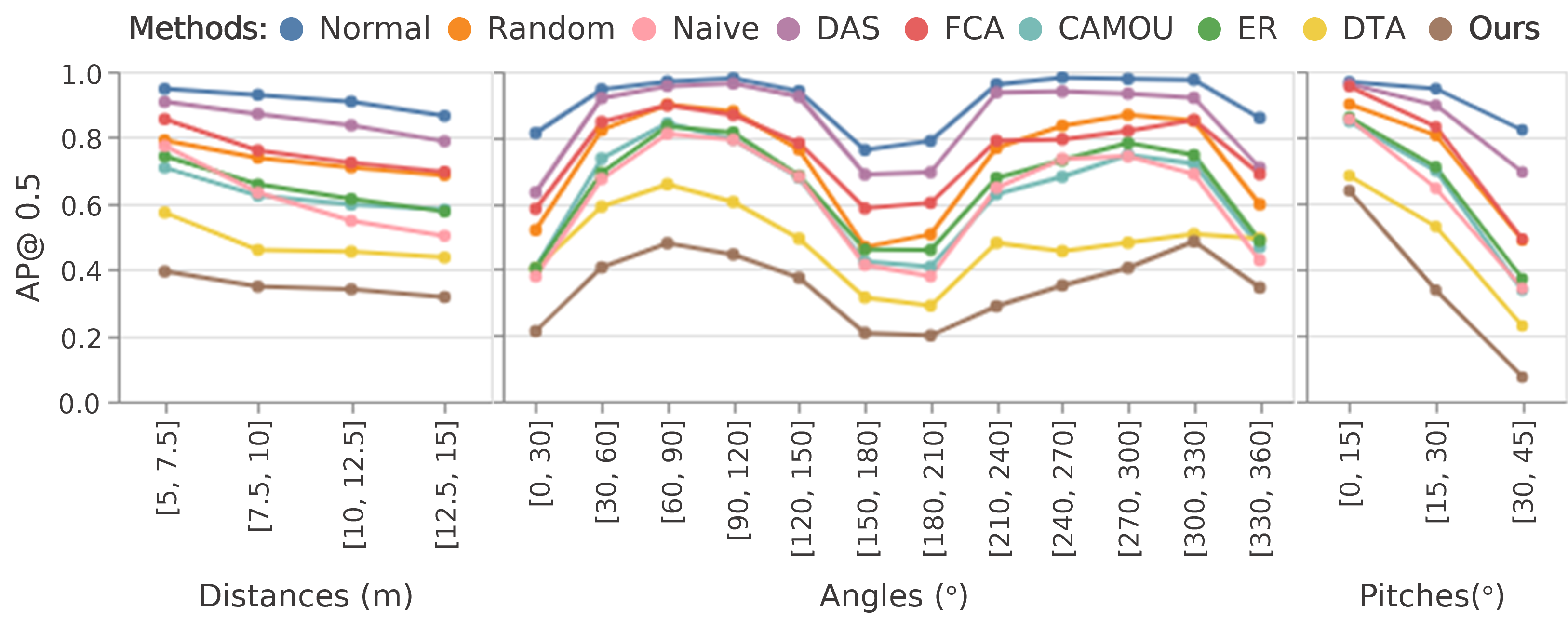}}
\vspace{-0.2cm}
\caption{Attack comparison on different camera poses. Values are AP@0.5 of the car averaged from all models.}
\label{fig:attack_comparison_graphs}
\end{figure}

Fig. \ref{fig:attack_comparison_graphs} shows the summarized performance of each camera pose; 
values are car AP@0.5, averaged from all evaluated models. It visualizes how our method invariably outperforms previous works on all viewing conditions. As implied, our method is relatively stable under various distances compared to other methods. Additionally, we observe that varied camera pitches have a greater impact on attack performance than other poses. %Add analysis about side-view$

\subsection{Universality Evaluations}
% This section provides extended experiments to evaluate how our proposal and related works attack performance in different settings, and also on different tasks. 

\textbf{Transferability Comparison on Different Settings.} For this experiment, we use various brand-new models, including YOLOv7 (YLv7) \cite{wang2022yolov7}, Dynamic R-CNN (DRCN) \cite{zhang2020dynamic}, Sparse R-CNN (SRCN) \cite{sun2021sparse}, Deformable DETR (DDTR) \cite{zhu2021deformable}, and Pyramid Vision Transformer (PVT) \cite{wang2021pyramid} to evaluate transferability over diverse modern architectures. All models are considered black-box except for YOLOv3 \cite{redmon2018yolov3}, used for texture generation in the previous section. The results of car AP@0.5 are presented in Tab. \ref{tab:transferability-evals}, showing that our method has the best performance on all models. It is interesting to see that YOLOv7 is so robust that Random, UPC \cite{Huang_2020_CVPR}, CAMOU, and ER can only slightly reduce the car AP@0.5. Nevertheless, even in the black-box setting, our method can significantly reduce the YOLOv7 performance. More surprisingly, our attack method is robust even in the case of the Transformer-based vision models, although they are generally known to be more resilient to adversarial attacks than CNNs and have low transferability from CNNs \cite{mahmood2021robustness,benz2021adversarial,aldahdooh2021reveal}. Considering most studies claiming their robustness only conduct digital attack experiments, it is a discovery that transformer-based model can be vulnerable to physical attacks such as our adversarial camouflage. This indicates our method is also \textit{model-agnostic}. 

\begin{table}[]
\caption{Universality evaluation where target objects, models, and scenes differ from the training. AP@0.5 (\%) of car.}
\label{tab:transferability-evals}
\vspace{-0.2cm}
%\resizebox{\columnwidth}{!}{
\scalebox{0.82}{
\begin{tabular}{l|cc|cc|cc}
\hline
\multirow{2}{*}{Methods}      & \multicolumn{2}{c|}{Single-Stage}       & \multicolumn{2}{c|}{Two-Stage} &  \multicolumn{2}{c}{Transformer} \\ \cline{2-7} 
                              & YLv3           & YLv7            & DRCN           & SRCN           & DDTR          & PVT            \\ \hline
Normal                        & 85.98          & 92.54           & 83.20          & 82.55          & 83.83          & 88.59          \\ \hline
Random                        & 66.93          & 85.65           & 69.59          & 70.43          & 47.77          & 78.42          \\ \hline
Naive Cam.                    & 60.73          & 69.62           & 55.68          & 64.08          & 49.91          & 67.45          \\ \hline
UPC \cite{Huang_2020_CVPR}    & 83.48          & 90.00           & 80.58          & 78.36          & 73.64          & 86.61          \\ \hline
CAMOU \cite{Zhang_2019_ICLR}  & 60.19          & 82.67           & 63.78          & 62.93          & 33.33          & 68.77          \\ \hline
ER \cite{Wu_CoRR_2020}        & 58.90          & 82.18           & 62.06          & 62.54          & 37.34          & 73.96          \\ \hline
DTA \cite{Suryanto_2022_CVPR} & 32.24          & 59.39           & 45.72          & 43.68          & 19.43          & 56.05          \\ \hline
Ours                          & \textbf{22.55} & \textbf{41.55}  & \textbf{30.38} & \textbf{42.00} & \textbf{14.69} & \textbf{51.54} \\ \hline
\end{tabular}
}
\end{table}

\begin{figure}
\centerline{\includegraphics[width=\columnwidth]{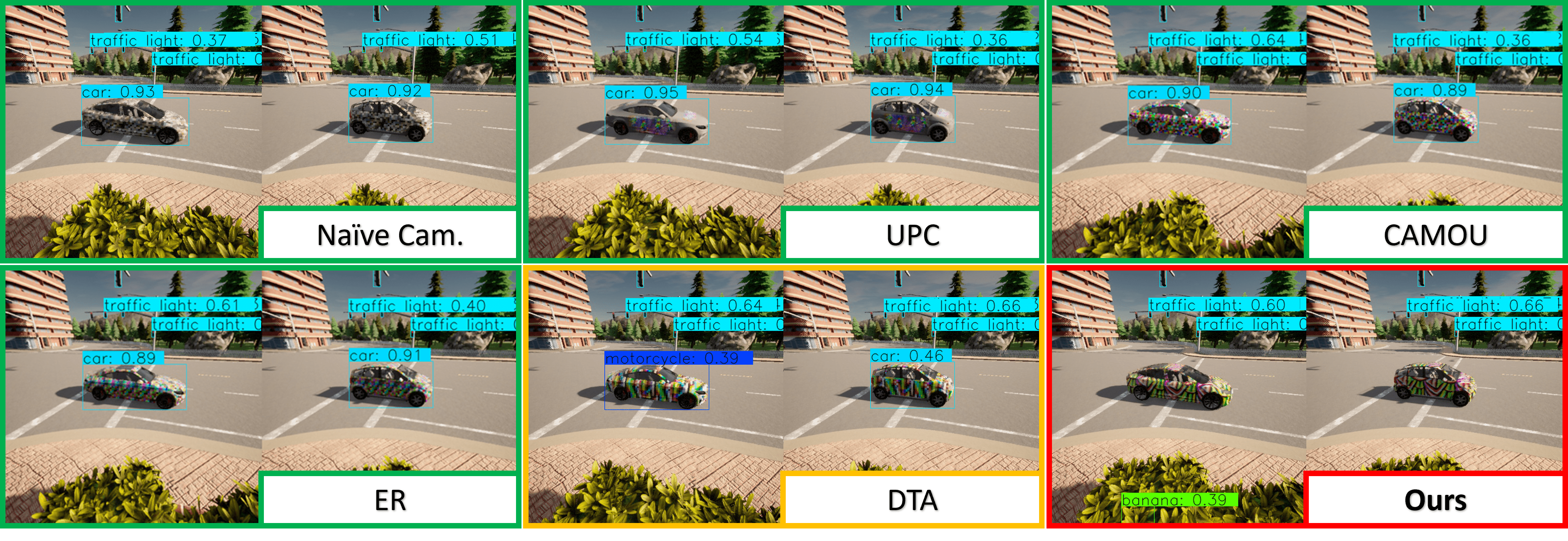}}
\vspace{-0.2cm}
\caption{Transferability to different settings. Zoom in.}
\label{fig:universality_comparison}
\end{figure}

Fig. \ref{fig:universality_comparison} displays the sample prediction results of YOLOv7 on evaluated methods. The model can correctly predict naive camouflage, UPC, CAMOU, and ER textured cars, but with consistently higher detection scores compared to previous evaluations. One of the DTA-textured cars results in misclassification, but the other is still detected correctly with a fairly low detection score. It is different from ours, which consistently makes the cars undetected, further verifying the universality of our method as \textit{instance-agnostic}.

\begin{figure}
\centerline{\includegraphics[width=\columnwidth]{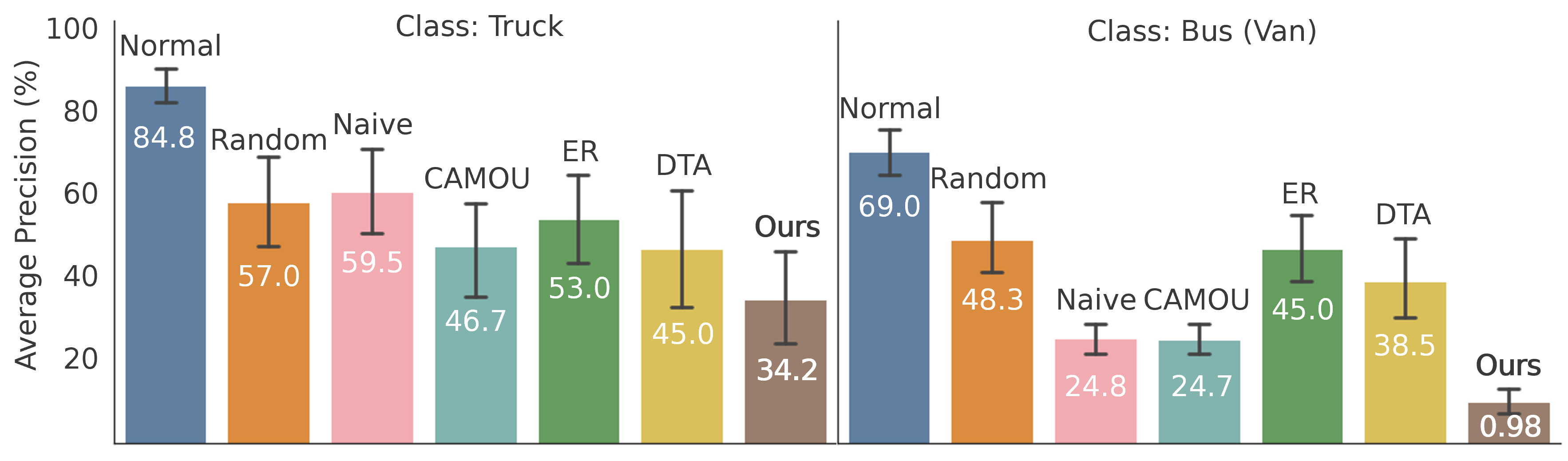}}
\vspace{-0.2cm}
\caption{Transferability to different class (Truck and Bus).}
\label{fig:class_comparison}
\end{figure}

\textbf{Transferability Comparison on Different Class.} 
We evaluate the textures when applied to different classes (i.e., truck and bus), use all evaluated models in the last experiment, and group the result in Fig. \ref{fig:class_comparison}. 
As shown, our method constantly outperforms others, demonstrating that ACTIVE is \textit{class-agnostic}, thanks to our stealth loss that considers all instead of specific class labels during optimization. 

\begin{figure}
\centerline{\includegraphics[width=\columnwidth]{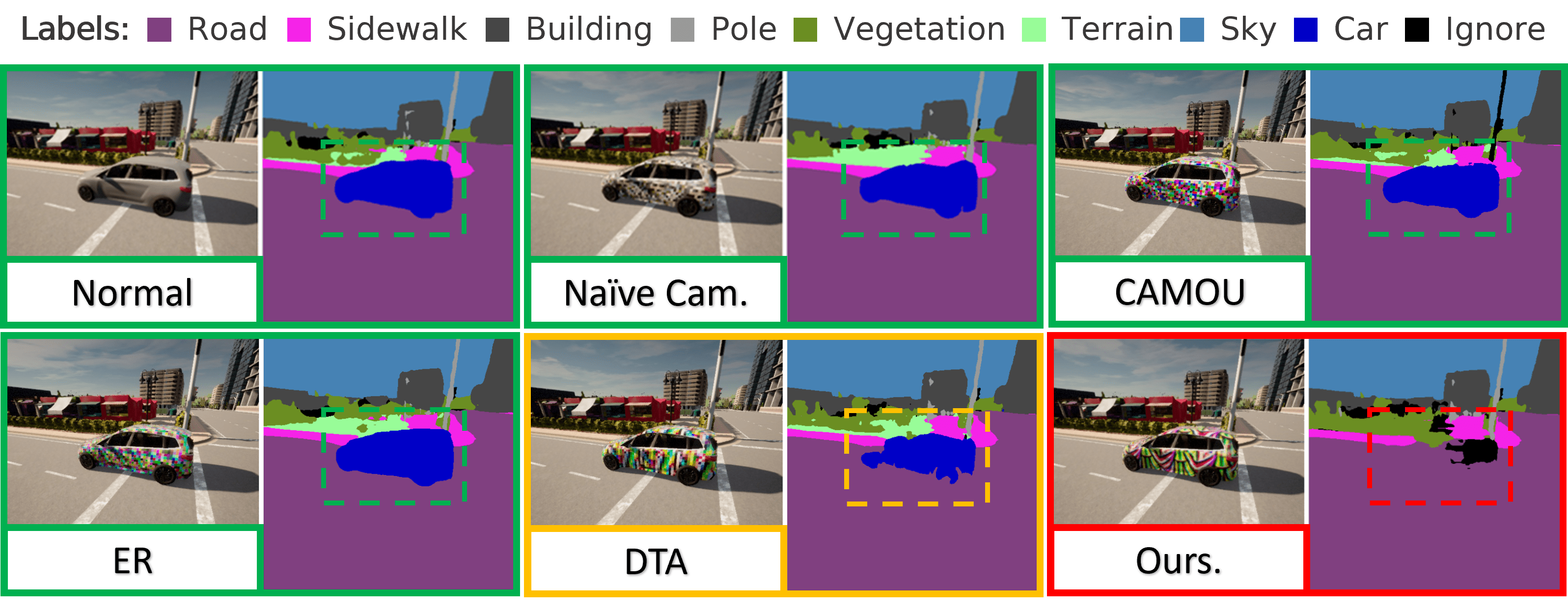}}
\vspace{-0.2cm}
\caption{Transferability to segmentation model. Zoom in.}
\label{fig:segmentation_comparison}
\end{figure}

\textbf{Transferability Comparison on Different Task.}
We further evaluate all methods' universality by testing the generated texture on a publicly available pretrained segmentation model: MaX-DeepLab-L \cite{wang2021max} and Axial-DeepLab \cite{wang2020axial} with SWideRNet \cite{swidernet_2020} Backbone. We test on both Cityscape \cite{Cordts2016Cityscapes} and COCO \cite{coco} pretrained models, and exclude high-pitch cameras because the Cityscape dataset only uses low-pitch data. Furthermore, we only evaluate the pixel accuracy (\%) of the car label to show how the texture can downgrade the prediction of the target object. Again, the experiment result in Tab. \ref{tab:segmentation-evals} shows that our method significantly outperforms the previous works. Additionally, Fig. \ref{fig:segmentation_comparison} visualizes the sample of how our method makes the car invisible from the segmentation model (either predicted as road or ignored), while other methods correctly predict as car, which also confirms our method produces \textit{task-agnostic} texture.

\begin{table}[]
\caption{Universality evaluation on a different task (i.e., Segmentation Model). Values are pixel accuracy (\%) of car.}
\label{tab:segmentation-evals}
\vspace{-0.2cm}
\resizebox{\columnwidth}{!}{
\begin{tabular}{l|cc|c}
\hline
\multirow{2}{*}{Methods}        & \multicolumn{2}{c|}{Cityscape Pretrained} & COCO Pretrained \\ \cline{2-4} 
                                & MaX-DL-L               & Axl-DL-SW      & MaX-DL-L        \\ \hline
Normal                          & 90.70                  & 92.76            & 89.77           \\ \hline
Random                          & 78.20                  & 88.24            & 81.68           \\ \hline
Naive Cam.                      & 50.23                  & 74.78            & 64.11           \\ \hline
UPC   \cite{Huang_2020_CVPR}    & 74.70                  & 83.66            & 75.85           \\ \hline
CAMOU \cite{Zhang_2019_ICLR}    & 62.12                  & 71.66            & 64.18           \\ \hline
ER    \cite{Wu_CoRR_2020}       & 71.55                  & 85.70            & 71.17           \\ \hline
DTA   \cite{Suryanto_2022_CVPR} & 31.53                  & 55.68            & 32.85           \\ \hline
Ours                            & \textbf{17.45}         & \textbf{32.04}   & \textbf{23.46}  \\ \hline
\end{tabular}
}
\end{table}

\begin{figure}
\centerline{\includegraphics[width=\columnwidth]{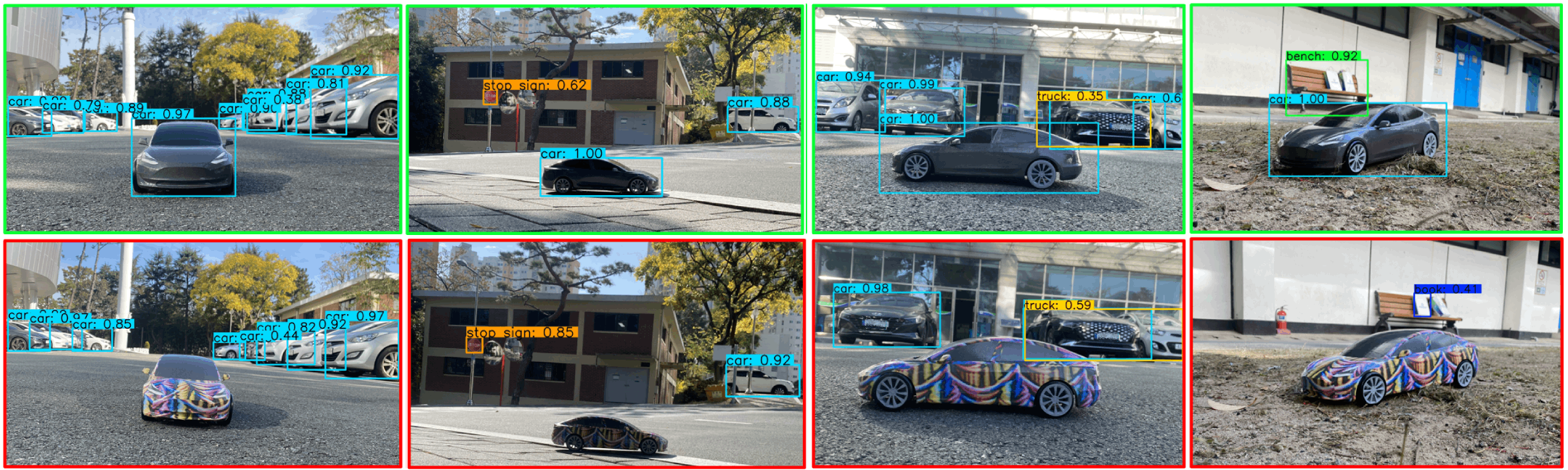}} 
\vspace{-0.2cm}
\caption{Real-world evaluation using two scaled cars. The upper row is the normal car model, while the bottom row is the adversarial camouflaged car model. Zoom for detail.}
\label{fig:real-world}
\end{figure}

\textbf{Transferability to Real World.}
Following \cite{Suryanto_2022_CVPR}, we conduct a real-world evaluation by constructing two 1:10-scaled Tesla Model 3s with a 3D printer and wrapping the texture onto the body of the car: one for a normal and another for our camouflaged car targeting YOLOv3.
Fig. \ref{fig:real-world} shows the normal car model is well detected, whereas the adversarial camouflaged car model is not detected as a car at all. 
Furthermore, we also evaluate practical real-time object detectors widely used in the real world
to show that our method transfers to the real-world (i.e., \textit{domain-agnostic}): MobileNetV2 (MbNetv2) \cite{sandler2018mobilenetv2}, EfficientDet-D2 (EfDetD2) \cite{tan2020efficientdet}, YOLOX-L (YLX-L) \cite{ge2021yolox}, YOLOv7 (YLv7) \cite{wang2022yolov7}, and also the target model, YOLOv3 (YLv3), shown in Tab. \ref{tab:realworld-evals}.

\begin{table}[]
\caption{AP@0.5 (\%) of target car in real-world evaluation.}
\label{tab:realworld-evals}
\vspace{-0.2cm}
\resizebox{\columnwidth}{!}{
\begin{tabular}{lccccc}
% \begin{tabular}{lC{0.7cm}C{1cm}ccc}
\hline
\multicolumn{1}{c}{\multirow{2}{*}{Methods}} & \multicolumn{5}{c}{Real-time Object Detector} \\ \cline{2-6} 
\multicolumn{1}{c}{} & YLv3            & MbNetv2        & EfDetD2         & YLX-L        & YLv7 \\ \hline
Normal               & 90.83           & 80.83          & 96.25            & 96.25          & 95.00 \\ \hline
Ours                 & \textbf{8.75}   & \textbf{26.27} & \textbf{26.25}   & \textbf{40.41} & \textbf{48.75} \\ \hline
\end{tabular}
}
\end{table}

\subsection{Ablation Study}

\begin{table}[]
\caption{Ablation study for each proposed module and loss.}
\label{tab:abs-study}
\vspace{-0.2cm}
\resizebox{\columnwidth}{!}{
\begin{tabular}{l |ccccc}
\hline
\multicolumn{1}{c|}{\textbf{Proposed Losses}}   & \multicolumn{5}{c}{\textbf{Proposed Modules} (Normal Car: \textit{93.01})}      \\ \cline{2-6} 
\multicolumn{1}{c|}{(YLv3 - Car AP@.5)} & Raw & w/ TPM & w/ ROA & \multicolumn{1}{c|}{Full} & Avg. (Std.) \\ \hline
$L_{atk}$ (DTA \cite{Suryanto_2022_CVPR})                                                    & 60.48                        & 48.21                          & 48.43                          & \multicolumn{1}{c|}{24.66}                        & 45$\pm$15                                 \\ \hline
{$L_{atk}$ (Stealth Loss)}                         & 60.13                        & 43.70                          & 38.85                          & \multicolumn{1}{c|}{22.48}               & 41$\pm$15                        \\
$L_{atk} + L_{sm}$                                                      & 58.90                        & 43.36                          & 36.00                          & \multicolumn{1}{c|}{\textbf{20.21}}                        & \textbf{39$\pm$16}                                 \\
$L_{atk} + L_{cm}$                                                      & 56.17                       & 42.04                          & 38.66                          & \multicolumn{1}{c|}{24.34}                        & 40$\pm$13                           \\
$L_{atk} + L_{sm} + L_{cm}$                                             & 57.81                       & 50.87                 & 40.70                 & \multicolumn{1}{c|}{28.22}               & 44$\pm$13                           \\ \hline
Avg. (Std.)                                                             & 59$\pm$2                         & 46$\pm$4                       & 41$\pm$5                       & \multicolumn{1}{c|}{\textbf{24$\pm$3}}            & \cellcolor[HTML]{333333} \\ \hline          
\end{tabular}
}
\vspace*{-1mm}
\end{table}

\textbf{Impact of Proposal on Performance and Naturalness.}
We evaluated our proposed components, including modules and losses, using ablation studies with default parameters. We used DTA \cite{Suryanto_2022_CVPR} as a baseline since our approach has a similar pipeline. The results in Table \ref{tab:abs-study} demonstrate that each of our proposed components plays a crucial role in enhancing the attack performance. Specifically, utilizing both TPM and ROA modules significantly impacts performance enhancement, with an average improvement of 35\%.

\begin{figure}
\centerline{\includegraphics[width=\columnwidth]{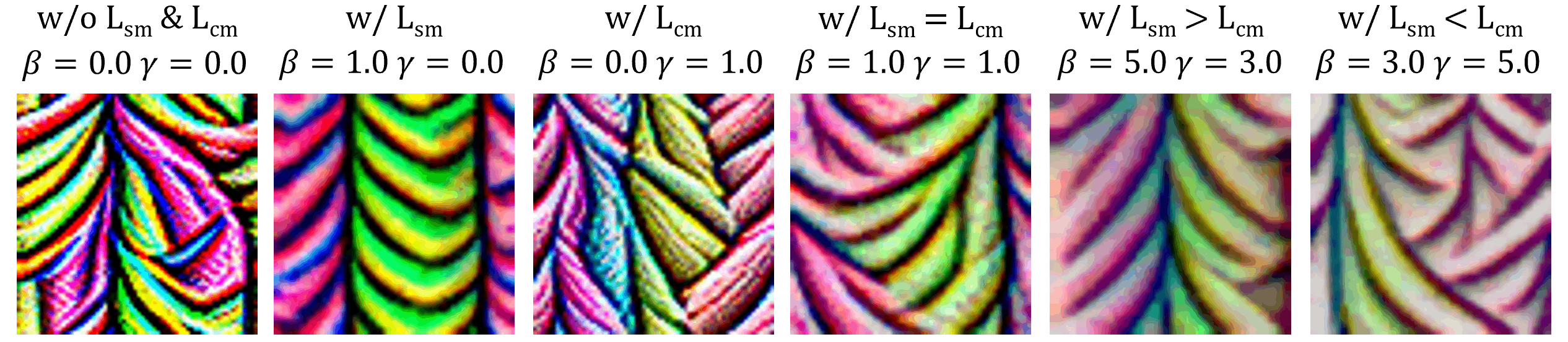}} 
\vspace{-0.2cm}
\caption{Visual textures comparison with different losses.}
\label{fig:naturalness}
\end{figure}

Although our proposed losses have a lesser impact on performance, they play an important role in texture naturalness, leading to trade-offs between the two. 
As illustrated in Fig. \ref{fig:naturalness},
% \hl{without $L_{sm}$, the texture is rough, whereas excluding $L_{cm}$ makes the texture more colorful and bright. 
omitting $L_{sm}$ outputs a rough texture, whereas excluding $L_{cm}$ makes it more colorful and bright.
Employing the stealth loss with only smooth loss yields the best performance, downgrading the car AP@0.5 to 20.21\%. More details are available in Supplementary Materials.
% When considering only attack performance, utilizing the stealth loss with the smooth loss achieves the best performance, downgrading the car AP@0.5 to 20.21%
% When maintaining full proposal except for the removed,
% \hl{Here we compare the performance of our method when one of our components is removed or replaced.} 
% excluding the ROA during optimization can result in a 20.6\% performance degradation, whereas replacing TPM with simple projection (i.e., RTP) can result in a 10.5\% decrement, as shown in Tab. \ref{tab:ablation-study-component}.

% \textbf{The Impact of Different Losses on Naturalness.}
% % \hl{TBD: Show the effect of raw loss, smooth loss, camouflage loss, and combined loss with respect to the performance and naturalness}
% As shown in Fig. \ref{fig:naturalness}, {$L_{sm}$ and $L_{cm}$ play an important role in texture naturalness. Particularly, without $L_{sm}$, the texture is rough, whereas excluding $L_{cm}$ may make the texture more colorful and bright. More details are available in Supplementary Materials. %However, both losses can lead to trade-offs with the attack's performance when not set correctly.}

%------------------------------------------------------------------------

\section{Discussions}
% Discuss the negative impact on safety critical applications such as self driving car.
% Discuss the importance of robust model that can stand to this camouflage before deploying the model to the real world.
% Discuss the limitation such, even more natural, the texture is still have the abstract shape.
% \hl{Harashta please rewrite}
% \textbf{Societal Implications.} As discussed in % by Suryanto et al.,
% \cite{Suryanto_2022_CVPR},  DTA as an attack poses harmful repercussions when implemented in a public setting, such as for attacking self-driving cars. The painting of cars with DTA texture, which can be done legally, also contributes to its danger \cite{Suryanto_2022_CVPR} since it is a highly feasible attack scenario. Our enhancements in DTAv2 that make it much more robust, universal, and natural to human vision \textemdash even succeeding in attacking YOLOv7 as the most up-to-date object detector \textemdash will further amplify its level of danger.
% \textbf{Societal Implications.} 
% \hl{Adversarial camouflage can harm self-driving cars since the car painting are legal} \cite{Zhang_2019_ICLR, Suryanto_2022_CVPR}\hl{. Enhancements in ACTIVE can make it more robust and universal will amplify the danger. As public detectors are still vulnerable, research in robustness is essential.}

\textbf{Societal Implications.} Adversarial camouflage poses harmful repercussions for self-driving cars since there exists a highly feasible attack scenario, e.g., legally painting cars with adversarial texture \cite{Zhang_2019_ICLR, Suryanto_2022_CVPR}. Enhancements in robustness and universality by ACTIVE can amplify the danger as existing public detectors are still highly vulnerable, signifying the importance of research in model robustness.

\textbf{Limitation.} Even though ACTIVE produces a more natural pattern similar to graffiti, the texture is still abstract.
% shape. Our next step is to generate patterns that can be tinkered to a specified shape.
% , unlike previous attacks (e.g., \hl{mention please}) that attached stickers on traffic signs which is considered a violation and 

\section{Conclusion}
\label{sec:conclusion}
We have presented ACTIVE, a physical camouflage attack framework for 3D objects for enhanced universality and robustness. Verified in our comprehensive evaluations, ACTIVE surpasses the performance of existing works \textemdash and notably, demonstrates its capability as a model, instance, class, task, and domain-agnostic % adversarial attack 
framework.

\textbf{Acknowledgments.} This work was supported by the Agency For Defense Development Grant Funded by the Korean Government (UE221150WD), and by the MSIT (Ministry of Science and ICT), Korea, under the Convergence security core talent training business (Pusan National University) support program (IITP-2023-2022-0-01201) supervised by the IITP (Institute for Information \& Communications Technology Planning \& Evaluation).

{\small
\bibliographystyle{ieee_fullname}
\bibliography{main}
}

\newpage
% \begin{appendices}
\onecolumn
\appendix
\appendixpage
{\Large\centering \textbf{Supplementary Material}}

\section{Overview}
This supplementary material describes the details of our implementation and evaluation results that can not be included in the main paper due to page limit. Furthermore, additional experiments and more evaluation samples for all experiments on the main paper are also presented.

\section{Algorithm Details}
\subsection{Triplanar Mapping}

\begin{figure}[h]
    \centering
    \begin{subfigure}[b]{0.475\columnwidth}
        \centering
        \includegraphics[width=0.95\columnwidth]{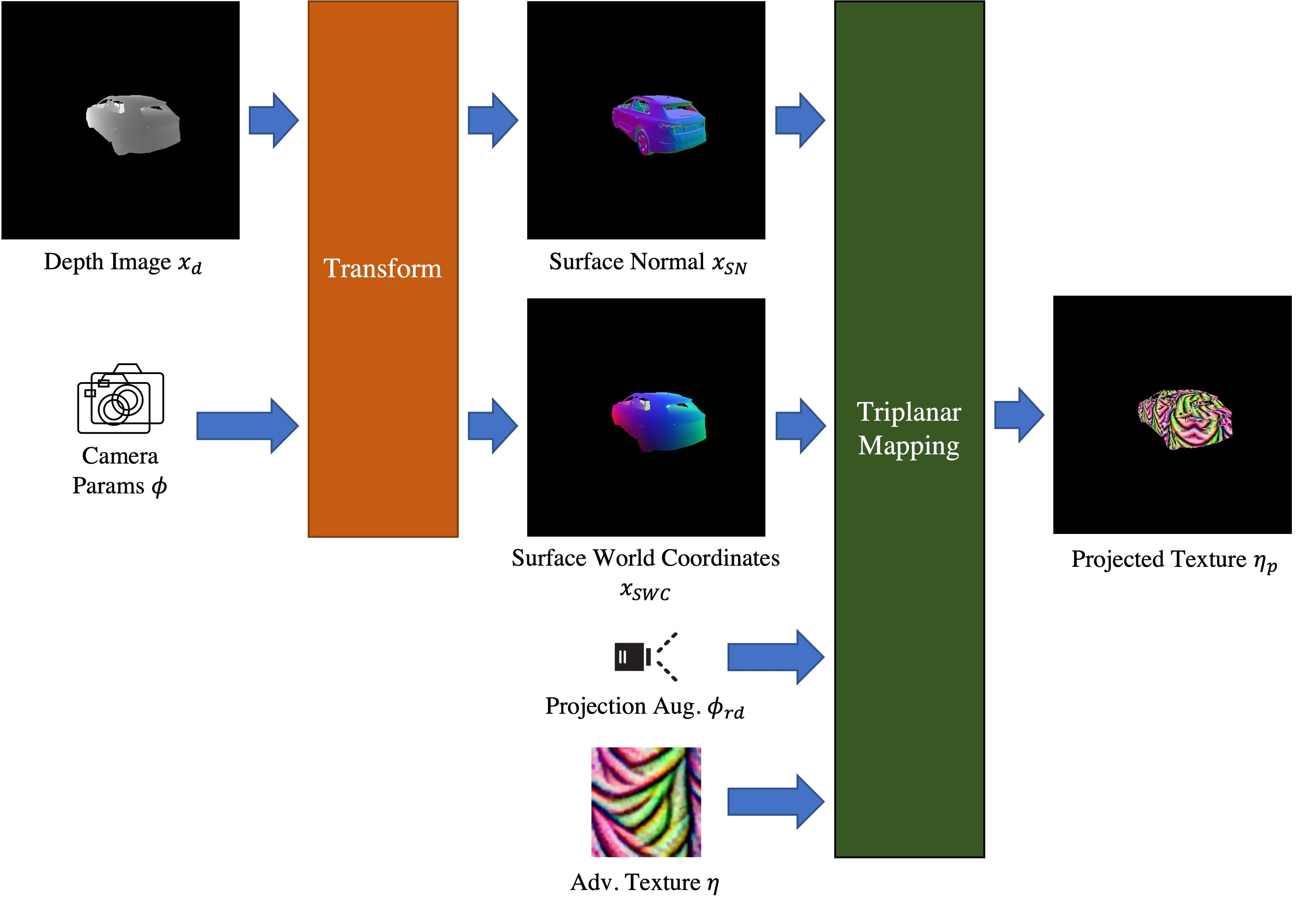}
        \caption{Triplanar mapping pipeline}
    \end{subfigure}
    \begin{subfigure}[b]{0.475\columnwidth}
        \centering
        \includegraphics[width=0.95\columnwidth]{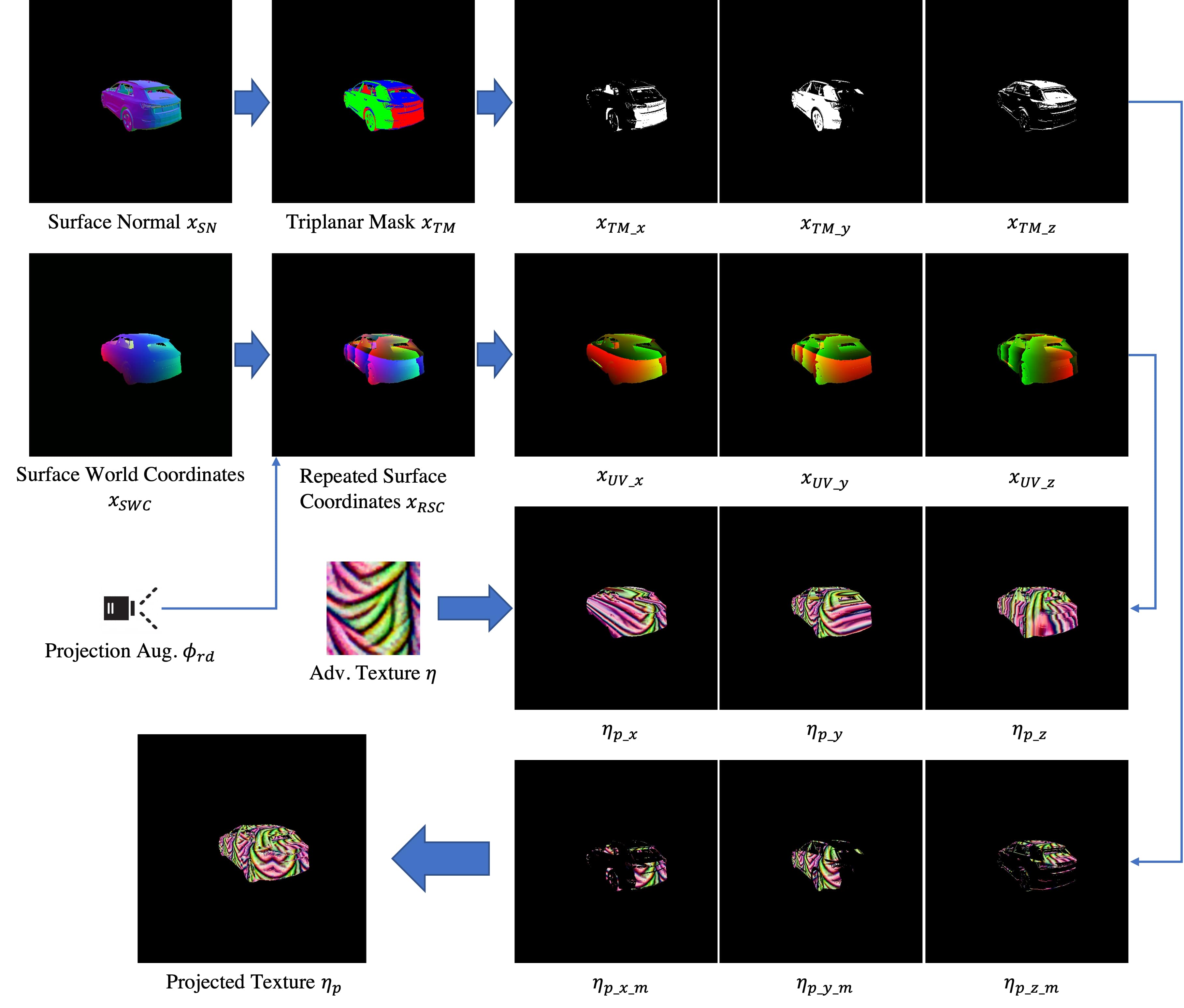}
        \caption{Triplanar mapping mechanism}
    \end{subfigure}
    \caption{Detailed pipeline and mechanism of Triplanar mapping \cite{nicholson2008gpu}}
    \label{fig:triplanar_mapping}
\end{figure}

In this section, we provide a detailed algorithm and figures for triplanar mapping \cite{nicholson2008gpu}. Algorithm \ref{alg:triplanar_mapping} and Figure \ref{fig:triplanar_mapping} describe the step-by-step process of extracting a projected texture using triplanar mapping.

\begin{algorithm}[H]
\caption{Triplanar Mapping Algorithm}
\label{alg:triplanar_mapping}
\begin{algorithmic}
\STATE {\bfseries Input:} Depth Image $x_d$, Camera Parameters $\phi$, Projection Augmentation $\phi_{rd}$, Adversarial Texture $\eta$
\STATE {\bfseries Output:} Projected Texture $\eta_p$ \\
(1) Extract Surface Normal $x_{SN}$ and Surface World Coordinates $x_{SWC}$
\STATE Extract camera intrinsic matrix $\phi_{in}$ and extrinsic matrix $\phi_{ex}$ from $\phi$
\STATE Compute 3D Local Coordinates $x_{LC}$ from $\phi_{in}$ and $x_d$
\STATE $x_{SWC} \gets x_{LC} \times \phi_{ex}$
\STATE Compute $x_{SN}$ from $x_{SWC}$ using the cross product of two tangent vectors \\
(2) Extract Triplanar Mask
\STATE Compute Triplanar Mask $x_{TM}$ based on the highest absolute value of $x_{SN}$ for each axis
\STATE Compute $x_{TM\_x}$, $x_{TM\_y}$, $x_{TM\_z}$ masked for each axis \\
(3) Extract Projected Texture $\eta_p$
\STATE Compute the Repeated Surface Coordinates $x_{RSC}$ by modulating the $x_{SWC}$ with the repeated texture size based on $\phi_{rd}$
\STATE Take the UV indices of $x_{UV\_x}$, $x_{UV\_y}$, $x_{UV\_z}$ from each axis of Repeated Surface Coordinates $x_{RSC}$
% \STATE Compute the flattened UV indices $x_{UV\_x}$, $x_{UV\_y}$, $x_{UV\_z}$ for each axis from $x_{RSC}$
\STATE Assign the projected texture $\eta_{p\_x}$, $\eta_{p\_y}$, $\eta_{p\_z}$ for each axis from $\eta$ mapped with $x_{UV\_x}$, $x_{UV\_y}$, $x_{UV\_z}$
\STATE $\eta_{p\_x\_m} \gets \eta_{p\_x} \times x_{TM\_x}, \eta_{p\_y\_m} \gets \eta_{p\_y} \times x_{TM\_y}, \eta_{p\_z\_m} \gets \eta_{p\_z} \times x_{TM\_z}$
\STATE $\eta_p \gets \eta_{p\_x\_m}+\eta_{p\_y\_m}+\eta_{p\_z\_m}$
\end{algorithmic}
\end{algorithm}

\subsection{Framework Procedure}
This section describes the optimization of our adversarial texture using the ACTIVE framework. Drawing inspiration from the DTA framework \cite{Suryanto_2022_CVPR}, we employ the Neural Texture Renderer (NTR) for rendering our triplanar-mapped textures onto target vehicles while maintaining various physical properties. Prior to optimizing the adversarial texture, it is necessary to train the NTR in advance. A detailed explanation of the NTR can be found in Section \ref{sec:ntr}. Subsequently, during the ACTIVE training phase, we generate the optimal adversarial texture by minimizing $L_{total}$. To further evaluate the attack performance, we conduct a physical simulation (on UE4) wherein our adversarial texture is applied as a world-aligned texture to the target vehicle in UE4, as illustrated in the ACTIVE test phase in Figure 2 of our main paper.

\section{Implementation Details}

\subsection{Dataset}

\begin{figure}[H]
    \centering
    \begin{subfigure}[b]{\columnwidth}
        \centering
        \includegraphics[width=\columnwidth]{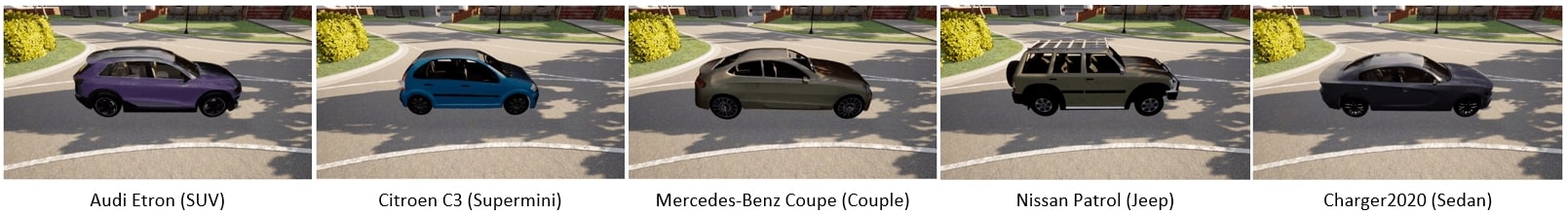}
        \caption{Multiple types of cars used for attack texture generation and robustness evaluation}
    \end{subfigure}
    \begin{subfigure}[b]{\columnwidth}
        \centering
        \includegraphics[width=\columnwidth]{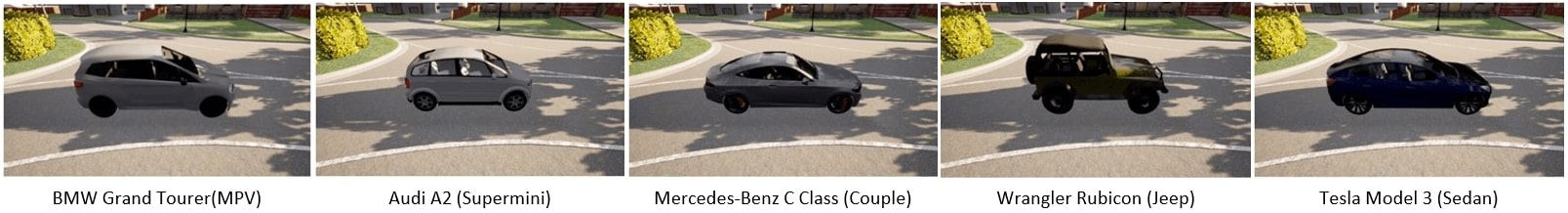}
        \caption{Multiple types of cars used for universality evaluation}
        \label{fig:car_universality}
    \end{subfigure}
    \caption{All cars used in our evaluation}
    \label{fig:dataset_cars}
\end{figure}

\textbf{Cars.}
We select multiple types of cars that are available on the CARLA \cite{carla} simulator.
Five types of cars used for attack texture generation and robustness evaluation are Audi Etron, Citreon C3, Mercedes-Benz Coupe, Nissan Patrol, and Charger 2020. Additionally, five other cars used for universality evaluation are BMW GrandTourer, Audi A2, Mercedes-Benz C-Class, Jeep Wrangler Rubicon, and Tesla Model 3. All cars are visualized in Fig. \ref{fig:dataset_cars}.

\textbf{For NTR Model Training.}
We select a map in CARLA and randomly choose 75 random locations for training and 50 different locations for testing. On each location, we generate a car dataset using a camera with a 5-meters distance, 15-degree pitch, and every 30-degree rotation. For each transformation, we gradually change the car types and textures with eight boundary colors and one gray color (as reference image) for training. Additionally, we select 50 random colors for testing. In summary, we employ 50,625 photo-realistic images for training (consisting 5 cars x 75 positions x 12 camera poses x 9 colors), and 150,000 images for testing (consisting 5 cars x 50 positions x 12 camera poses x 50 colors).

\textbf{For Attack Texture Generation.}
We use the same cars and camera poses, then choose 250 random locations on the same map to synthesize the reference images. In summary, we use 15,000 photo-realistic images as reference $x_{ref}$ (consisting 5 cars x 250 positions x 12 camera poses x 1 color). For each dataset, we also synthesize the car mask (i.e., the target object).

\textbf{For Attack Texture Evaluation.}
For robustness, we evaluate Audi Etron at 50 random locations, while for universality, we assess multiple cars at 20 random locations. We use wider camera transformation to evaluate attacks on untrained camera distributions as default. We also define multiple camera settings, including 
$[[5, 7.5], [7.5, 10], [10, 12.5], [12.5, 15]]$ meter distances,
$[[0, 15], [15, 30], [30, 45]]$ pitch degrees, and 
$[[0, 30], [30, 60], ..., [330, 360]]$ rotation degrees. 
Each value is selected randomly within the range for each evaluation sample. A total of 14,400 images (consisting 5 cars x 20 positions x 4 distances x 3 pitches x 12 angles) for a single texture evaluation are used in our experiments. 
% Fig. \ref{fig:camera_pose_transformations} \hl{illustrates the camera setting used by our evaluation.}

\subsection{Compared Methods}
We provide detailed parameter setup of compared methods: Dual Attention Suppression (DAS) \cite{Wang_2021_CVPR}, Full-coverage Camouflage Attack (FCA) \cite{fcaattack}, Universal Physical Camouflage (UPC) \cite{Huang_2020_CVPR} CAMOU \cite{Zhang_2019_ICLR}, Enlarge-and-Repeat (ER) \cite{Wu_CoRR_2020}, and Differentiable Transformation Attack (DTA) \cite{Suryanto_2022_CVPR} in our experiment. 
Except for texture taken from original codes, we re-optimize all methods targeting YOLOv3 \cite{redmon2018yolov3} on Audi Etron car, and use the same simulated town for a fair comparison.
Additionally, we include random and naive camouflage patterns to evaluate the model's robustness against arbitrary textures.

\textbf{DAS and FCA.} 
These methods use Neural Mesh Renderer (NMR) \cite{kato2018neural} to optimize the adversarial camouflage over a 3D car and attach it to the background, which is CARLA simulated town. We use the original texture from their official codes (DAS:{ \href{https://github.com/nlsde-safety-team/DualAttentionAttack}{https://github.com/nlsde-safety-team/DualAttentionAttack}, FCA:{ \href{https://github.com/idrl-lab/Full-coverage-camouflage-adversarial-attack}{https://github.com/idrl-lab/Full-coverage-camouflage-adversarial-attack}) for our evaluations.
Specifically, both textures are optimized for Audi Etron and CARLA Town10. We follow FCA for generating texture targeting YOLOv3 \cite{redmon2018yolov3} as the white box model. 
Since these methods depend on a specific 3D UV map, we only evaluate them on robustness evaluation.

\textbf{UPC.}
This method outputs a masked patch that can be attached to the target surface. Since UPC has different settings to reproduce in our environments and is specifically designed for targeting Faster R-CNN \cite{ren2015faster}, we execute their official code:{ \href{https://github.com/mesunhlf/UPC-tf}{https://github.com/mesunhlf/UPC-tf} and keep the original attack setup on cars, then use their attack patches in our environment. Following \cite{Suryanto_2022_CVPR}, we utilize decals in Unreal Engine to apply UPC patches onto target car surfaces, covering the hood, doors, rooftop, and back to ensure visibility from diverse perspectives. Since UPC has a different setting from ours, we only evaluate this method in transferability comparison.
Fig. \ref{fig:das_fca_upc_implementation} shows samples taken from related works' original papers (left) and how we fully transfer the texture on Unreal Engine 4 \cite{unrealengine} (right).

\begin{figure}[h]
    \centering
    \begin{subfigure}[b]{0.32\columnwidth}
        \centering
        \includegraphics[width=0.99\columnwidth]{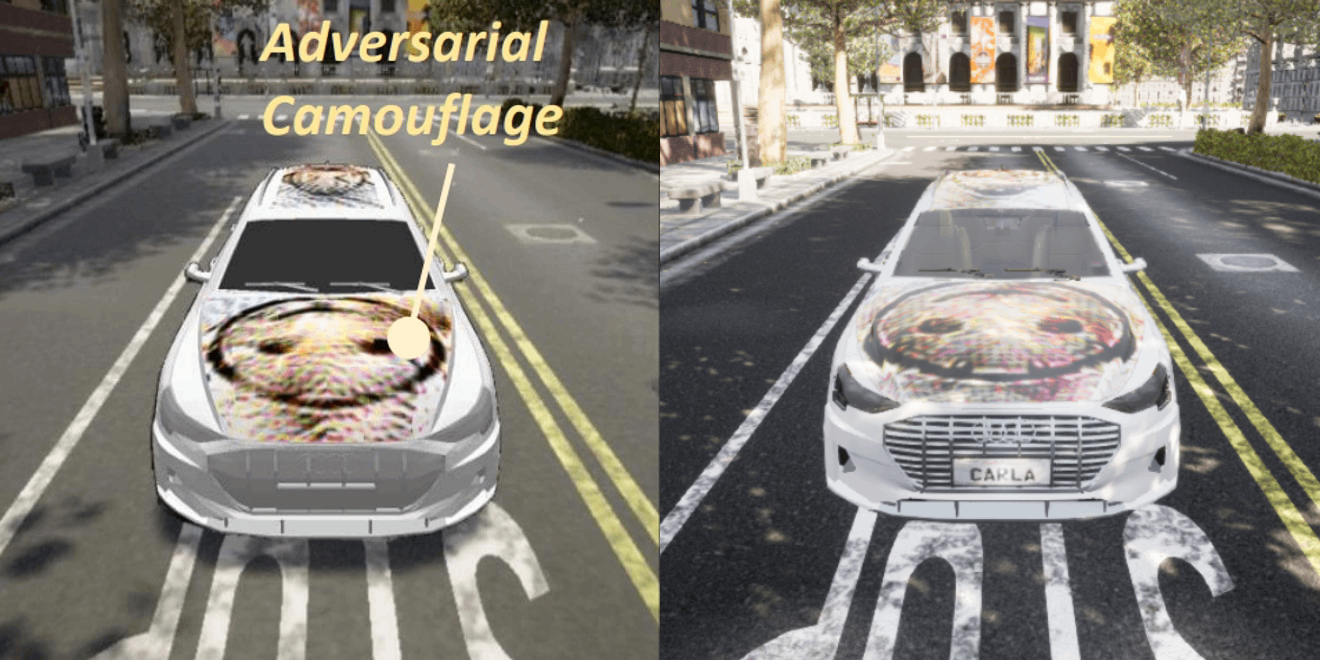}
        \caption{DAS \cite{Wang_2021_CVPR}}
    \end{subfigure}
    \begin{subfigure}[b]{0.32\columnwidth}
        \centering
        \includegraphics[width=0.99\columnwidth]{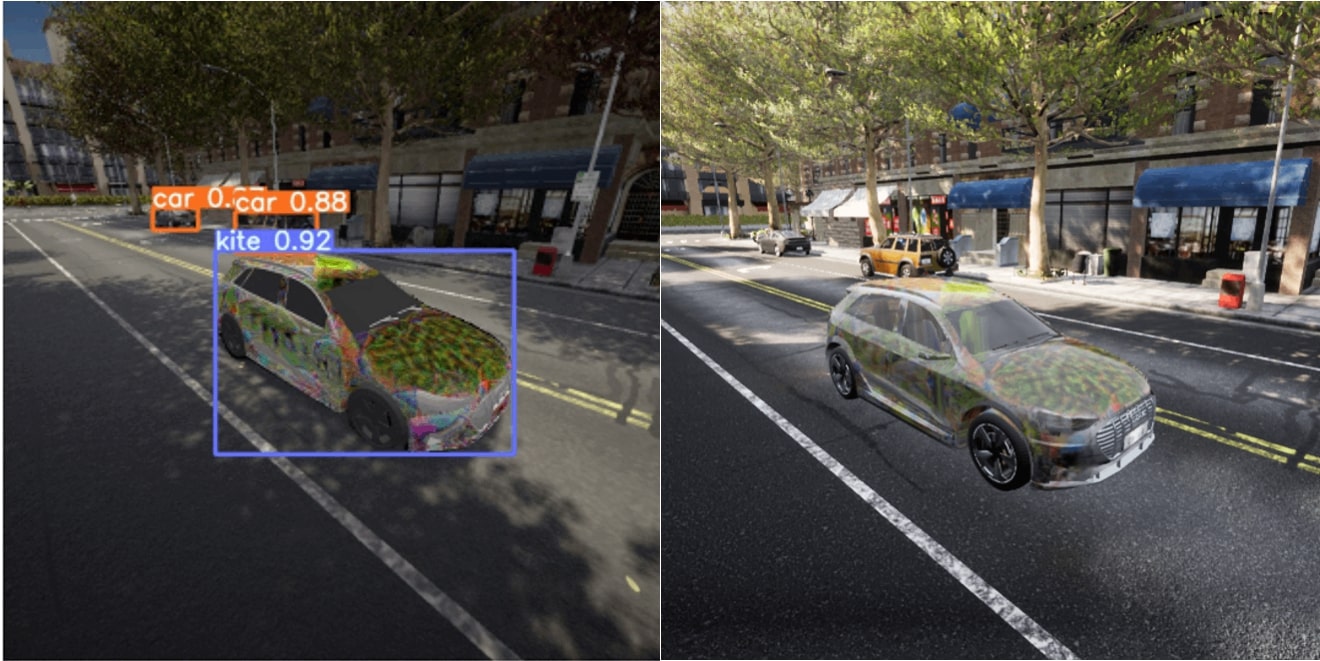}
        \caption{FCA \cite{fcaattack}}
    \end{subfigure}
     \begin{subfigure}[b]{0.32\columnwidth}
        \centering
        \includegraphics[width=0.99\columnwidth]{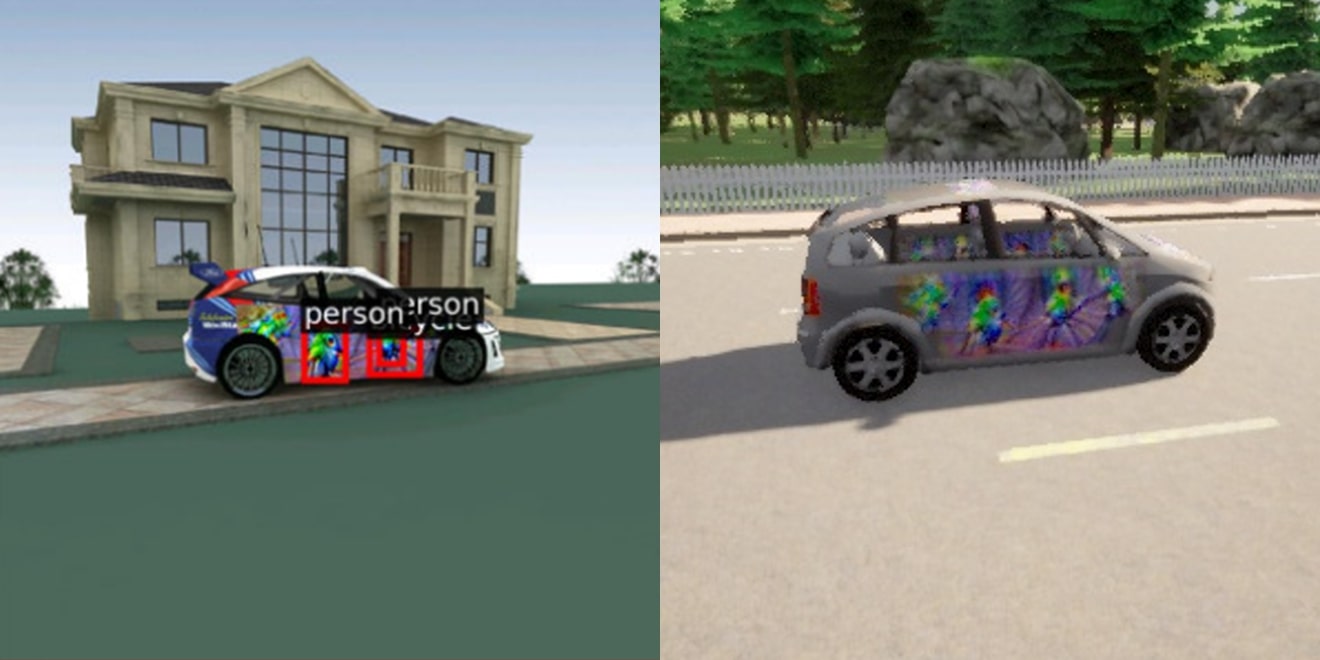}
        \caption{UPC \cite{fcaattack}}
    \end{subfigure}
    \caption{Transferring textures from the original paper (left) to our evaluation setting by fully rendering it on Unreal Engine 4 with epic settings (right).}
    \label{fig:das_fca_upc_implementation}
\end{figure}

\textbf{CAMOU, ER, and DTA.} 
These methods output repeated patterns that cover all target object surfaces as physical camouflage. We choose the same $16 \times 16$ texture resolution as the best original setting that forms mosaic-like patterns. 
We closely replicate the original papers' approach, but we rebuild the environment and target models 
%based on our same evaluation setup
same as our evaluation setup (optimizing texture on Audi Etron and targeting YOLOv3 \cite{redmon2018yolov3}). For CAMOU, we generate the camouflage pattern using the same clone network architecture and other parameters as the original paper. For ER, we use the same parameters as the original paper, except changing the $p = 3$ and $r = 1$ parameters to produce the same $16\times16$ texture. For DTA, we use NTR as neural renderer in our paper and optimize the texture using the original parameters. All generated textures and the rendered cars are shown in Fig. \ref{fig:repeated_pattern_implementation}.
% @Yongsu please check the setting for CAMOU and ER

\begin{figure}[h]
    \centering
    \begin{subfigure}[b]{0.32\columnwidth}
        \centering
        \includegraphics[width=0.99\columnwidth]{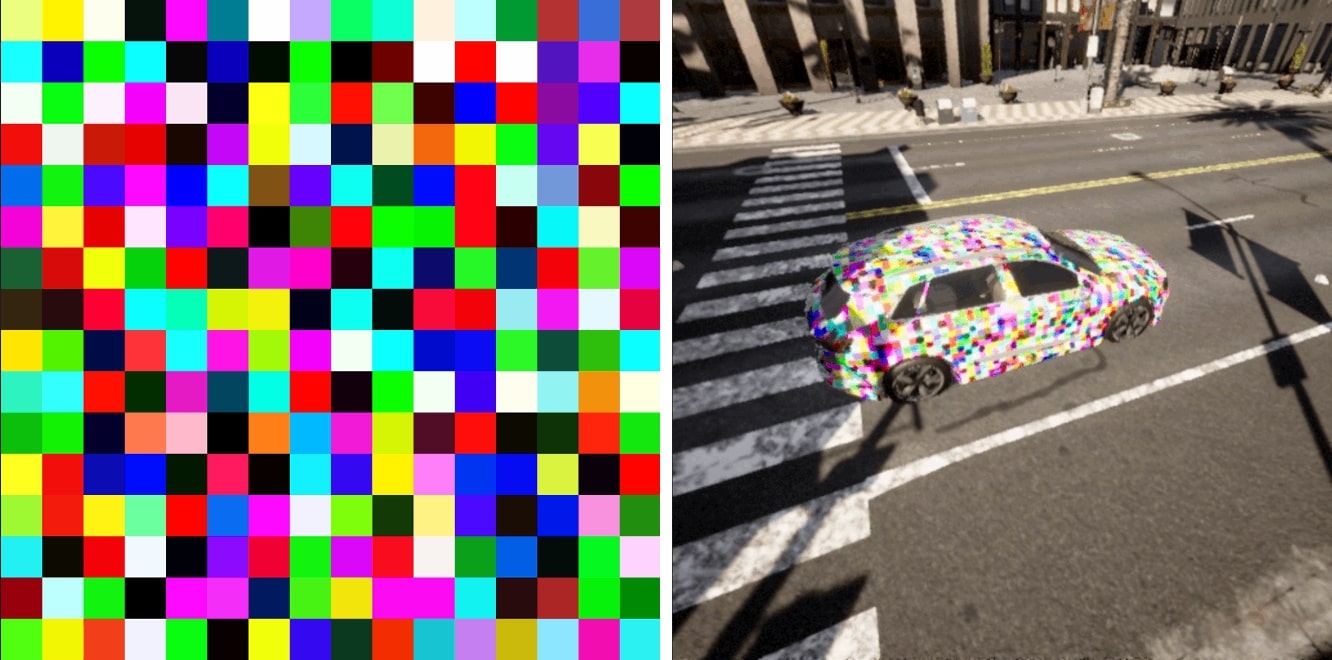}
        \caption{CAMOU \cite{Zhang_2019_ICLR}}
    \end{subfigure}
    \begin{subfigure}[b]{0.32\columnwidth}
        \centering
        \includegraphics[width=0.99\columnwidth]{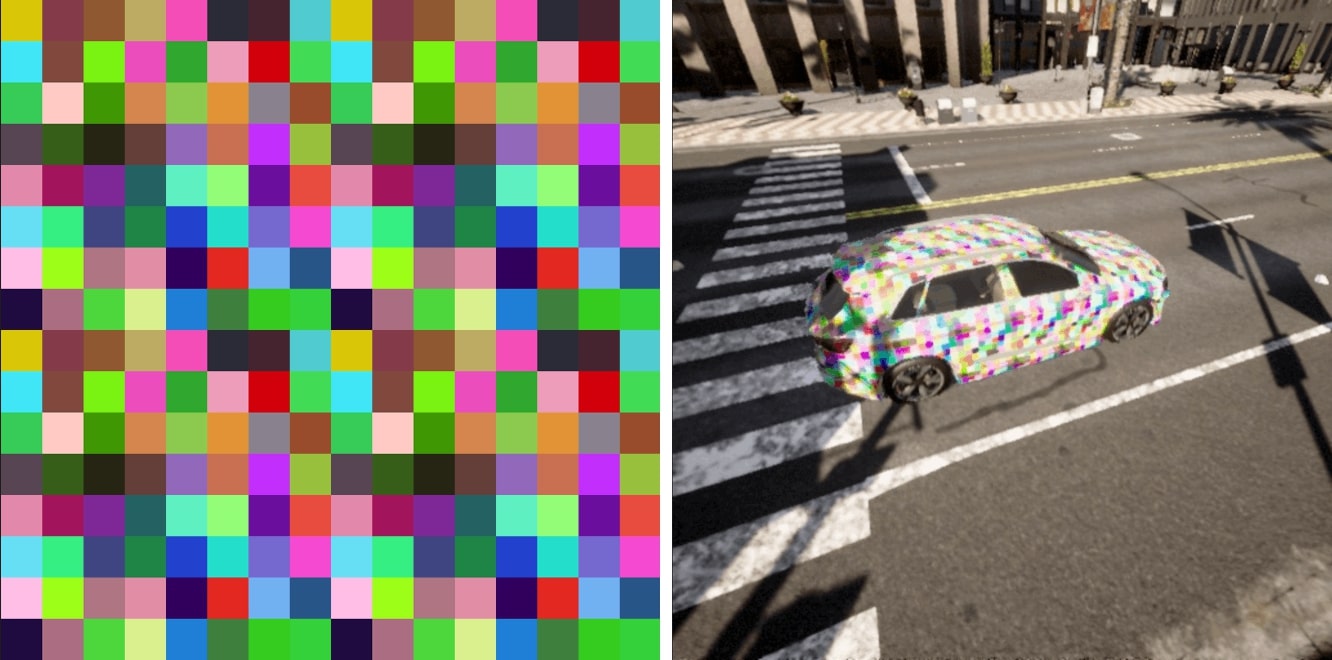}
        \caption{ER \cite{Wu_CoRR_2020}}
    \end{subfigure}
    \begin{subfigure}[b]{0.32\columnwidth}
        \centering
        \includegraphics[width=0.99\columnwidth]{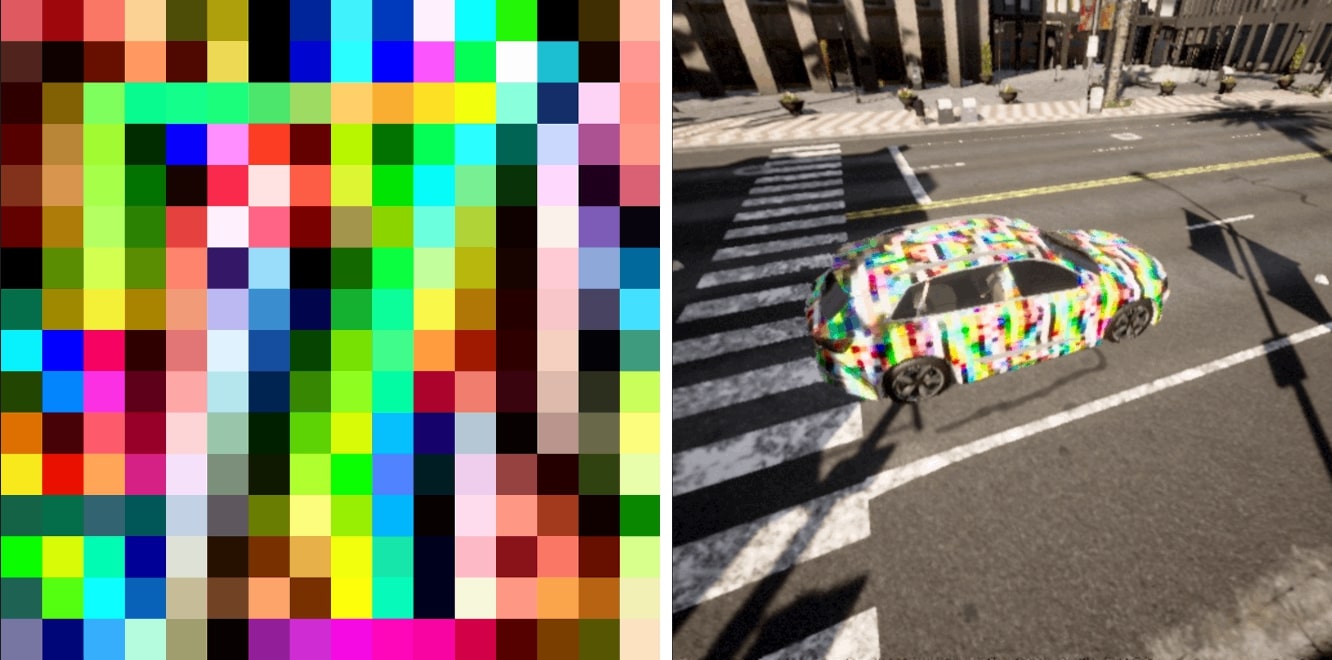}
        \caption{DTA \cite{Suryanto_2022_CVPR}}
    \end{subfigure}
    \caption{Re-implementation of previous works in our test environment. For each figure, the left side depicts the generated camouflage pattern, while the right side depicts the result of a car rendered by repeated texture using world-aligned texture.}
    \label{fig:repeated_pattern_implementation}
\end{figure}

\newpage

% To conclude, regarding Tables 2-5 in our main paper, each one was designed with \textit{specific criteria} in mind. In particular, Tab. 2 compares only adversarial camouflage methods based on \textit{Neural Renderer} (i.e., DAS, FCA, and DTA), aimed at assessing its suitability for digital-to-physical transferability. Next, Tab. 3 compares methods that can directly attack the \textit{same target model} (YOLOv3), with Random and Naive camouflage serving 
% as additional benchmarks for robustness against arbitrary texture. Subsequently, UPC was \textit{omitted} in Tab. 3 due to its loss being designed \textit{only for attacking Faster R-CNN}. In contrast, Tabs. 4 and 5 place UPC for comparing universality but \textit{exclude} DAS and FCA due to their \textit{object-dependent textures} cannot be universally applied to different cars.
In conclusion, the tables in our main paper were designed with \textit{specific criteria} in mind. In particular, Tab. 2 compares only adversarial camouflage methods based on \textit{Neural Renderer} (i.e., DAS, FCA, and DTA), aimed at assessing its suitability for digital-to-physical transferability. Next, Tab. 3 compares methods that can directly attack the \textit{same target model} (YOLOv3), \textit{vehicle}, and \textit{simulated town}, with Random and Naive camouflage serving as additional benchmarks for robustness against arbitrary texture. Subsequently, UPC was \textit{omitted} in Tab. 3 due to its loss being designed \textit{only for attacking Faster R-CNN}. In contrast, Tabs. 4 and 5 place UPC for comparing universality but \textit{exclude} DAS and FCA due to their \textit{object-dependent textures} cannot be universally applied to different cars.

\subsection{Evaluated Models}
We evaluate our method on multiple publicly-available state-of-the-art COCO \cite{coco} pre-trained object detection models including EfficientDet-D2 \cite{tan2020efficientdet}, YOLOv3 \cite{redmon2018yolov3}, SSD \cite{liu2016ssd}, Faster R-CNN \cite{ren2015faster}, Mask R-CNN \cite{he2017mask}, YOLOv7 \cite{wang2022yolov7}, Dynamic R-CNN \cite{zhang2020dynamic}, Sparse R-CNN \cite{sun2021sparse}, Deformable DETR \cite{zhu2021deformable}, Pyramid Vision Transformer \cite{wang2021pyramid}, MobileNetV2 \cite{sandler2018mobilenetv2}, and YOLOX-L \cite{ge2021yolox}, with the default score threshold set to 0.30.
Furthermore, we also evaluate the attack patterns on state-of-the-art pre-trained panoptic segmentation model including Max-DeepLab-L \cite{wang2021max} and Axial-DeepLab \cite{wang2020axial} with SWideRNet \cite{swidernet_2020} Backbone either pretrained on COCO \cite{coco} or Cityscape \cite{Cordts2016Cityscapes} dataset. For the object detection task, we evaluate the target car AP@0.5, while for the segmentation task, we evaluate the target car pixel accuracy. The details of all evaluated models, including model variant, publication venue, and implementation framework, are available in Table \ref{tab:all_evaluated_object_detection_models} for object detection models and Table \ref{tab:all_evaluated_segmentation_models} for segmentation models.

\begin{table}[h]
\caption{All COCO pre-trained object detection models used in our evaluations. Models are sorted by publication year.}
\label{tab:all_evaluated_object_detection_models}
\resizebox{\columnwidth}{!}{
\begin{tabular}{|c|c|c|c|c|c|c|}
\hline
\rowcolor[HTML]{FFFC9E}
\textbf{Model Name} & \textbf{Model Variant} & \textbf{Venue}             & \textbf{Year}             & \textbf{ML Framework} & \textbf{Implementation} & \textbf{Link} \\ \hline
Faster R-CNN               & ResNet50           & NPIS  & 2015 & TensorFlow   & TF Object Detection API \cite{tfod} & \href{https://github.com/tensorflow/models/blob/master/research/object_detection/g3doc/tf2_detection_zoo.md}{Here} \\ \hline
SSD                        & ResNet50-FPN       & ECCV  & 2016 & TensorFlow   & TF Object Detection API \cite{tfod} & \href{https://github.com/tensorflow/models/blob/master/research/object_detection/g3doc/tf2_detection_zoo.md}{Here} \\ \hline
Mask R-CNN                 & Inception ResNetV2 & ICCV  & 2017 & TensorFlow   & TF Object Detection API \cite{tfod} & \href{https://github.com/tensorflow/models/blob/master/research/object_detection/g3doc/tf2_detection_zoo.md}{Here} \\ \hline
YOLOv3                     & YOLOv3             & ArXiv & 2018 & TensorFlow   & TF Re-implementation    & \href{https://github.com/hunglc007/tensorflow-yolov4-tflite}{Here} \\ \hline
MobileNetv2                & FPN Lite           & CVPR  & 2018 & TensorFlow   & TF Object Detection API \cite{tfod} & \href{https://github.com/tensorflow/models/blob/master/research/object_detection/g3doc/tf2_detection_zoo.md}{Here} \\ \hline
EfficientDet               & EfficientDet-D2    & CVPR  & 2020 & TensorFlow   & TF Object Detection API \cite{tfod} & \href{https://github.com/tensorflow/models/blob/master/research/object_detection/g3doc/tf2_detection_zoo.md}{Here} \\ \hline
Dynamic R-CNN              & ResNet50           & ECCV  & 2020 & Pytorch      & MMDetection \cite{mmdetection}            & \href{https://github.com/open-mmlab/mmdetection/tree/master/configs/dynamic_rcnn}{Here} \\ \hline
Pyramid Vision Transformer & PVT-S              & ICCV  & 2021 & Pytorch      & MMDetection \cite{mmdetection}            & \href{https://github.com/open-mmlab/mmdetection/tree/master/configs/pvt}{Here} \\ \hline
Deformable DETR            & Two-Stage DDTR     & ICLR  & 2021 & Pytorch      & MMDetection \cite{mmdetection}             & \href{https://github.com/open-mmlab/mmdetection/tree/master/configs/deformable_detr}{Here} \\ \hline
Sparse R-CNN               & ResNet50-FPN       & CVPR  & 2021 & Pytorch      & MMDetection \cite{mmdetection}             & \href{https://github.com/open-mmlab/mmdetection/tree/master/configs/sparse_rcnn}{Here} \\ \hline
YOLOX                      & YOLOX-L            & ArXiv & 2021 & TensorFlow   & TF Re-implementation    & \href{https://github.com/leondgarse/keras_cv_attention_models/tree/main/keras_cv_attention_models/yolox}{Here} \\ \hline
YOLOv7                     & YOLOv7             & ArXiv & 2022 & Pytorch      & Official Implementation & \href{https://github.com/WongKinYiu/yolov7}{Here} \\ \hline
\end{tabular}
}
\end{table}

\vspace{-0.5mm}

\begin{table}[h]
\caption{All pre-trained panoptic segmentation models used in our evaluations. Models are sorted by publication year.}
\label{tab:all_evaluated_segmentation_models}
\resizebox{\columnwidth}{!}{
\begin{tabular}{|c|c|c|c|c|c|c|}
\hline
\rowcolor[HTML]{FFFC9E} 
\textbf{Model Name} & \textbf{Model Variant} & \textbf{Venue} & \textbf{Year} & \textbf{Pre-trained Dataset} & \textbf{Implementation} & \textbf{Link} \\ \hline
Axial-DeepLab & Axial-SWideRNet-(1, 1, 4.5) & ECCV & 2020 & Cityscape & TF DeepLab2 & \href{https://github.com/google-research/deeplab2/blob/main/g3doc/projects/axial_deeplab.md}{Here} \\ \hline
MaX-DeepLab   & MaX-DeepLab-L               & CVPR & 2021 & Cityscape & TF DeepLab2 & \href{https://github.com/google-research/deeplab2/blob/main/g3doc/projects/axial_deeplab.md}{Here}  \\ \hline
MaX-DeepLab   & MaX-DeepLab-L               & CVPR & 2021 & COCO      & TF DeepLab2 & \href{https://github.com/google-research/deeplab2/blob/main/g3doc/projects/max_deeplab.md}{Here} \\ \hline
\end{tabular}
}
\end{table}

\section{Neural Texture Renderer (NTR)}
\label{sec:ntr}
\subsection{Improvement from Differentiable Transformation Network (DTN)}
In this section, we present Neural Texture Renderer (NTR), the improvement of DTN by Suryanto et al. \cite{Suryanto_2022_CVPR}.
In particular, we found several redundancies in the DTN, especially in the Transformation Features \textit{TF} and training process. Here, we describe per point how we managed to improve DTN
%, \hl{in which we refer to as NTR}.
in our NTR. %%HAR: atau mau lebih soft, here referred to as  NTR?

\textbf{Redundancy in Transformation Features.} DTN as in \cite{Suryanto_2022_CVPR} uses four \textit{TF}, including \textit{TF Subtractor} + ReLU, \textit{TF Adder}, \textit{TF Multiplier}, and \textit{TF Final Adder} that are assumed to cover all basic transformations. We theorize that the first two layers are redundant; the last two are already sufficient for encoding the \textit{TF}. Therefore, we only use \textit{TF Multiplier} and \textit{TF Final Adder} in our final Transformation Features \textit{TF} that are similar to weight and bias in Neural Network design. Our NTR architecture using the dense connection is depicted in Fig. \ref{fig:dtn_improvement_architecture}.

\textbf{Redundancy in Dataset.} DTN uses 50 flat random color textures for training the network. We believe this is inefficient, as we want to train the network with multiple objects. We hypothesize that training the network with eight boundary colors in RGB space and gray $[128, 128, 128]$ as the reference color is enough to make the network generalize other colors. The eight boundary colors consist of the primary colors (red $[255, 0, 0]$, green $[0, 255, 0]$, blue $[0, 0, 255]$), secondary colors (magenta $[255, 0, 255]$,  yellow $[255, 255, 0]$, cyan $[0, 255, 255]$), white $[255, 255, 255]$, and black $[0, 0, 0]$, as in Fig. \ref{fig:dtn_improvement_boundary}. 

\begin{figure}[h]
    \centering
    \begin{subfigure}[b]{0.6\columnwidth}
        \centering
        \includegraphics[width=0.9\columnwidth]{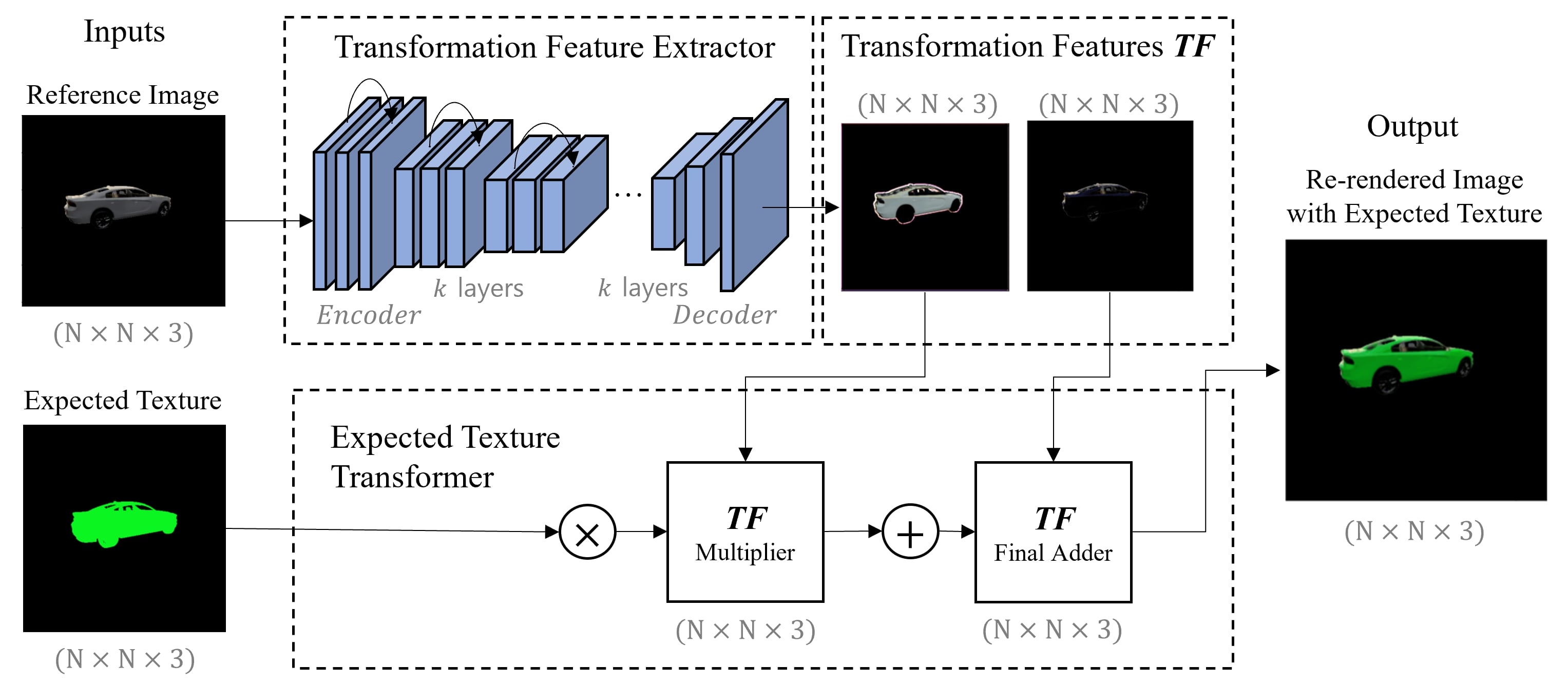}
        \caption{Our Implemented NTR Architecture}
        \label{fig:dtn_improvement_architecture}
    \end{subfigure}
    \begin{subfigure}[b]{0.39\columnwidth}
        \centering
        \includegraphics[width=0.6\columnwidth]{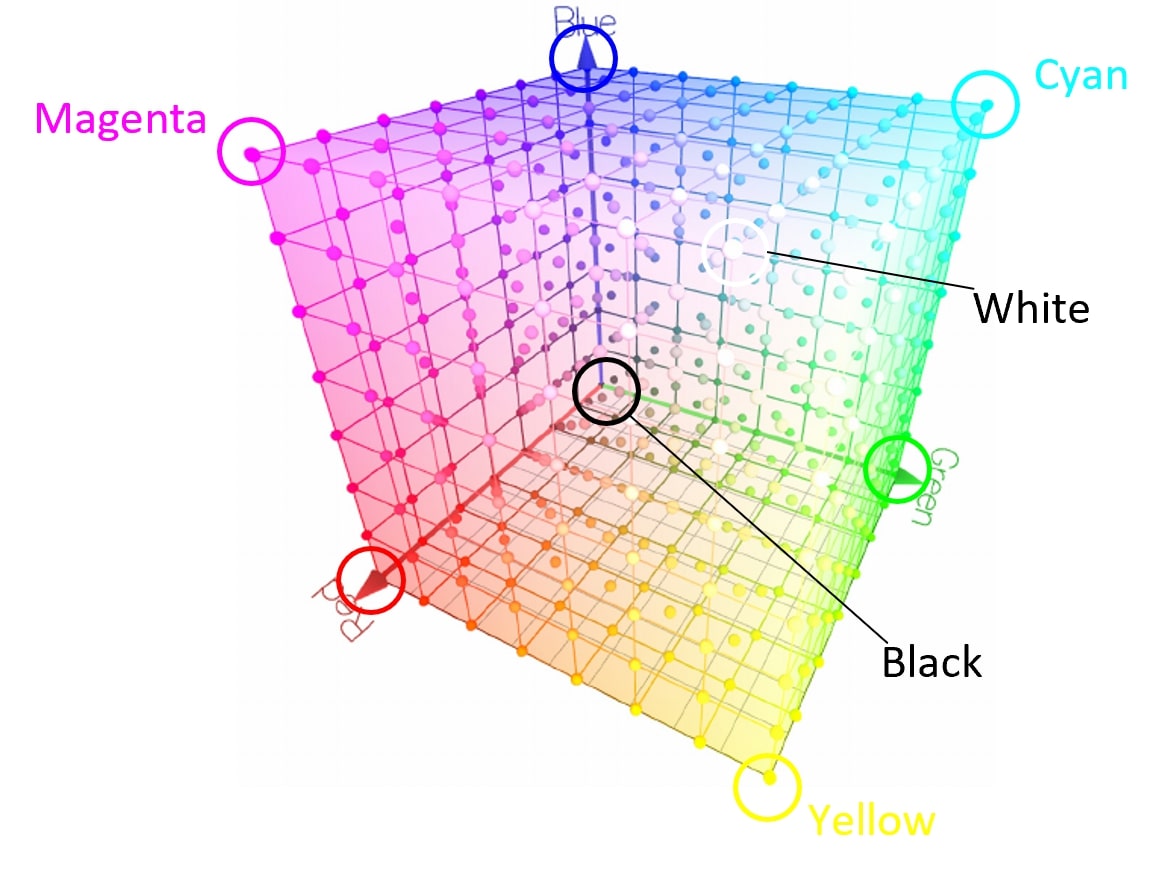}
        \caption{Selected Boundary Colors in RGB Space}
        \label{fig:dtn_improvement_boundary}
    \end{subfigure}
    % \caption{Improvement in DTN (Differentiable Transformation Network)}
    \vspace{-1mm}
    \caption{Neural Texture Renderer (NTR) Architecture and Improvement from DTN}
    \label{fig:dtn_improvement}
\end{figure}

\vspace{-4mm}

\subsection{NTR Evaluation}
\vspace{-2mm}
% We then perform experiments to evaluate our hypotheses about redundancy in the original DTN design. We train several DTN models using the same machine, and each model is trained with 32 batch sizes for 20 epochs using a single Nvidia Tesla V100 GPU and Intel Xeon E5-2696 CPU. Evaluation results are as shown in Tab. \ref{tab:dtn_evaluations}. From the table, we can infer that training DTN with only eight boundary colors and gray is enough to make DTN able to generalize the test set, resulting in 82\% efficiency. Furthermore, removing redundant \textit{TF} does not hurt the performance, showing that the last two \textit{TF} are enough to encode the expected transformation features.
We then perform experiments to evaluate our hypotheses about redundancy in the DTN design. In particular, we train several models using the same machine, and each model is trained with 32 batch sizes for 20 epochs using a single Nvidia Tesla V100 GPU and Intel Xeon E5-2696 CPU. Evaluation results are as shown in Tab. \ref{tab:dtn_evaluations}, with DTN as the baseline. From the table, we can infer that training the model with only eight boundary colors and gray is enough to enable generalizing the test set, resulting in 82\% efficiency. Furthermore, removing redundant \textit{TF} does not hurt the performance, showing that the last two \textit{TF} are enough to encode the expected transformation features. 

\vspace{-2mm}

\begin{table}[h]
\caption{NTR Evaluation}
\vspace{-2mm}
\label{tab:dtn_evaluations}
\resizebox{\columnwidth}{!}{
\begin{tabular}{|l|r|r|r|r|r|}
\hline
\rowcolor[HTML]{FFFC9E} 
\multicolumn{1}{|c|}{\cellcolor[HTML]{FFFC9E}Model} & N Colors & Total Dataset & Training Time & Tess Loss & Test SSIM \\ \hline
DTN (/w dense connection, k\_layers=4)     & 50       & 281,250       & 20.4 hours    & 0.0312    & 0.986     \\ \hline
+ Trained with Boundary Colors   & 9 & 50625 & 3.7 hours & 0.0314 & 0.985 \\ \hline
++ Removed Redundant \textit{TF} & 9 & 50625 & 3.7 hours & 0.0314 & 0.985 \\ \hline
\end{tabular}
}
\end{table}

\vspace{-2mm}

\subsection{NTR Sample Predictions}
\vspace{-2mm}
    Here, we present the sample prediction of NTR on the test set after the training is completed. Fig. \ref{fig:dtn_prediction_sample} shows sample prediction result consisting of the reference image containing the target object, expected texture, predicted image (i.e., the output of NTR), ground truth, and MSE (i.e., the error between the predicted image and ground truth). As shown, NTR can learn how to render the expected texture given the reference image. Fig. \ref{fig:dtn_transformation_process} shows the process of transforming expected texture using the encoded \textit{TF}. We can observe that \textit{TF} multiplier is enough to encode the color transformation with the expected texture. \textit{TF} multiplier can encode how the texture is minimized or emphasized and remove the non-applied texture part. The \textit{TF} final adder enhances the image from the \textit{TF} multiplier and adds the non-applied texture part from the reference.

\vspace{-1mm}

\begin{figure}[h]
    \centering
    \begin{subfigure}[b]{0.425\columnwidth}
        \centering
        \includegraphics[width=\columnwidth]{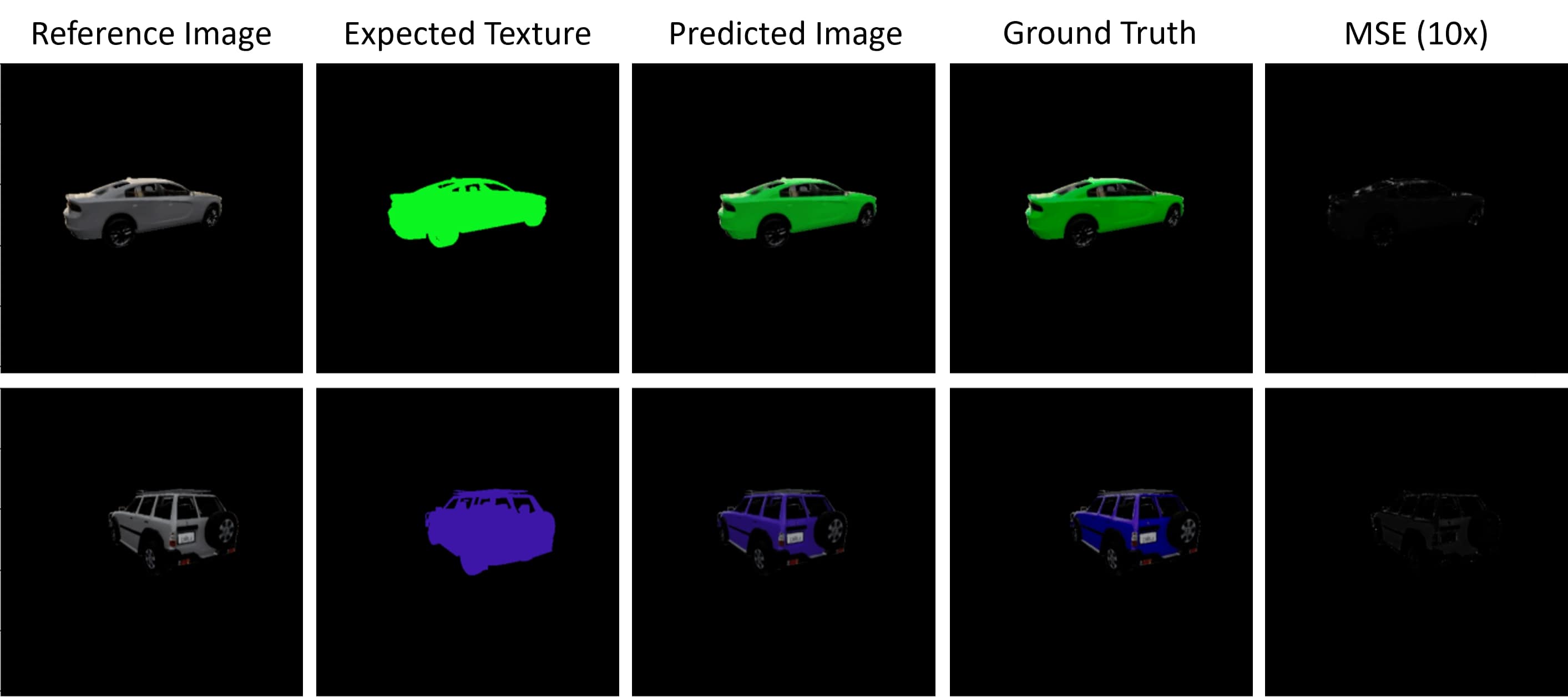}
        \caption{NTR sample predictions on test set}
        \label{fig:dtn_prediction_sample}
    \end{subfigure}
    \begin{subfigure}[b]{0.435\columnwidth}
        \centering
        \includegraphics[width=\columnwidth]{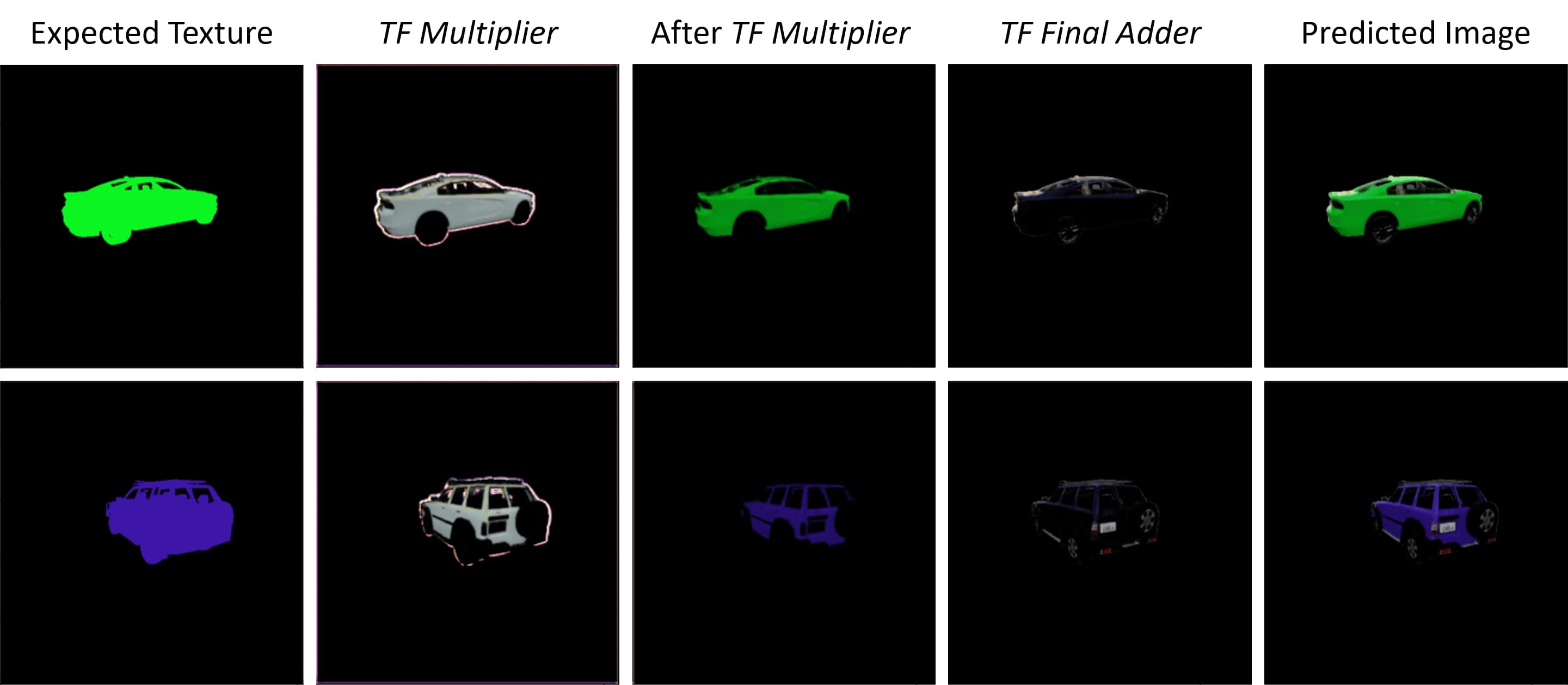}
        \caption{Transforming expected texture with the encoded \textit{TF}}
        \label{fig:dtn_transformation_process}
    \end{subfigure}
    \vspace{-2mm}
    \caption{NTR Sample Predictions}
    \label{fig:dtn_sample}
\end{figure}

\newpage

\section{Robustness Evaluation Details}
% TODO: Add paragraphs about what we are doing here.
% We provide the detailed experimental results of the robustness evaluation. 
\subsection{Detailed Performance on Various Camera Poses}
% TODO: Add paragraphs about the graphs
% TODO: Add paragraphs about the sample images
We evaluate YOLOv3 \cite{redmon2018yolov3} (a white-box model), and SSD \cite{liu2016ssd}, Faster R-CNN \cite{ren2015faster}, Mask R-CNN \cite{he2017mask} (which are black-box models) for each distance, rotation, and pitch. As shown in Fig. \ref{fig:smooth_and_camouflage_textures} and \ref{fig:comparison_sample_predictions}, our method has better attack performance than other methods and is relatively stable under various camera poses in most cases.

\begin{figure}[h]
    \centering
    \begin{subfigure}[b]{0.49\columnwidth}
        \centering
        \includegraphics[width=0.90\columnwidth]{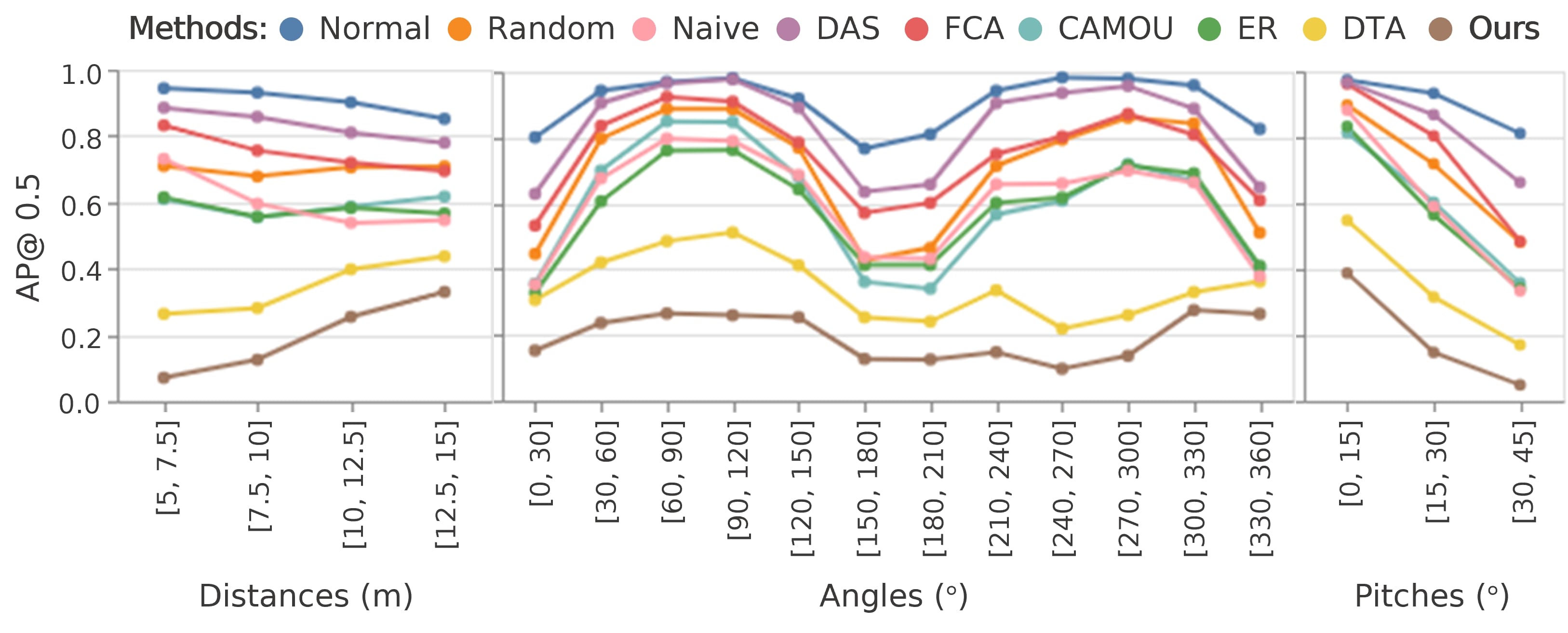}
        \caption{YOLOv3 \cite{redmon2018yolov3}}
        \label{fig:comparison_yolov3}
    \end{subfigure}
    \begin{subfigure}[b]{0.49\columnwidth}
        \centering
        \includegraphics[width=0.90\columnwidth]{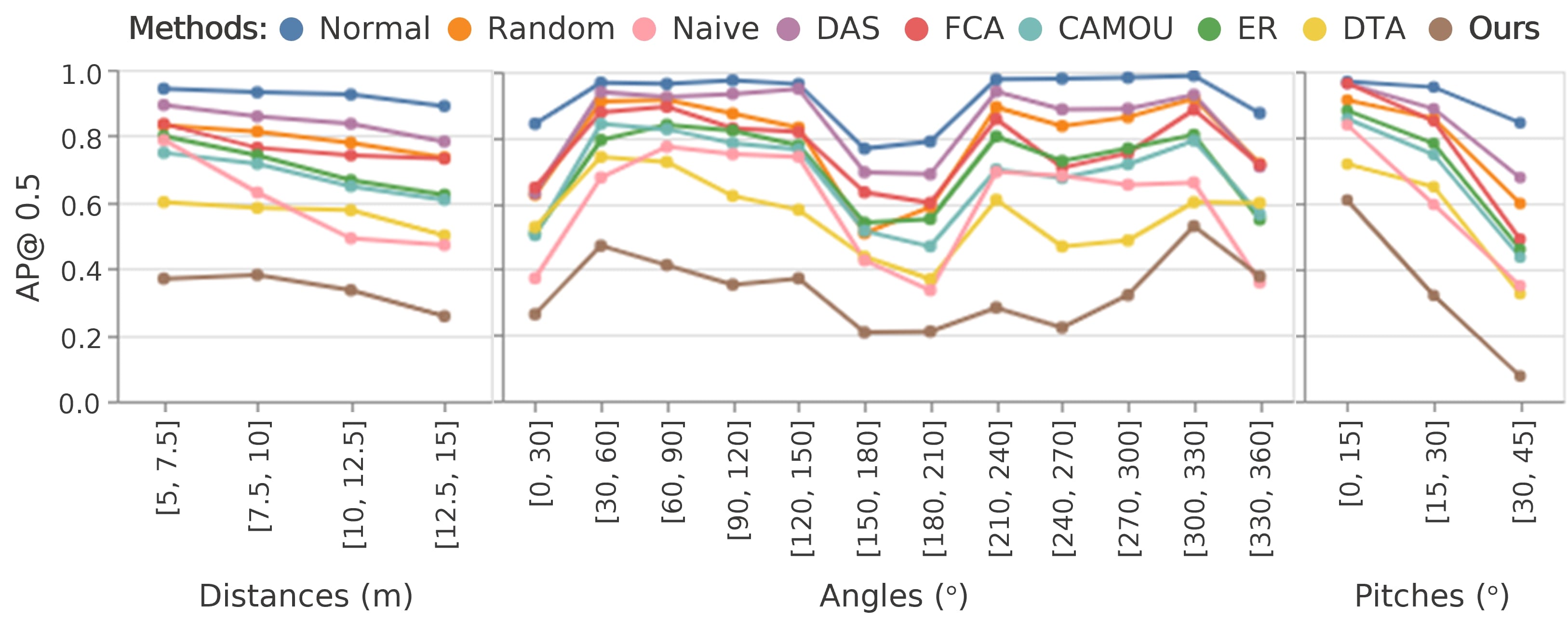}
        \caption{SSD \cite{liu2016ssd}}
        \label{fig:comparison_ssd}
    \end{subfigure}
    \begin{subfigure}[b]{0.49\columnwidth}
        \centering
        \includegraphics[width=0.90\columnwidth]{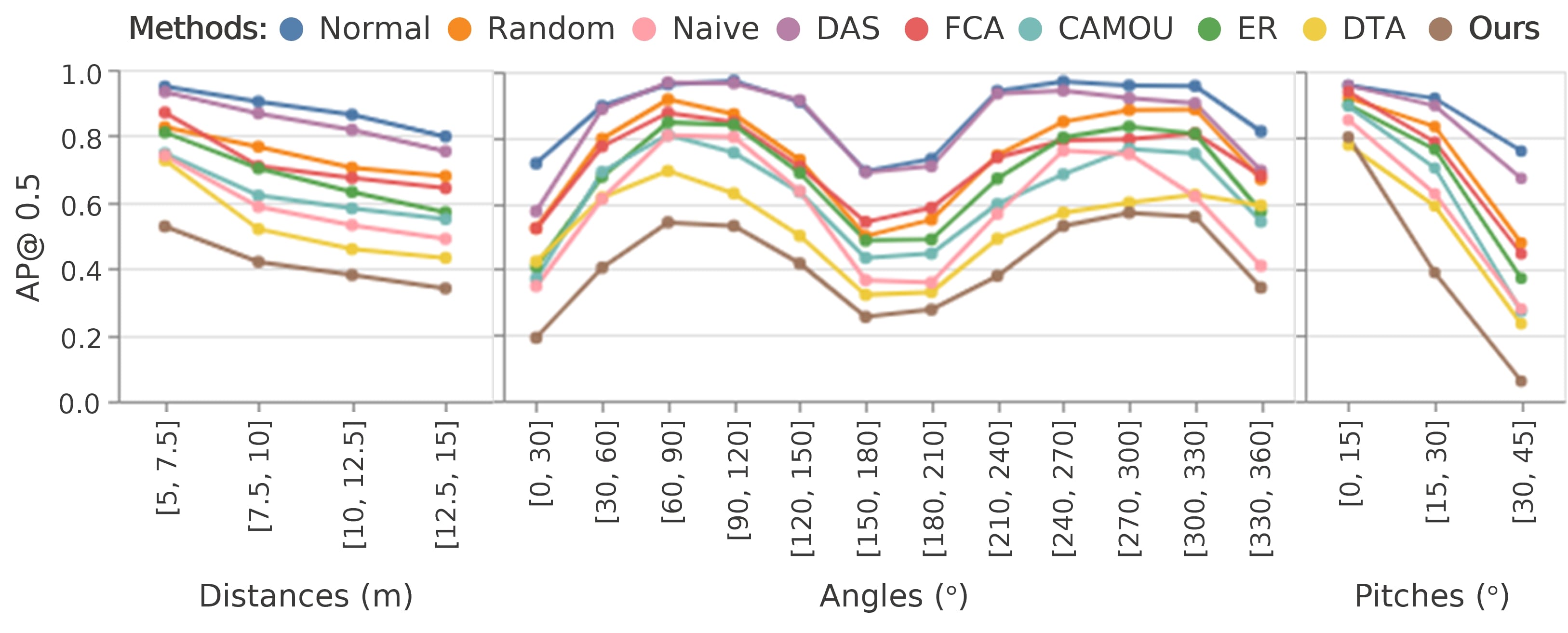}
        \caption{Faster R-CNN \cite{ren2015faster}}
        \label{fig:comparison_fasterrcnn}
    \end{subfigure}
    \begin{subfigure}[b]{0.49\columnwidth}
        \centering
        \includegraphics[width=0.90\columnwidth]{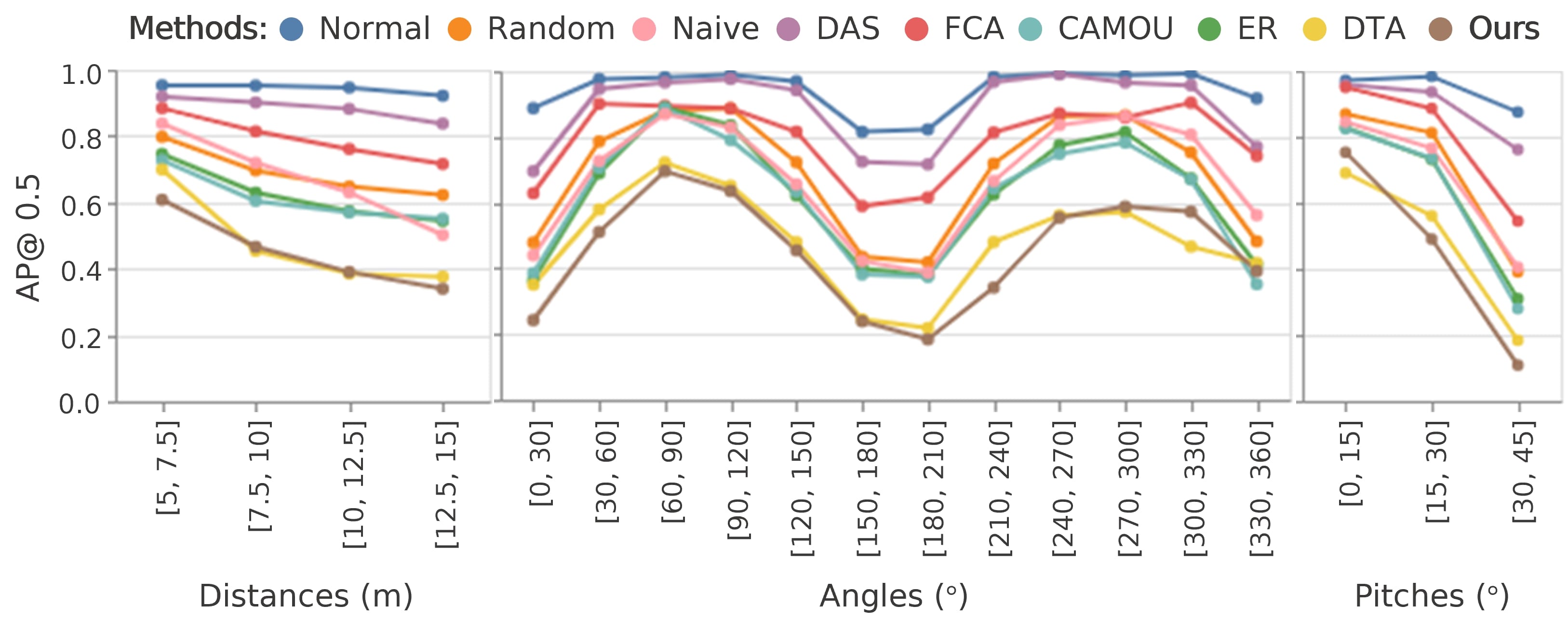}
        \caption{Mask R-CNN \cite{he2017mask}}
        \label{fig:comparison_maskrcnn}
    \end{subfigure}
    \caption{Attack comparison on different camera poses. Values are Average Precision @0.5 of the target car.}
    \label{fig:smooth_and_camouflage_textures}
\end{figure}

\begin{figure}[h]
\centerline{\includegraphics[width=\columnwidth]{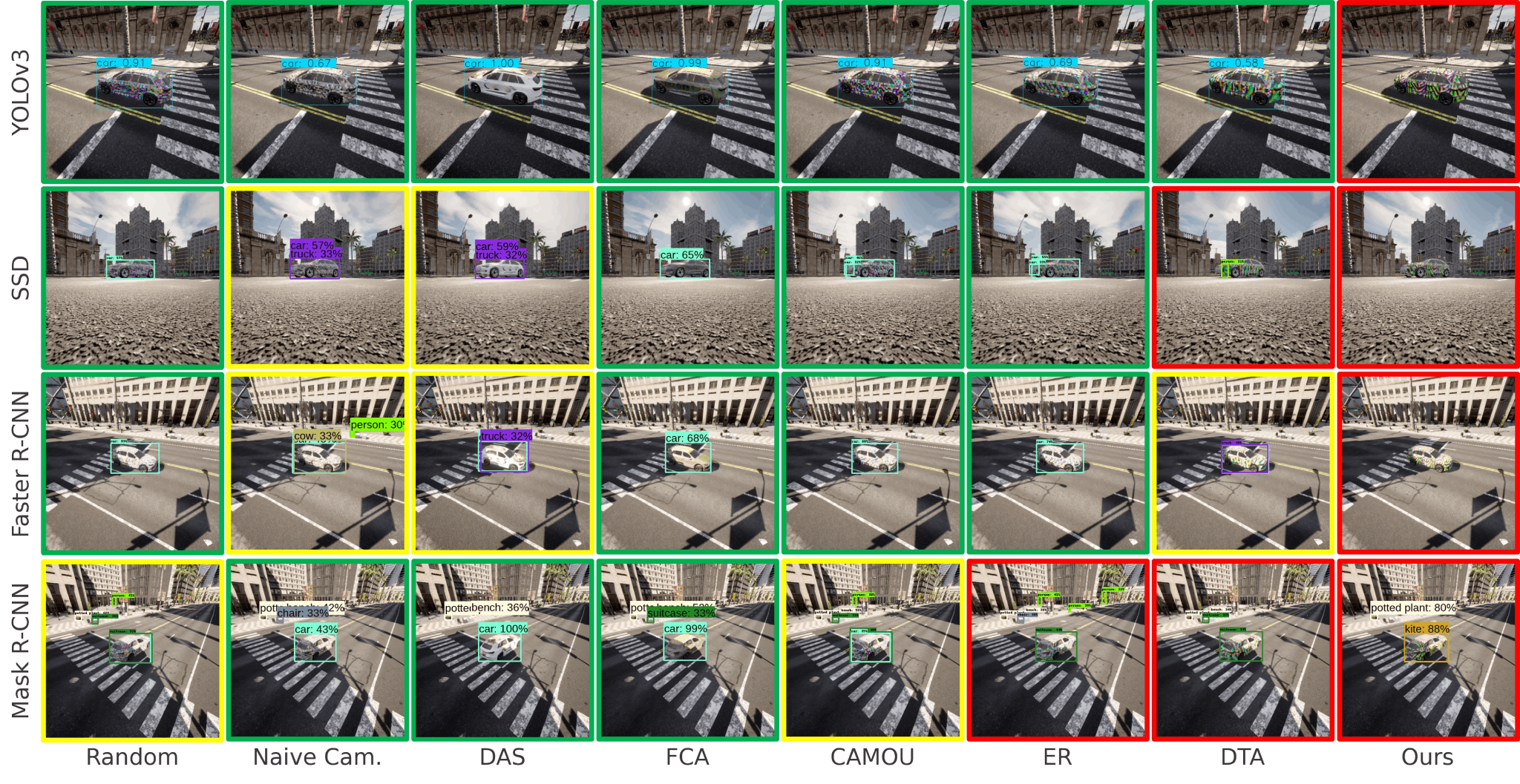}}
\caption{Sample predictions for each method and evaluated model. Green = correct; Yellow = partially correct; Red = misdetection. Zoom for detail.}
\label{fig:comparison_sample_predictions}
\end{figure}

\section{Universality Evaluation Details}
% TODO: Add paragraphs about what we are doing here.
% We provide the detailed experimental results of the universality evaluation. First, we evaluate the transferability by using the different settings with the robustness evaluation. We also evaluate the transferability on a different task (e.g., segmentation models). Finally, we fabricate car models with a 3D printer to evaluate the transferability to the real world.
We provide the detailed experimental results of the universality evaluation. First, we evaluate the transferability by trying out our method in different settings (i.e., different from the settings used in the robustness evaluation). We also evaluate the transferability on different classes (i.e., truck and bus) and a different task (i.e., segmentation models). Finally, we fabricate the car models with a 3D printer to evaluate our method's transferability to the real world.

\subsection{Transferability to Different Settings}
% TODO: Add paragraphs about what we are doing here.
% TODO: Add paragraphs about the sample images
Here, we use different settings from the robustness evaluation: different cars (as in Fig. \ref{fig:car_universality}), different scenes, and diverse modern object detection architecture such as YOLOv7 \cite{wang2022yolov7}, Dynamic R-CNN \cite{zhang2020dynamic}, Sparse R-CNN \cite{sun2021sparse}, Deformable DETR (DDTR) \cite{zhu2021deformable}, and Pyramid Vision Transformer (PVT) \cite{wang2021pyramid}. As shown in Fig. \ref{fig:transferability_evaluations}, it is observed that our method outperforms previous works in most cases, and even when being evaluated in different settings as a black-box attack. Sample prediction for each model can be seen in Fig. \ref{fig:transferability_samples}.

\begin{figure}[h]
    \centering
    \begin{subfigure}[b]{0.49\columnwidth}
        \centering
        \includegraphics[width=0.95\columnwidth]{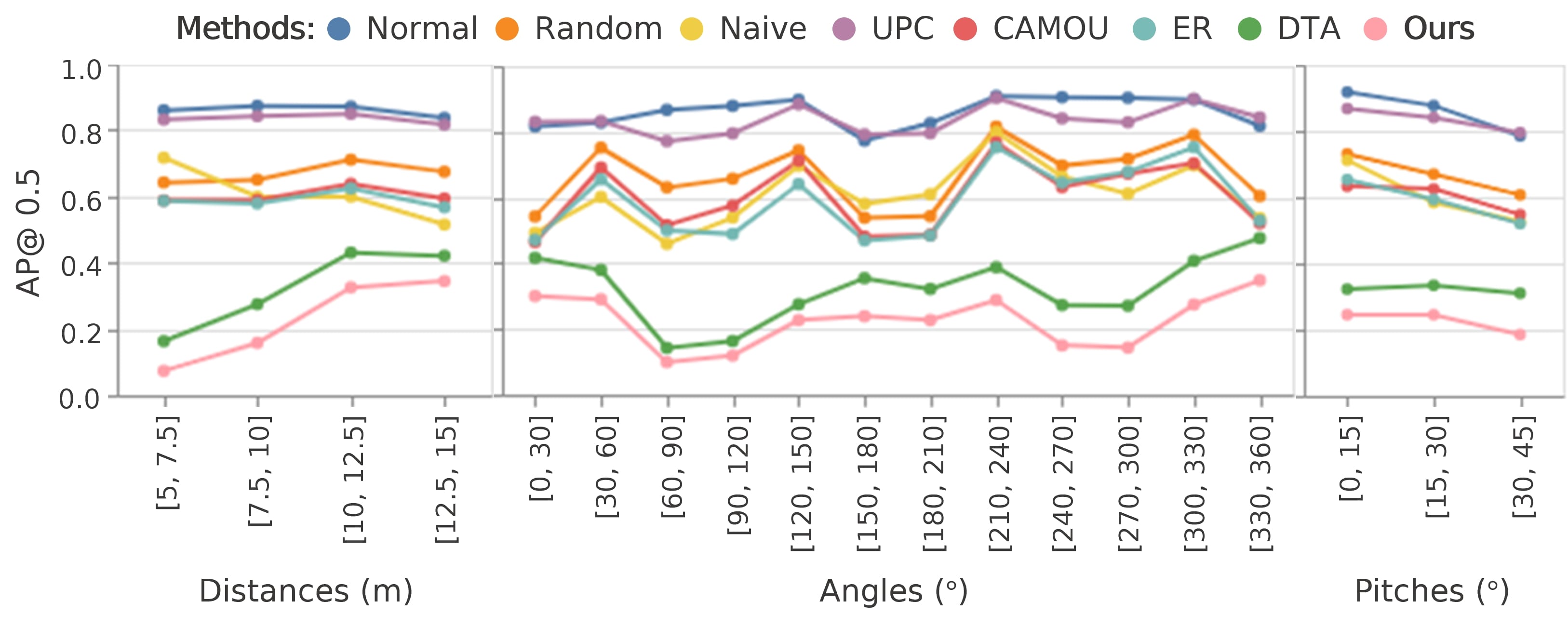}
        \caption{YOLOv3 \cite{redmon2018yolov3}}
        \label{fig:transferability_yolov3}
    \end{subfigure}
    \begin{subfigure}[b]{0.49\columnwidth}
        \centering
        \includegraphics[width=0.95\columnwidth]{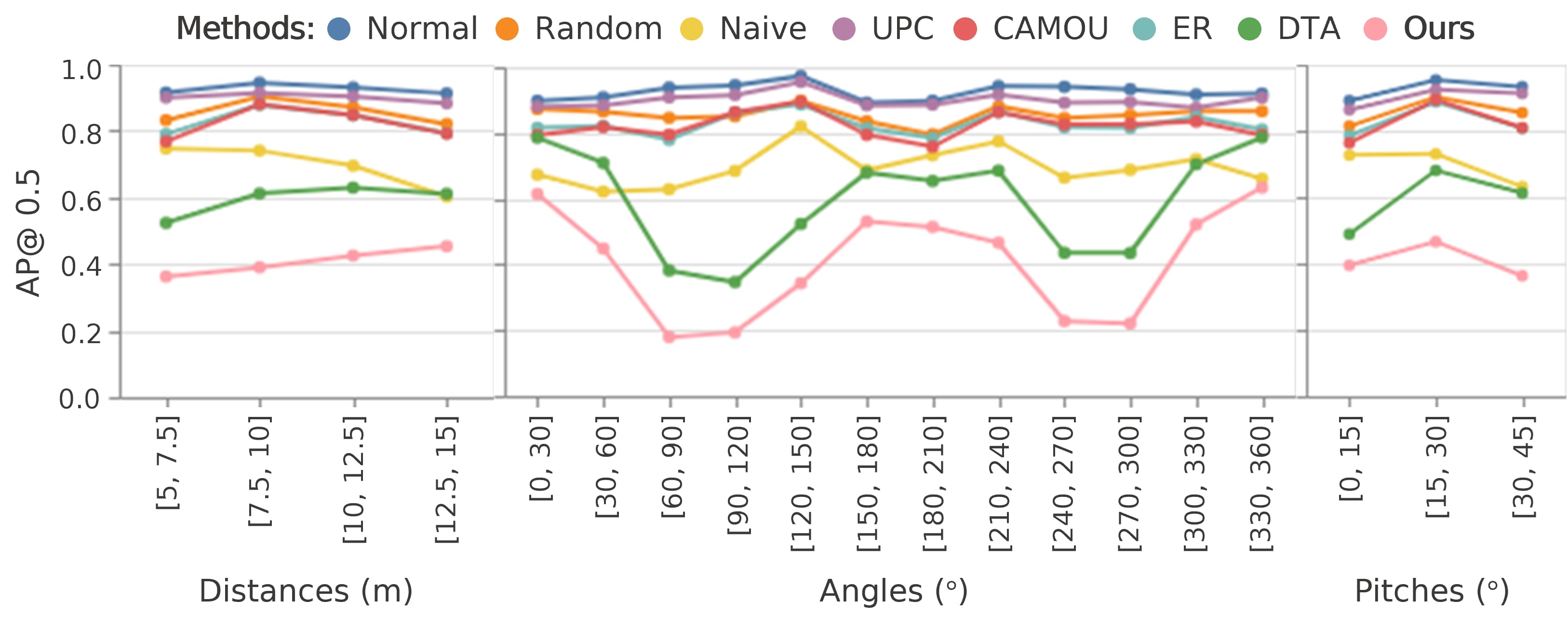}
        \caption{YOLOv7 \cite{wang2022yolov7}}
        \label{fig:transferability_yolov7}
    \end{subfigure}
    \begin{subfigure}[b]{0.49\columnwidth}
        \centering
        \includegraphics[width=0.95\columnwidth]{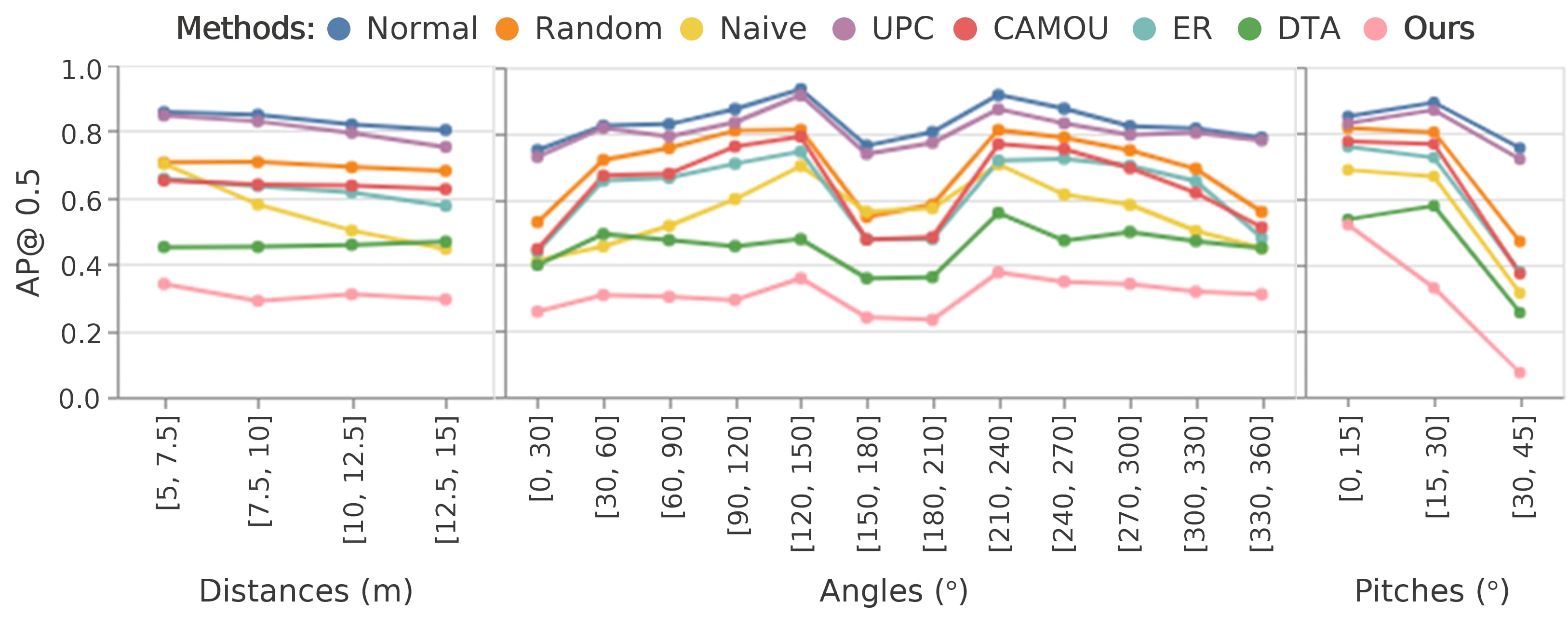}
        \caption{Dynamic R-CNN \cite{zhang2020dynamic}}
        \label{fig:transferability_dynamicrcnn}
    \end{subfigure}
    \begin{subfigure}[b]{0.49\columnwidth}
        \centering
        \includegraphics[width=0.95\columnwidth]{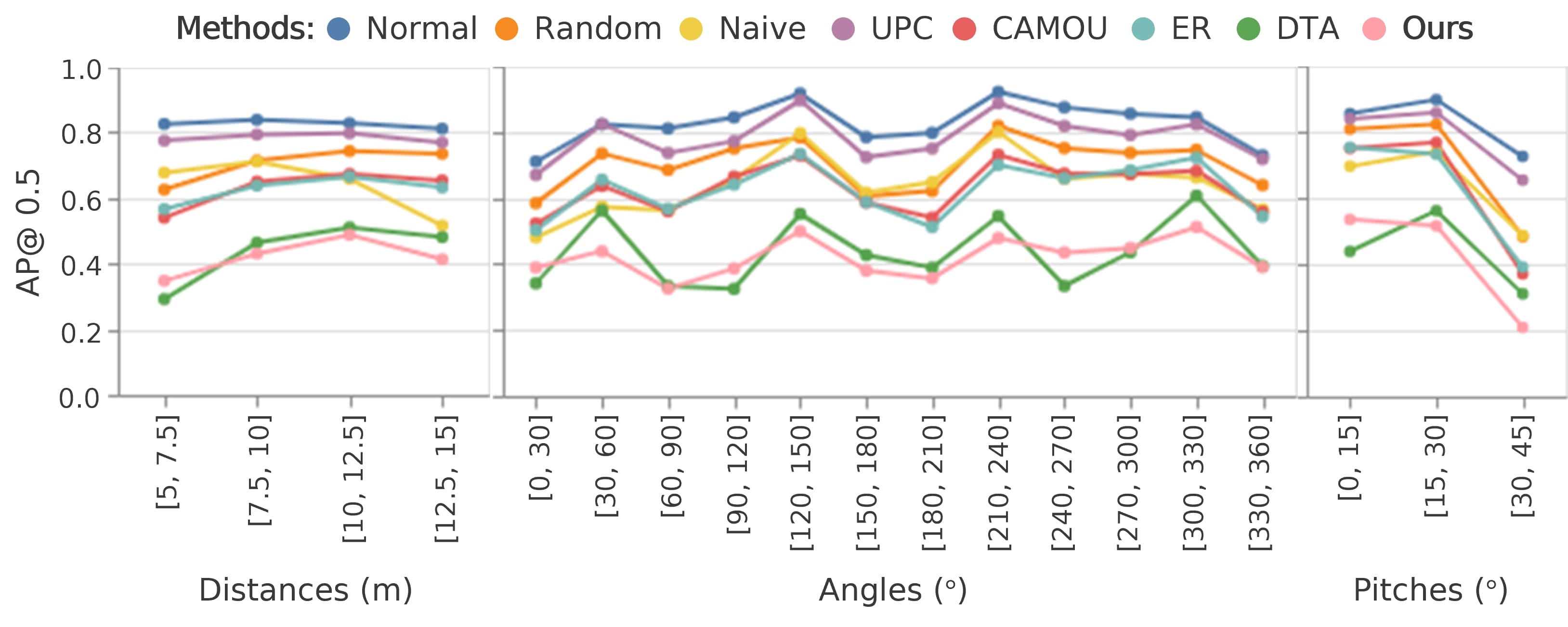}
        \caption{Sparse R-CNN \cite{sun2021sparse}}
        \label{fig:transferability_sparsercnn}
    \end{subfigure}
    \begin{subfigure}[b]{0.49\columnwidth}
        \centering
        \includegraphics[width=0.95\columnwidth]{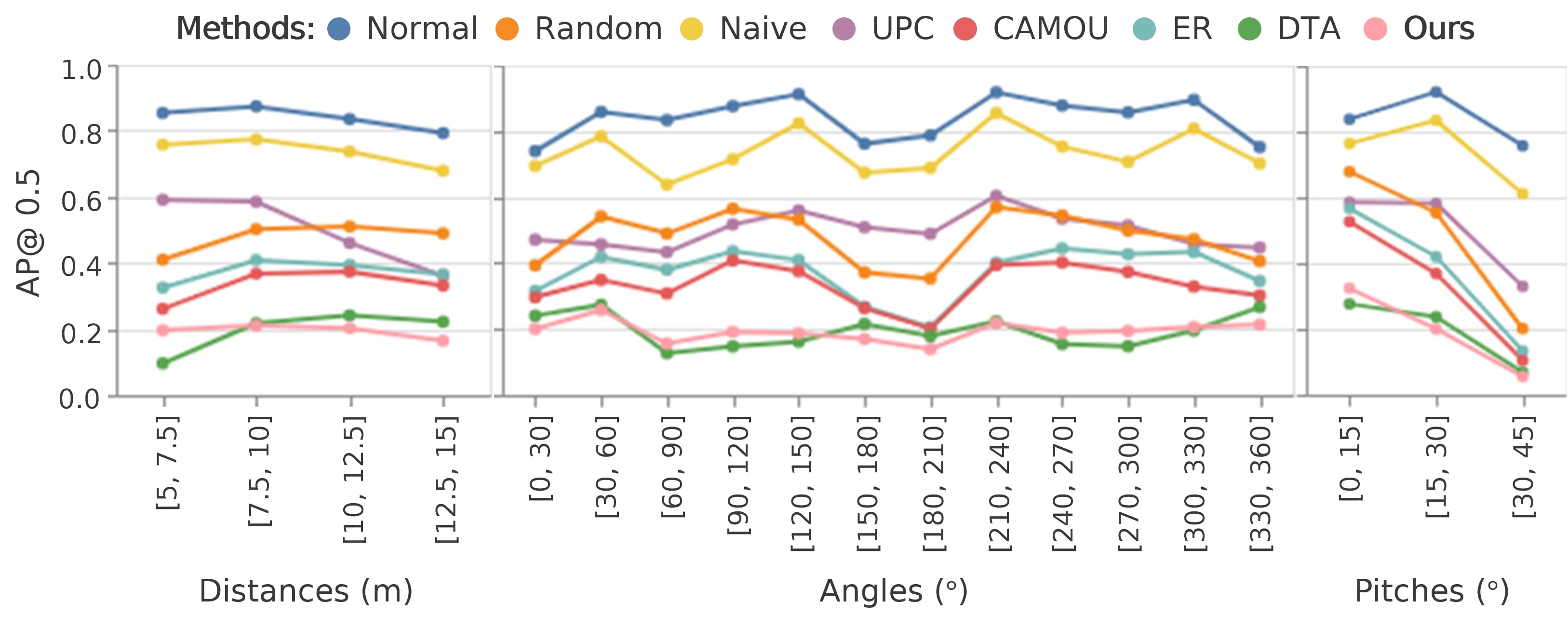}
        \caption{Deformable DETR \cite{zhu2021deformable}}
        \label{fig:transferability_ddtr}
    \end{subfigure}
    \begin{subfigure}[b]{0.49\columnwidth}
        \centering
        \includegraphics[width=0.9\columnwidth]{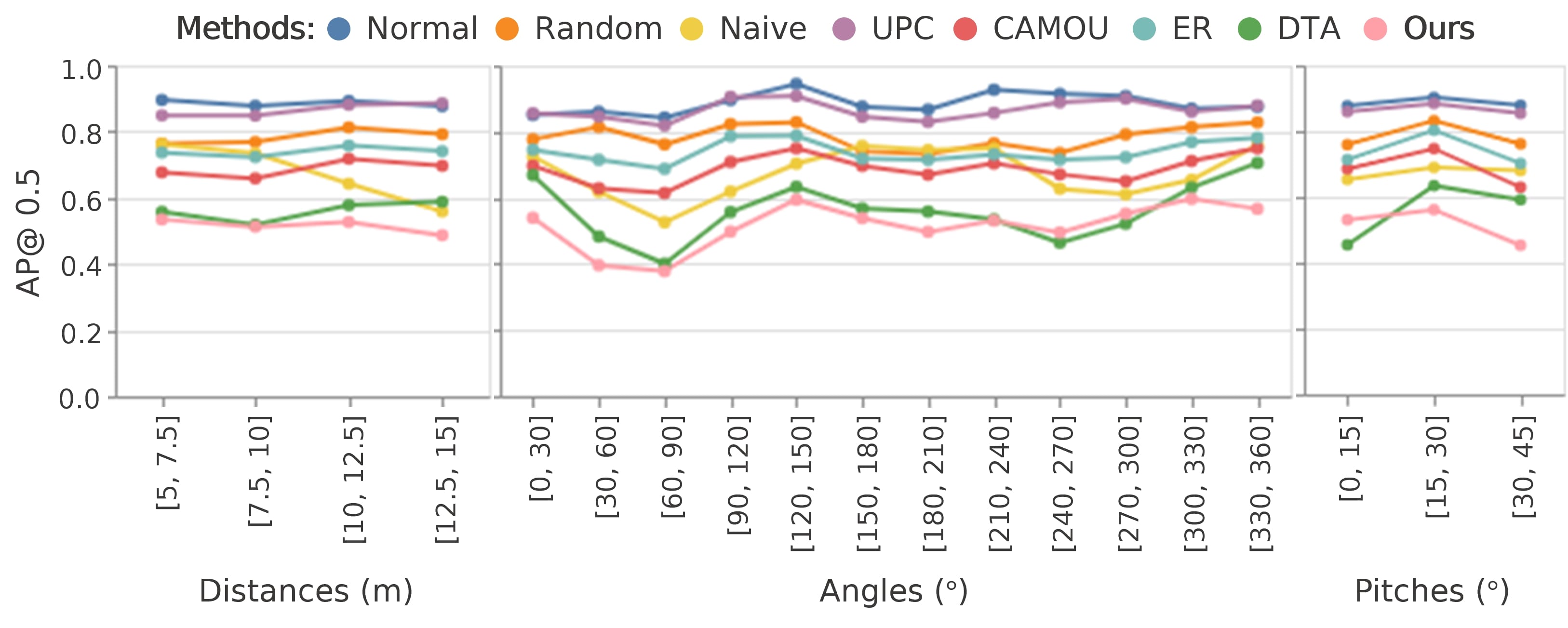}
        \caption{Pyramid Vision Transformer \cite{wang2021pyramid}}
        \label{fig:transferability_pvt}
    \end{subfigure}
    \caption{Universality evaluation on different camera poses. Target objects, models, and scenes differ from the training setting. Values are Average Precision @0.5 of the target car.}
    \label{fig:transferability_evaluations}
\end{figure}

\begin{figure}[h]
    \centering
    \begin{subfigure}[h]{\columnwidth}
        \centering
        \includegraphics[width=\columnwidth]{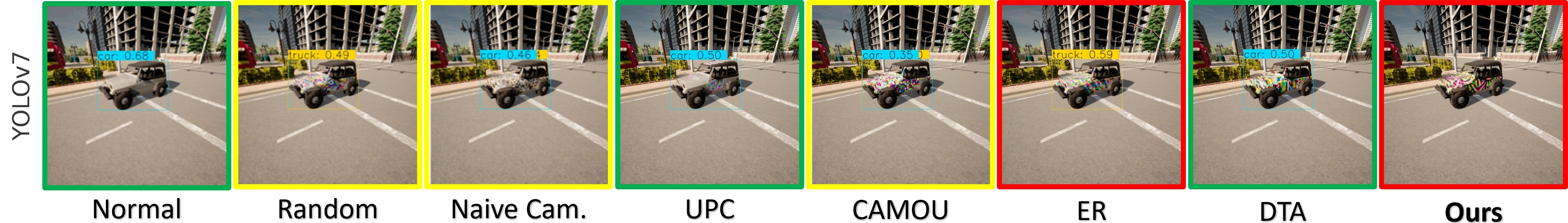}
    \end{subfigure}
\end{figure}

\begin{figure}[h] \ContinuedFloat
    \centering
    \begin{subfigure}[h]{\columnwidth}
        \centering
        \includegraphics[width=\columnwidth]{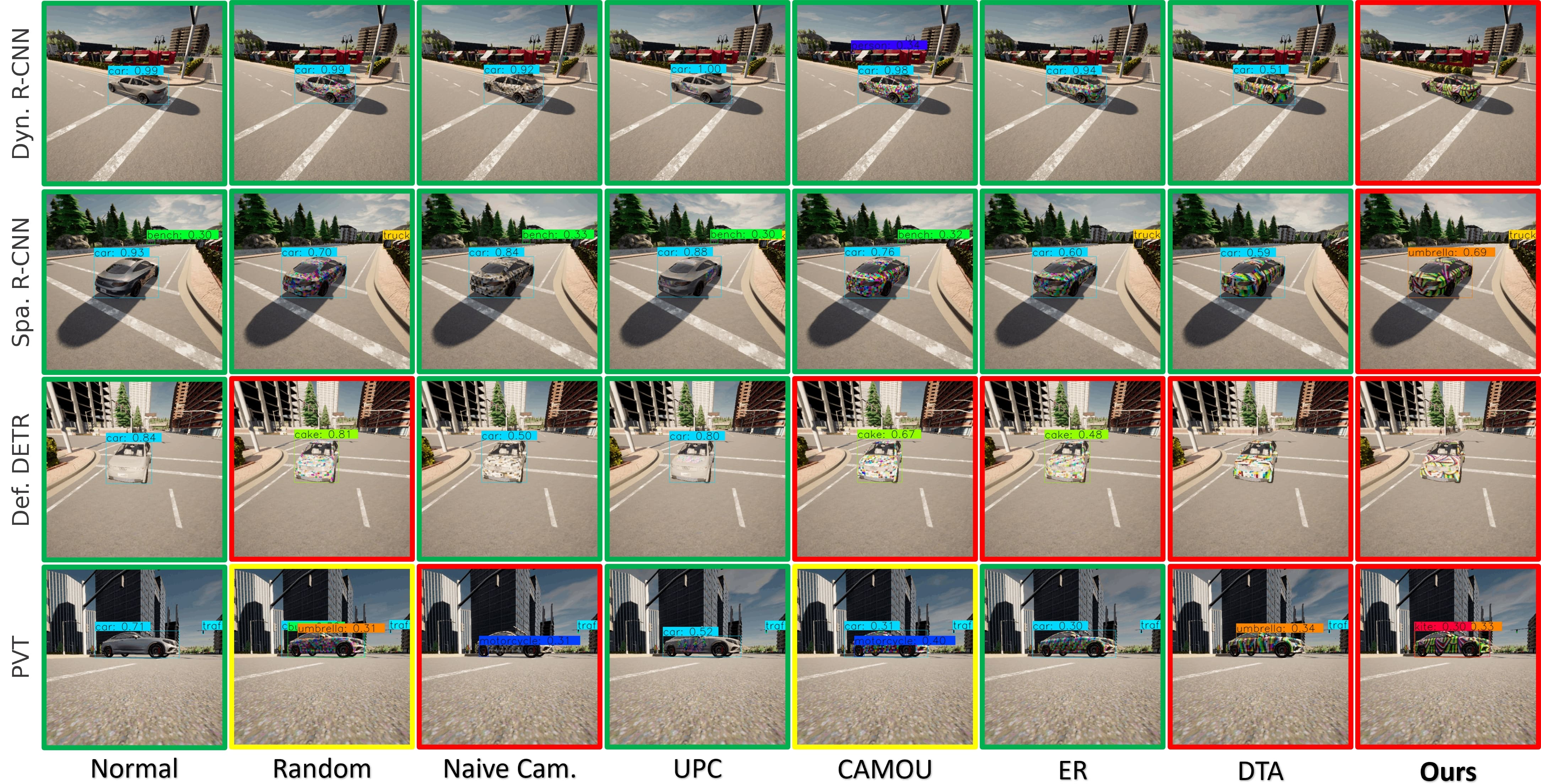}
    \end{subfigure}
    % \begin{subfigure}[h]{\columnwidth}
    %     \centering
    %     \includegraphics[width=\columnwidth]{images/supplementary_material/universality/universality_sample_dynamic_rcnn.jpg}
    % \end{subfigure}
    %   \begin{subfigure}[h]{\columnwidth}
    %     \centering
    %     \includegraphics[width=\columnwidth]{images/supplementary_material/universality/universality_sample_sparse_rcnn.jpg}
    % \end{subfigure}
    % \begin{subfigure}[h]{\columnwidth}
    %     \centering
    %     \includegraphics[width=\columnwidth]{images/supplementary_material/universality/universality_sample_ddetr.jpg}
    % \end{subfigure}
    % \begin{subfigure}[h]{\columnwidth}
    %     \centering
    %     \includegraphics[width=\columnwidth]{images/supplementary_material/universality/universality_sample_pvt.jpg}
    % \end{subfigure}
    \caption{Sample predictions for each method on various models and cars. Green = correct; Yellow = partially correct; Red = misdetection. Zoom for detail.}
    \label{fig:transferability_samples}
\end{figure}

\subsection{Transferability to Different Class (Truck and Bus)}
We perform evaluations to inspect whether the proposed adversarial camouflage can be transferred to other vehicle classes. We select a truck and a Volkswagen bus \textemdash which are available on the CARLA \textemdash as the target objects, and use the same textures and evaluated models from the previous evaluation. The detailed attack performance comparison for each method and evaluated model can be seen in Fig. \ref{fig:transferability_to_different_class}, showing that our camouflage still performs best compared to others. Additionally, YOLOv7 sample prediction for each method and object can be seen in Fig.} \ref{fig:different_class_samples}.

\begin{figure}[h]
\centerline{\includegraphics[width=0.825\columnwidth]{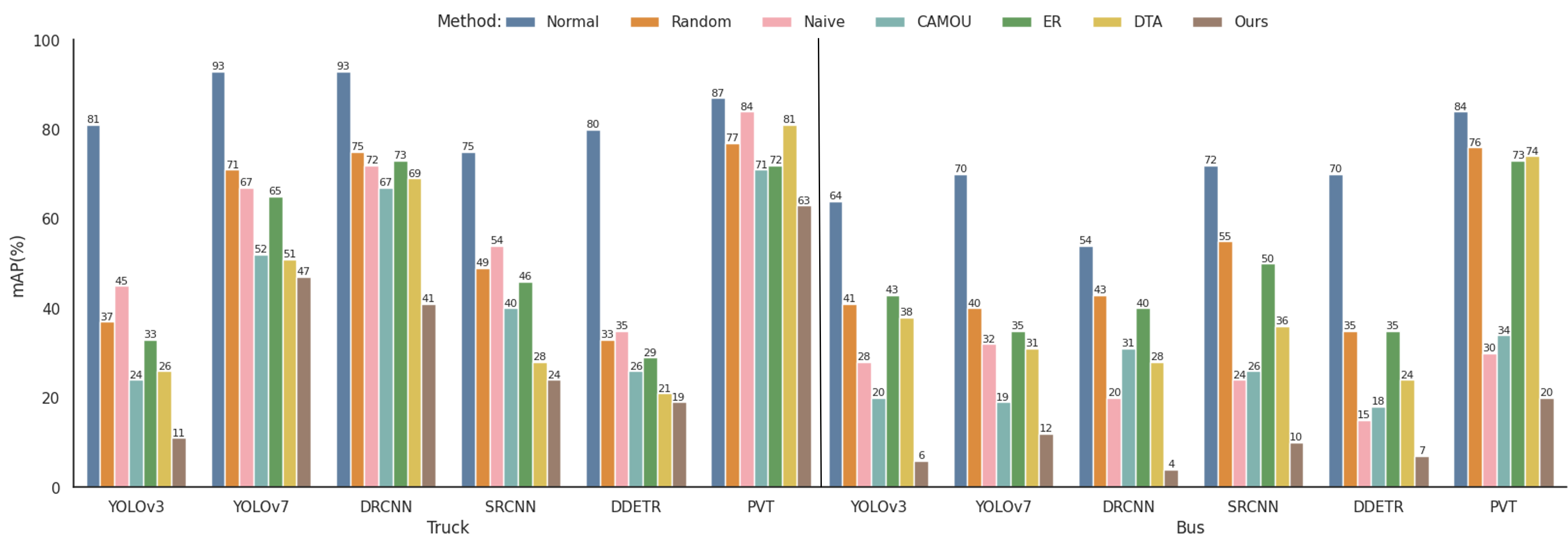}}
\caption{Transferability to different vehicle classes (truck and bus).}
\label{fig:transferability_to_different_class}
\end{figure}

\begin{figure}[h]
\centerline{\includegraphics[width=\columnwidth]{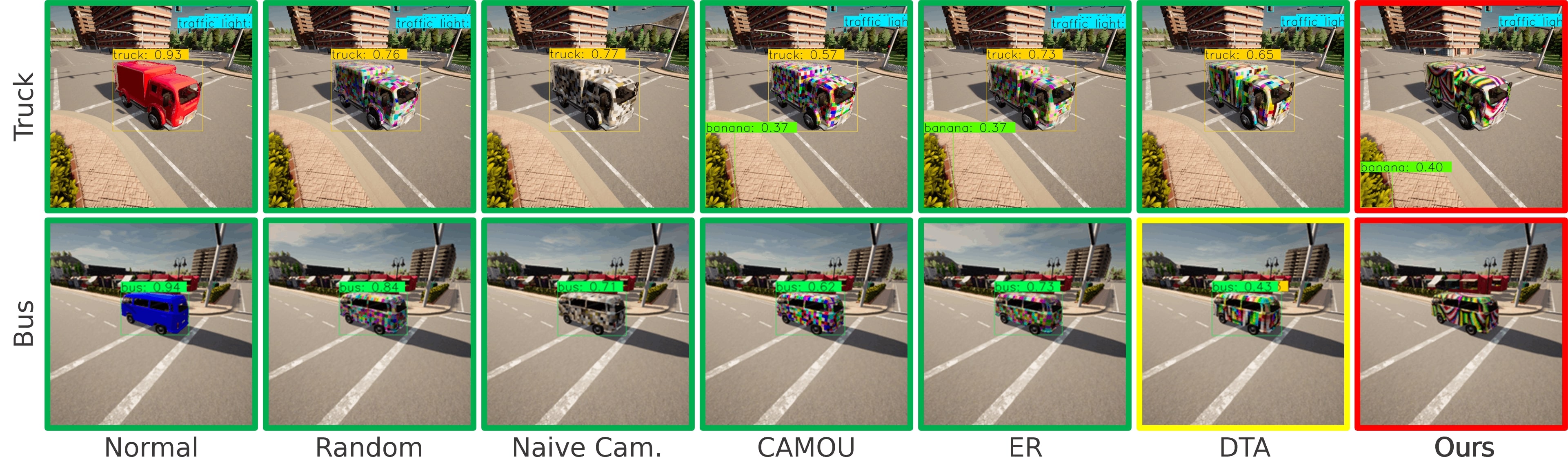}}
\caption{YOLOv7 sample predictions for truck and bus. Green = correct; Yellow = partially correct; Red = misdetection. Zoom for detail.}
\label{fig:different_class_samples}
\end{figure}

% \begin{figure}[h]
%     \centering
%     \begin{subfigure}[h]{\columnwidth}
%         \centering
%         \includegraphics[width=\columnwidth]{images/supplementary_material/different_class/different_class_sample_truck.jpg}
%     \end{subfigure}
% \end{figure}

% \begin{figure}[h] % \ContinuedFloat
%     \centering
%     \begin{subfigure}[h]{\columnwidth}
%         \centering
%         \includegraphics[width=\columnwidth]{images/supplementary_material/different_class/different_class_sample_truck.jpg}
%     \end{subfigure}
%     \begin{subfigure}[h]{\columnwidth}
%         \centering
%         \includegraphics[width=\columnwidth]{images/supplementary_material/different_class/different_class_sample_bus.jpg}
%     \end{subfigure}
%      \caption{YOLOv7 sample predictions for truck and bus. Green = correct; Yellow = partially correct; Red = misdetection. Zoom for detail.}
%     \label{fig:different_class_samples}
% \end{figure}

\subsection{Transferability to Different Task (Segmentation Models)}
% TODO: Add paragraphs about what we are doing here.
% TODO: Add paragraphs about the sample images
We use segmentation models such as MaX-DeepLab-L \cite{wang2021max} and Axial-DeepLab \cite{wang2020axial} to evaluate the transferability to different tasks. We evaluate the pixel accuracy of the car label to show how the adversarial camouflage can decrease the prediction of the target object. Fig. \ref{fig:segmentation_transferability} shows that our method yields significantly higher attack performance compared to other methods, even for the segmentation models. Sample predictions can be seen in Fig. \ref{fig:segmentation_sample}.

\begin{figure}[h]
    \centering
    \begin{subfigure}[b]{0.49\columnwidth}
        \centering
        \includegraphics[width=0.95\columnwidth]{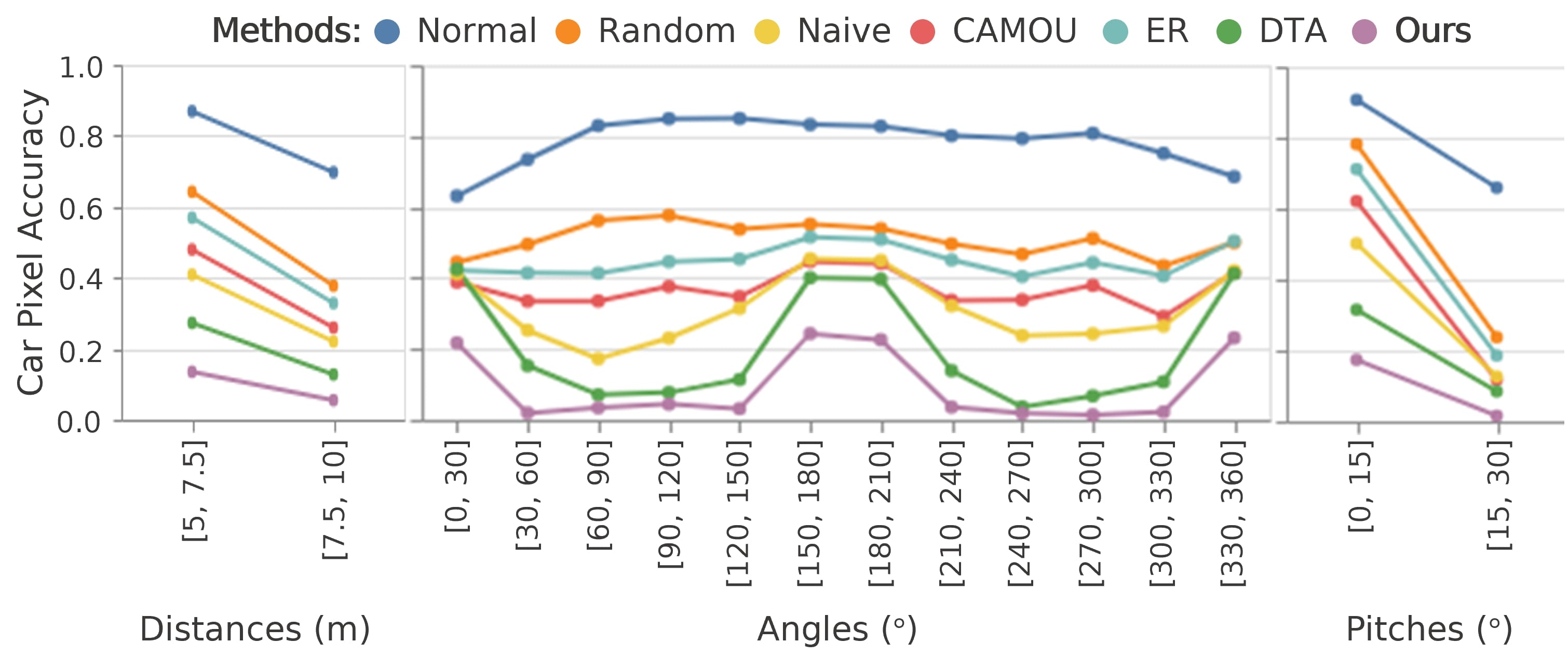}
        \caption{Cityscape Pretrained MaX-DeepLab-L \cite{wang2021max}}
        \label{fig:segmentation_maxdeeplab_cityscape}
    \end{subfigure}
    \begin{subfigure}[b]{0.49\columnwidth}
        \centering
        \includegraphics[width=0.95\columnwidth]{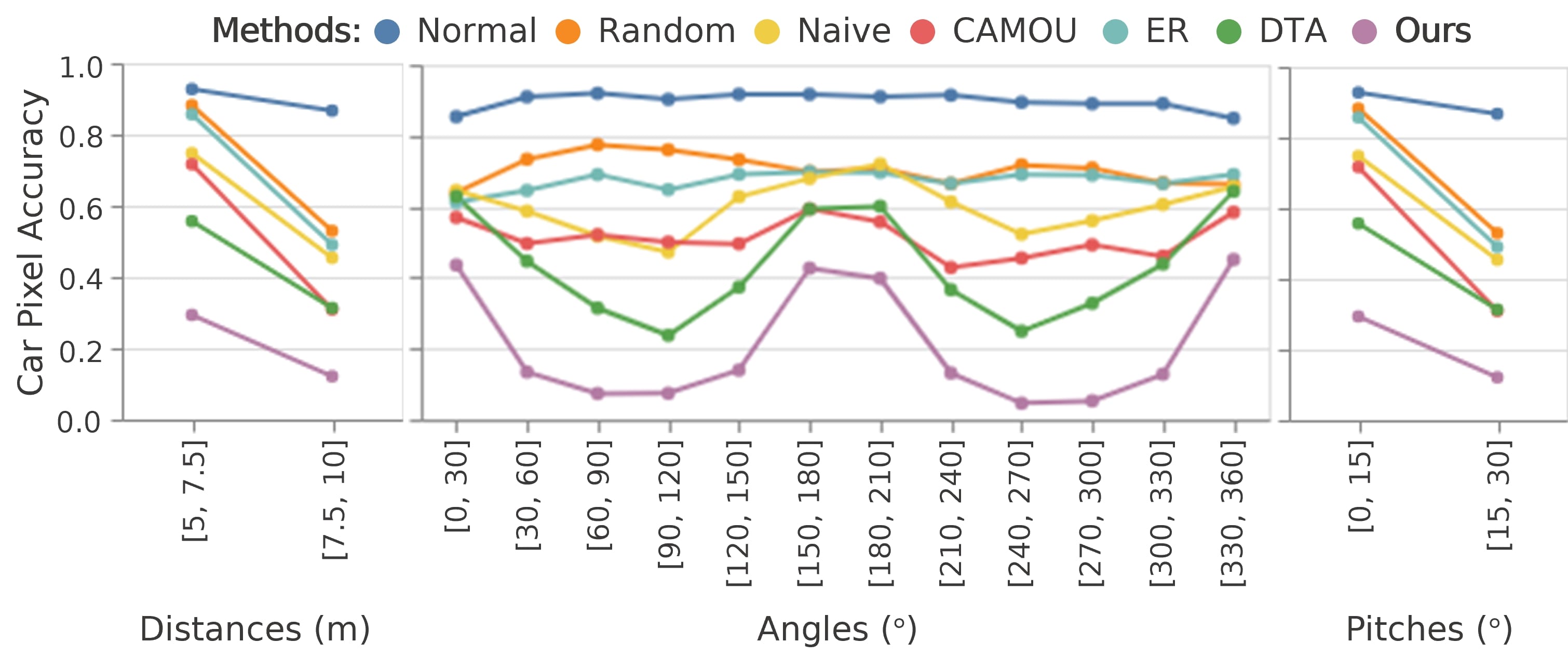}
        \caption{Cityscape Pretrained Axial-DeepLab \cite{wang2020axial}}
        \label{fig:segmentation_axial_cityscape}
    \end{subfigure}
    \begin{subfigure}[b]{0.49\columnwidth}
        \centering
        \includegraphics[width=0.95\columnwidth]{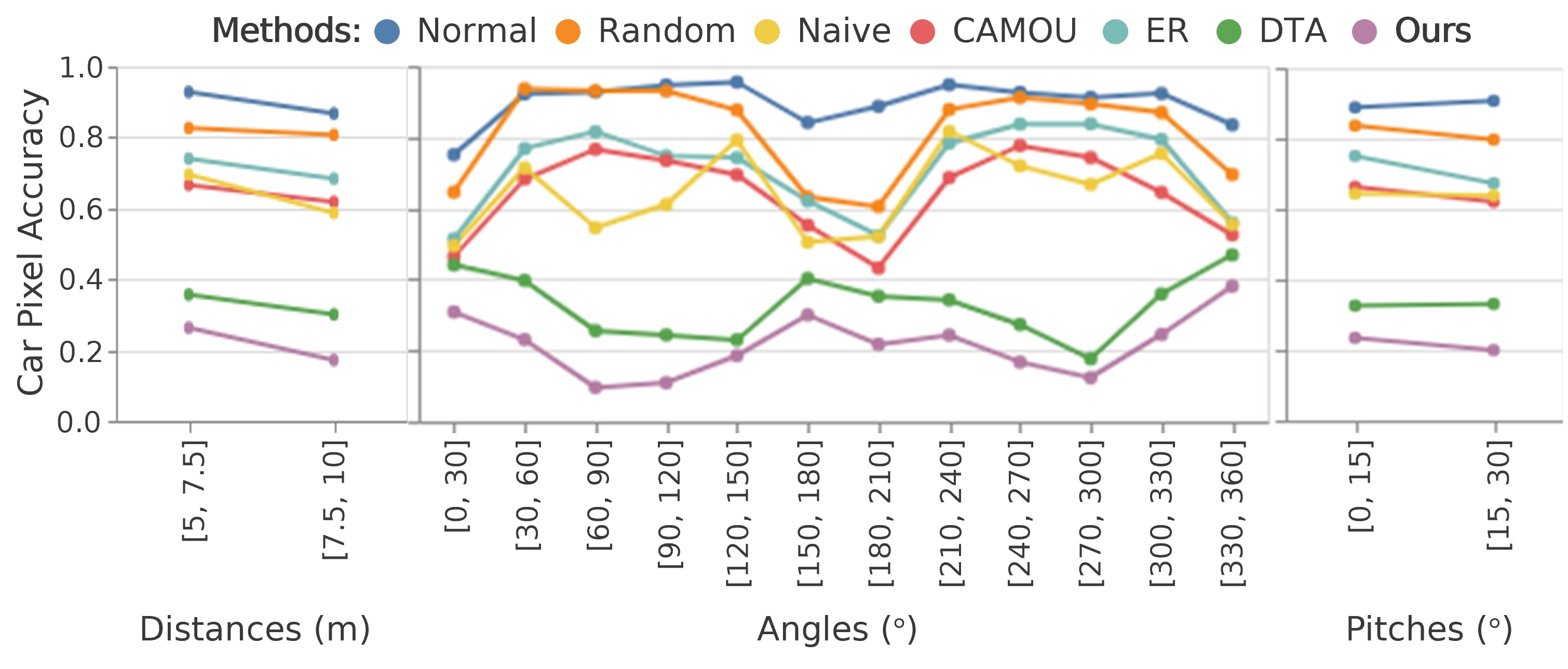}
        \caption{COCO Pretrained MaX-DeepLab-L \cite{wang2021max}}
        \label{fig:segmentation_maxdeeplab_coco}
    \end{subfigure}
    \caption{Transferability to segmentation models on different camera poses. Values are Pixel Accuracy of the target car.}
    \label{fig:segmentation_transferability}
\end{figure}

\begin{figure}[h]
    \centering
    \begin{subfigure}[h]{\columnwidth}
        \centering
        \includegraphics[width=\columnwidth]{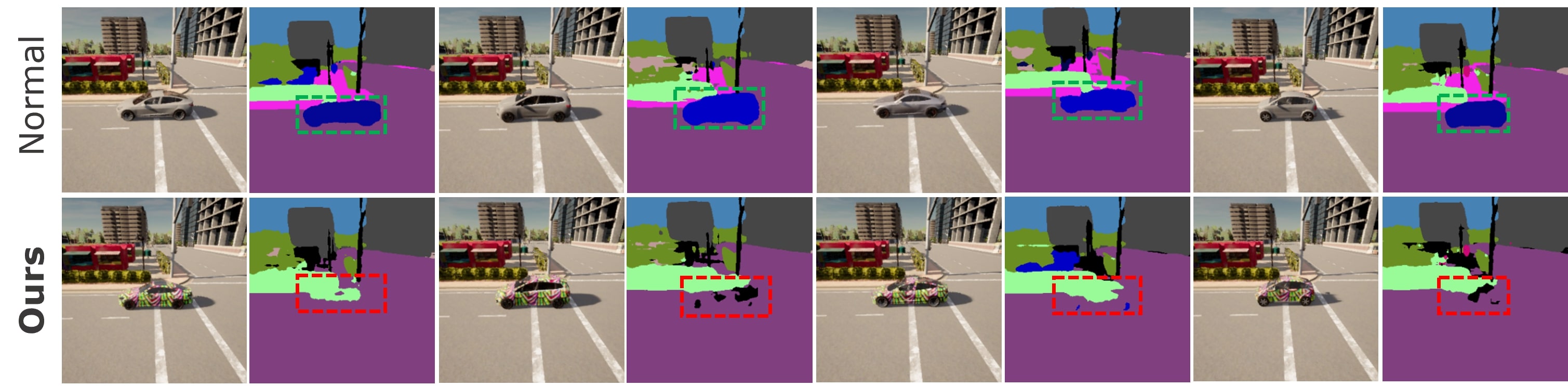}
    \end{subfigure}
    \caption{Sample predictions of normal and our camouflage car on Axial-DeepLab. The camouflage works universally on multiple cars.}
    \label{fig:segmentation_sample}
\end{figure}

\newpage

\subsection{Transferability to the Real World}
% TODO: Add paragraphs about what we are doing here.
% Describe experiment setting for Real World Evaluation
% Describe camera config used in Real World Evaluation
% Add paragraphs about the sample images

We conduct a real-world experiment by fabricating two 1:10-scaled Tesla Model 3s with a 3D printer: one for a normal and another for our camouflaged car with the texture targeting YOLOv3. We select the adversarial camouflage with the best attack performance (i.e., the lowest AP of YOLOv3) to reliably measure the transferability to the real world when producing a camouflaged car. We place the car models in five real-world locations and capture car images for every 45-degree interval in angles ([0, 45], [45, 90], $\dots$, [315, 360] degrees) with three different distances ([5, 7.5], [7.5, 10], [10, 12.5] meters) and two different pitches ([0, 15], [15, 30] degrees) using iPhone 11 Pro Max. As a result, a total of 240 images (5 locations $\times$ 8 rotations $\times$ 3 distances $\times$ 2 pitches) are taken for each camera pose, and 240 images for each car model. In summary, a total of 480 images are used for evaluation. 

We evaluate practical real-time object detectors widely used in the real world: MobileNetV2 \cite{sandler2018mobilenetv2}, EfficientDet-D2 \cite{tan2020efficientdet}, YOLOX-L \cite{ge2021yolox}, and YOLOv7 \cite{wang2022yolov7}. From Fig. \ref{fig:real_world_evaluations}, our adversarial camouflaged car can significantly decrease AP@0.5 of all object detection models in the real world. Fig. \ref{fig:real_world_evaluation_samples} shows that the normal car model is well detected, whereas the camouflaged car model is not detected as a car at all under various camera poses.

\begin{figure}[H]
    \centering
    \begin{subfigure}[b]{0.49\columnwidth}
        \centering
        \includegraphics[width=\columnwidth]{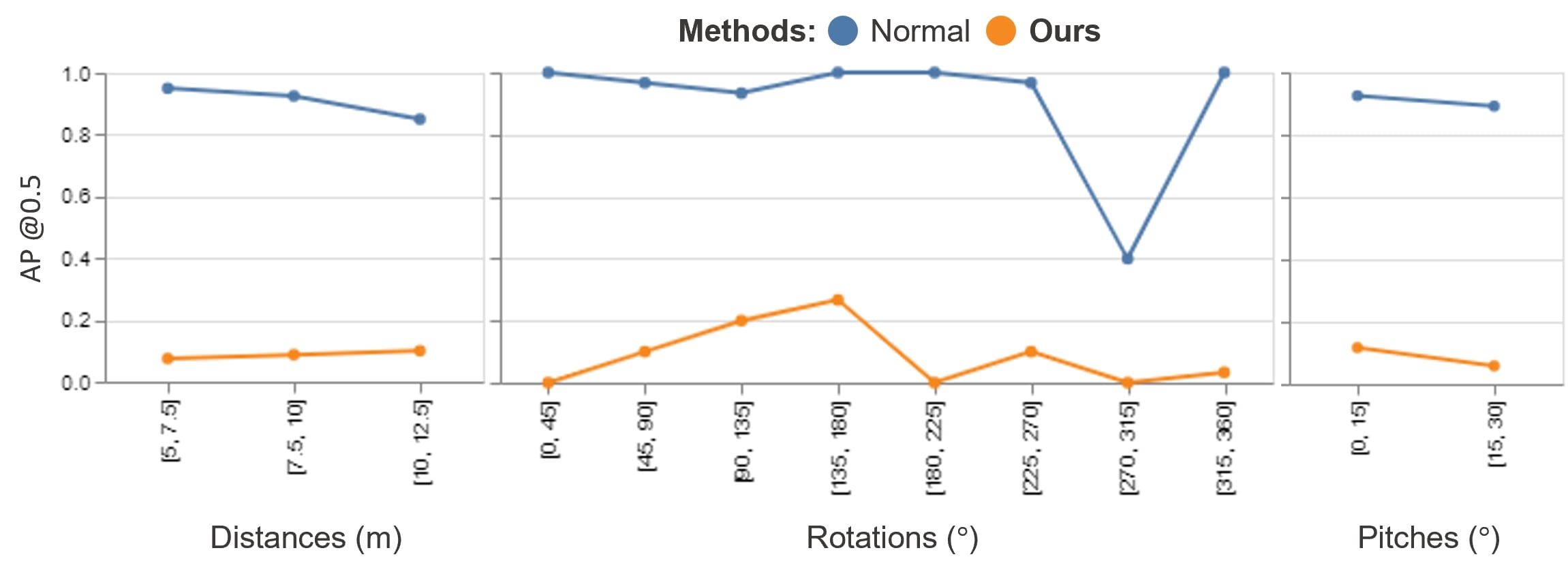}
        \caption{YOLOv3 \cite{redmon2018yolov3}}
        \label{fig:real_world_yolov3}
    \end{subfigure}
    \begin{subfigure}[b]{0.49\columnwidth}
        \centering
        \includegraphics[width=\columnwidth]{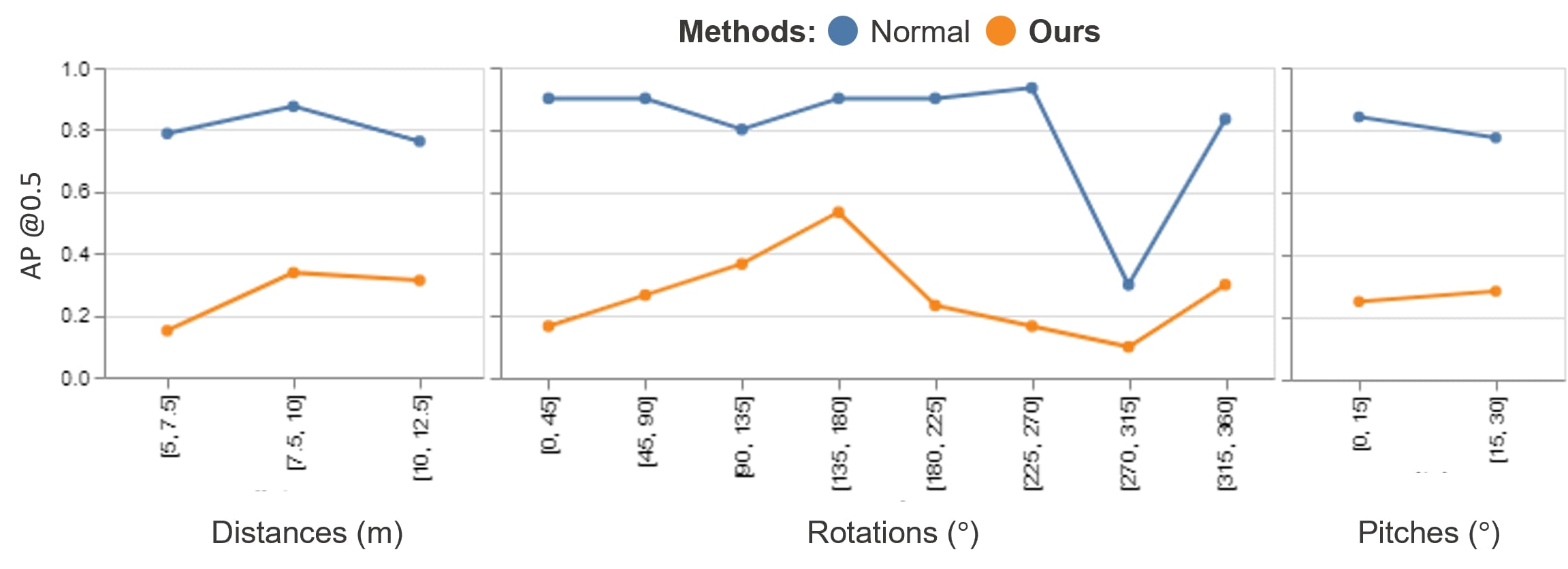}
        \caption{MobileNetv2 \cite{sandler2018mobilenetv2}}
        \label{fig:real_world_mobilenetv2}
    \end{subfigure}
    \begin{subfigure}[b]{0.49\columnwidth}
        \centering
        \includegraphics[width=\columnwidth]{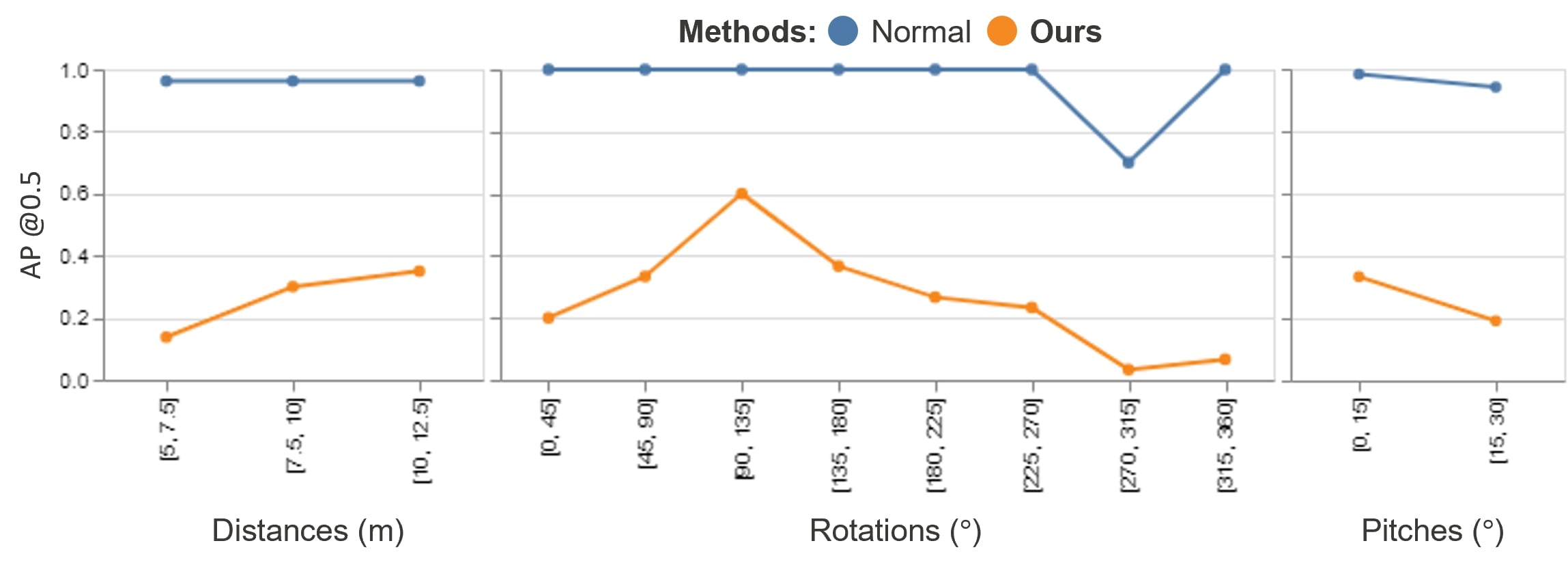}
        \caption{EfficientDet-D2 \cite{tan2020efficientdet}}
        \label{fig:real_world_effdetd2}
    \end{subfigure}
    \begin{subfigure}[b]{0.49\columnwidth}
        \centering
        \includegraphics[width=\columnwidth]{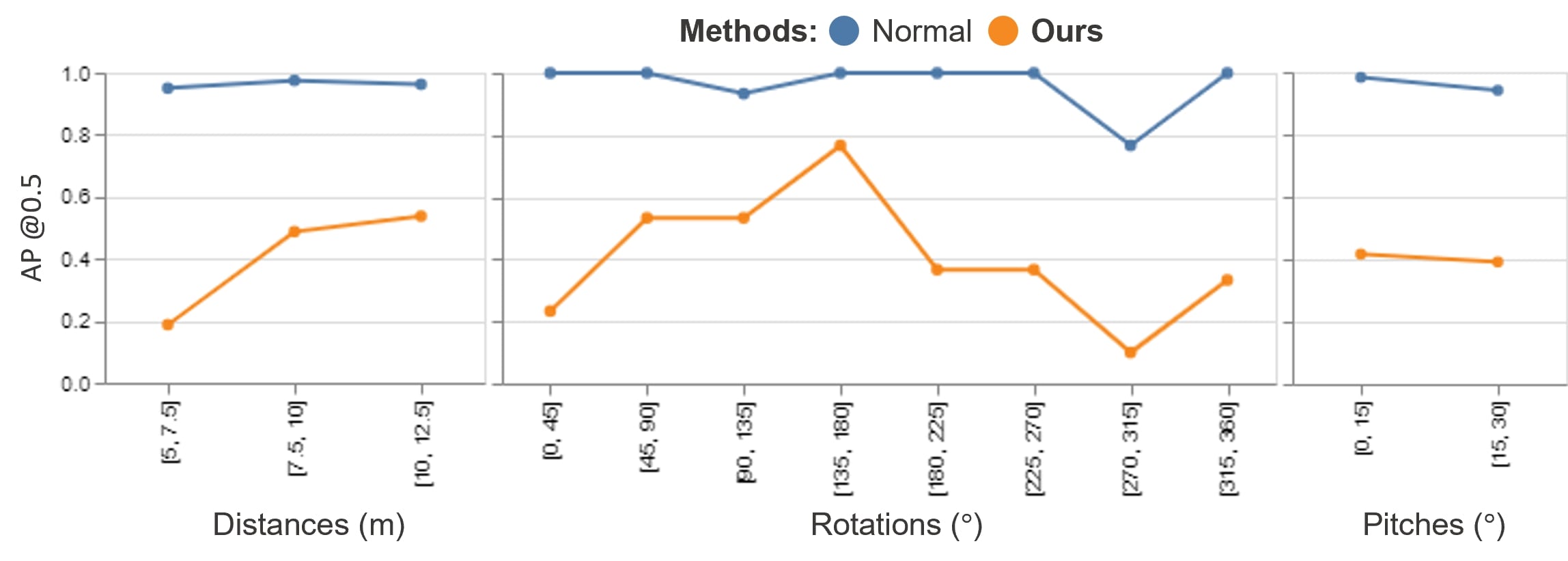}
        \caption{YOLOX-L \cite{ge2021yolox}}
        \label{fig:real_world_yolox}
    \end{subfigure}
    \begin{subfigure}[b]{0.49\columnwidth}
        \centering
        \includegraphics[width=\columnwidth]{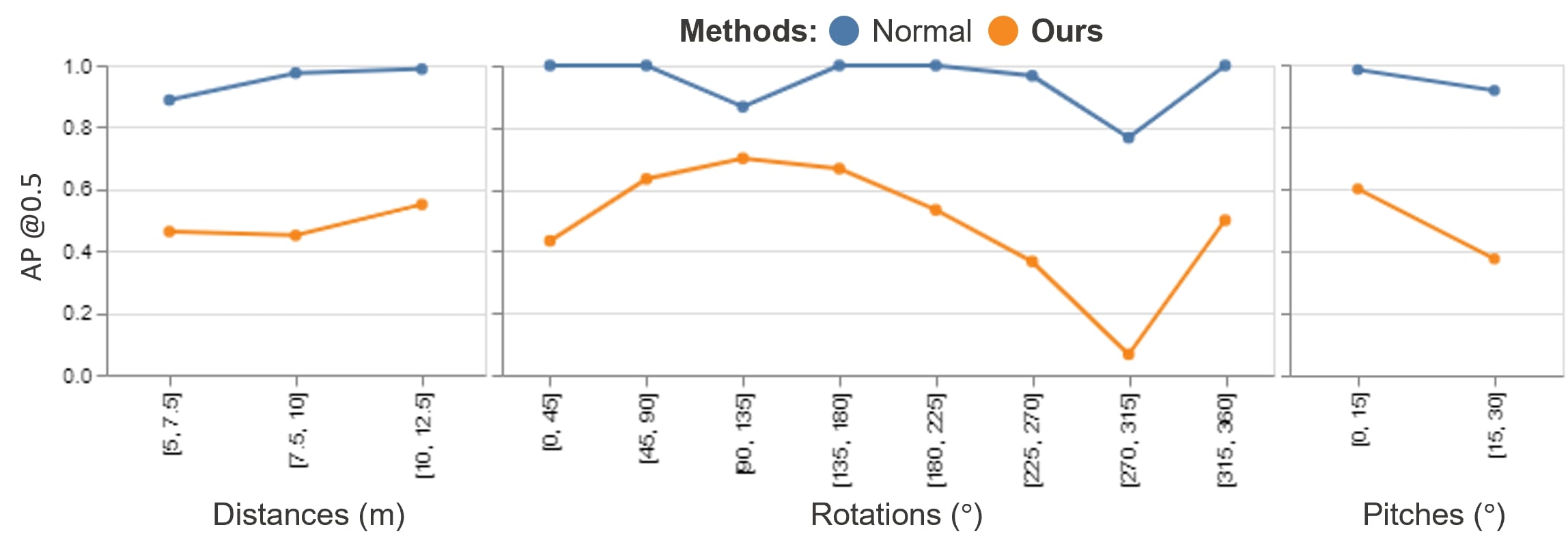}
        \caption{YOLOv7 \cite{wang2022yolov7}}
        \label{fig:real_world_yolov7}
    \end{subfigure}
    \caption{Transferability to real world on different camera poses. Values are Average Precision @0.5 of the target car.}
    \label{fig:real_world_evaluations}
\end{figure}

\begin{figure}[H]
    \centering
    \begin{subfigure}[b]{\columnwidth}
        \centering
        \includegraphics[width=0.90\columnwidth]{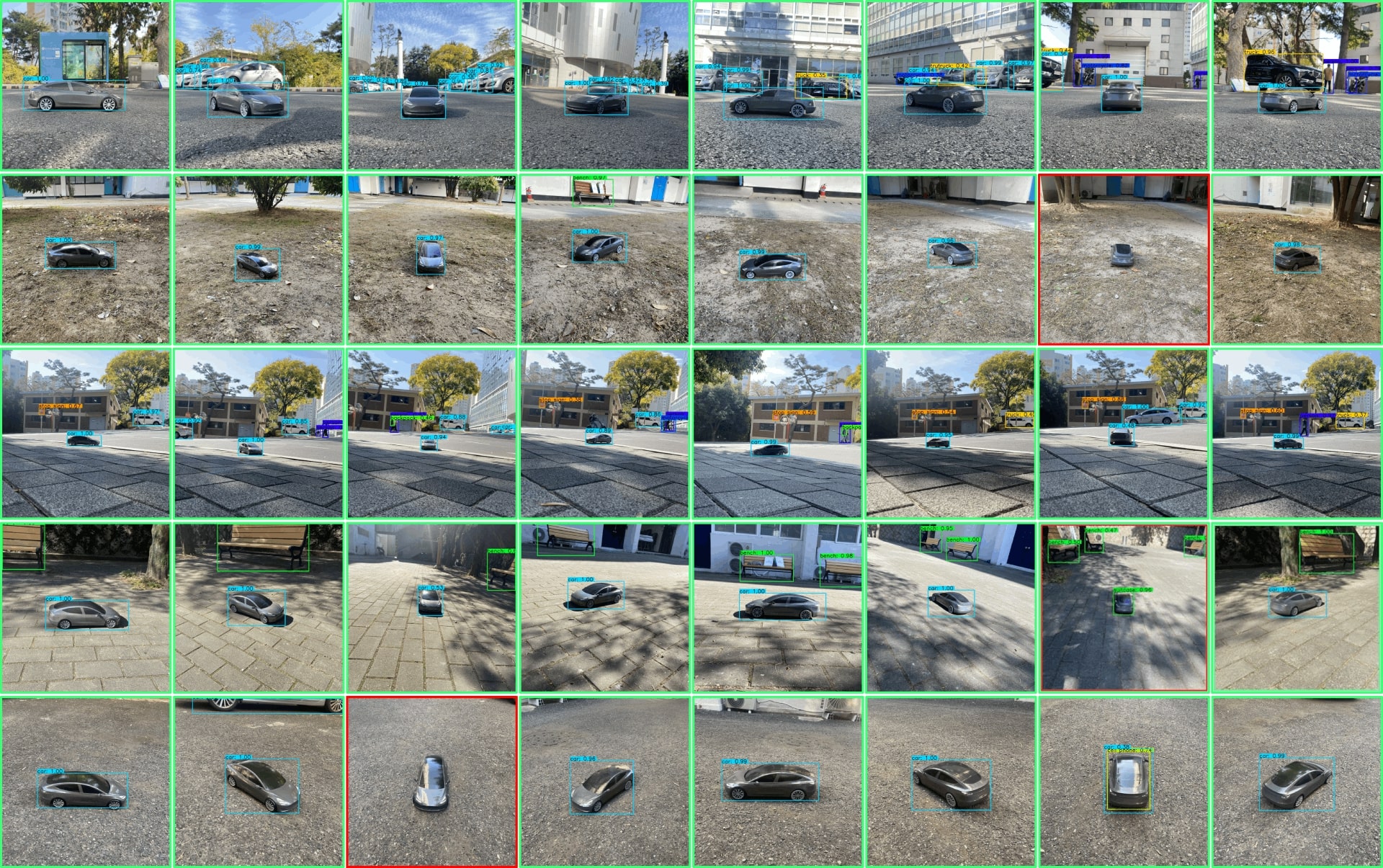}
        \caption{Normal car. Cars can be consistently detected with relatively high scores.}
        \label{fig:real_world_normal}
    \end{subfigure}
    \begin{subfigure}[b]{\columnwidth}
        \centering
        \includegraphics[width=0.90\columnwidth]{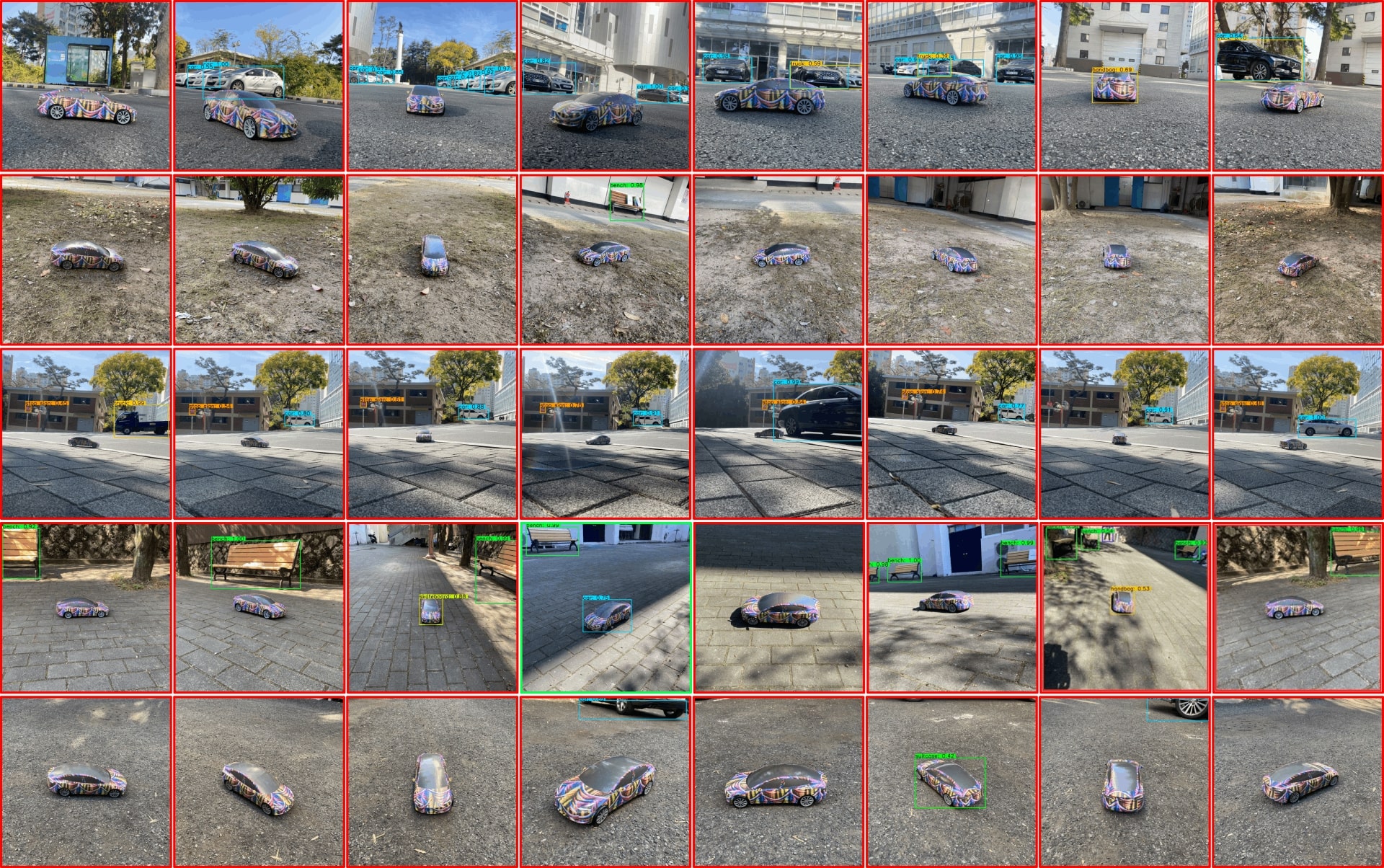}
        \caption{Our camouflaged car. It can evade most of the detection with a high success rate, while background objects can still be detected correctly.}
        \label{fig:real_world_attack}
    \end{subfigure}
    \caption{Sample predictions of our real-world evaluation using two scaled car models. We randomly sample 240 real-world images for a single car consisting of five locations, three distances, eight rotations, and two pitch angles as camera transformation sets. Zoom for detail.}
    \label{fig:real_world_evaluation_samples}
\end{figure}

\newpage

\section{Ablation Study}
\subsection{Effect of Different Losses With Respect to Performance}
We perform a detailed analysis of our loss hyperparameters, including our smooth loss $\beta$ and camouflage loss $\gamma$ with respect to the attack performance in the white-box setting (targeting YOLOv3). We fix our stealth loss $\alpha$ as 1.0 and run a grid search for both $\beta$ and $\gamma$ with the value in the range of $[0.0, 0.25, 0.5, 0.75, 1.0 ]$.} As shown in Fig \ref{fig:loss_to_performance}, the smooth loss has a better correlation for increasing the attack performance than camouflage loss. Also, as mentioned by \cite{Mahendran_2015_CVPR}, extreme differences between adjacent pixels in the perturbation are unlikely to be accurately captured by cameras and may not be physically realizable; thus, the smooth loss becomes important in physical adversarial attacks, as verified in our experiment. On the other hand, utilizing our camouflage loss may result in a trade-off between attack performance and naturalness since it suppresses texture by using colors close to the extracted dominant background, limiting the use of other colors to enhance the attack. This can be seen from the AP@0.5 grouped by camouflage $\gamma$ showing that $\gamma=0$ has the lowest mean compared to others.

\begin{figure}[H]
\centerline{\includegraphics[width=0.95\columnwidth]{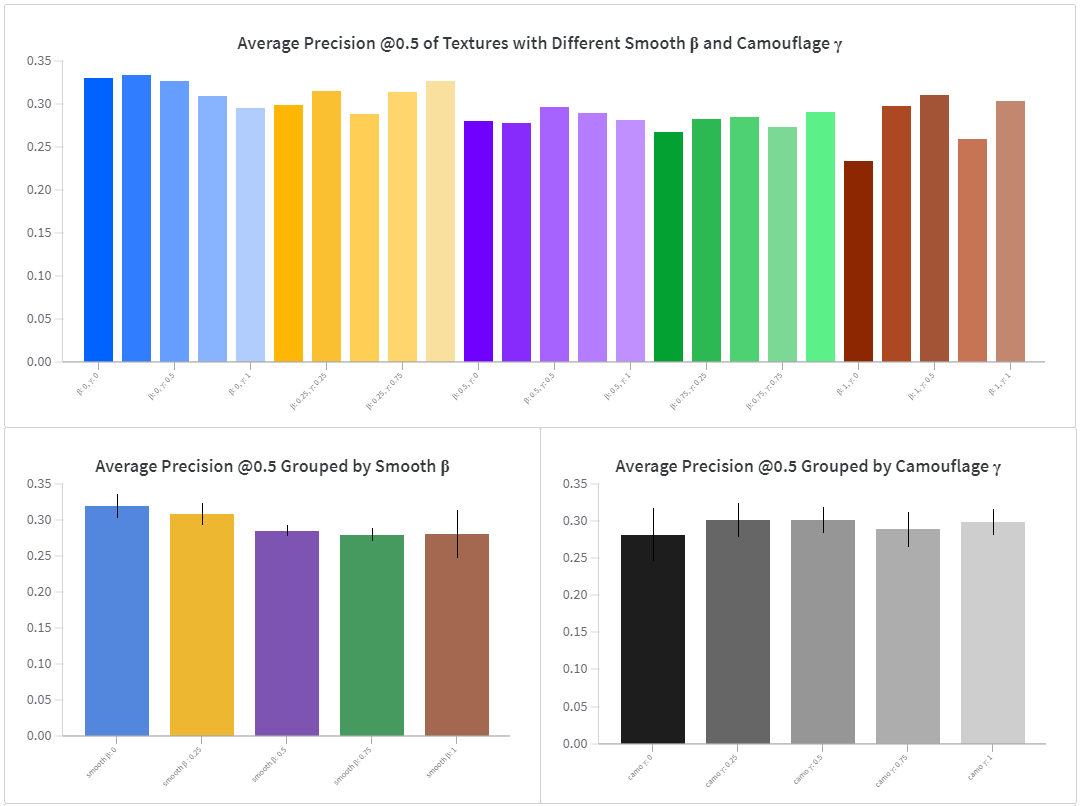}}
\caption{Performance graph for various smooth loss $\beta$ and camouflage loss $\gamma$.}
\label{fig:loss_to_performance}
\end{figure}

\subsection{Effect of Different Losses with Respect to Naturalness}
Here, we visualize the effect of different losses with respect to texture naturalness. As shown in Fig \ref{fig:loss_to_naturallness}, we can see that both smooth and camouflage loss have a role in texture naturalness. We also observe that higher smooth $\beta$ results in lower pixels variance, yielding a smooth texture, while higher camouflage $\gamma$ results in softer color texture. Additionally, only utilizing the smooth loss may result in a bright and garish color texture, attracting more human attention, whereas only employing camouflage loss may result in a rough texture. Thus, utilizing both smooth and camouflage loss yields in a more natural result with a smoother and softer texture. 

\begin{figure}[H]
\centerline{\includegraphics[width=\columnwidth]{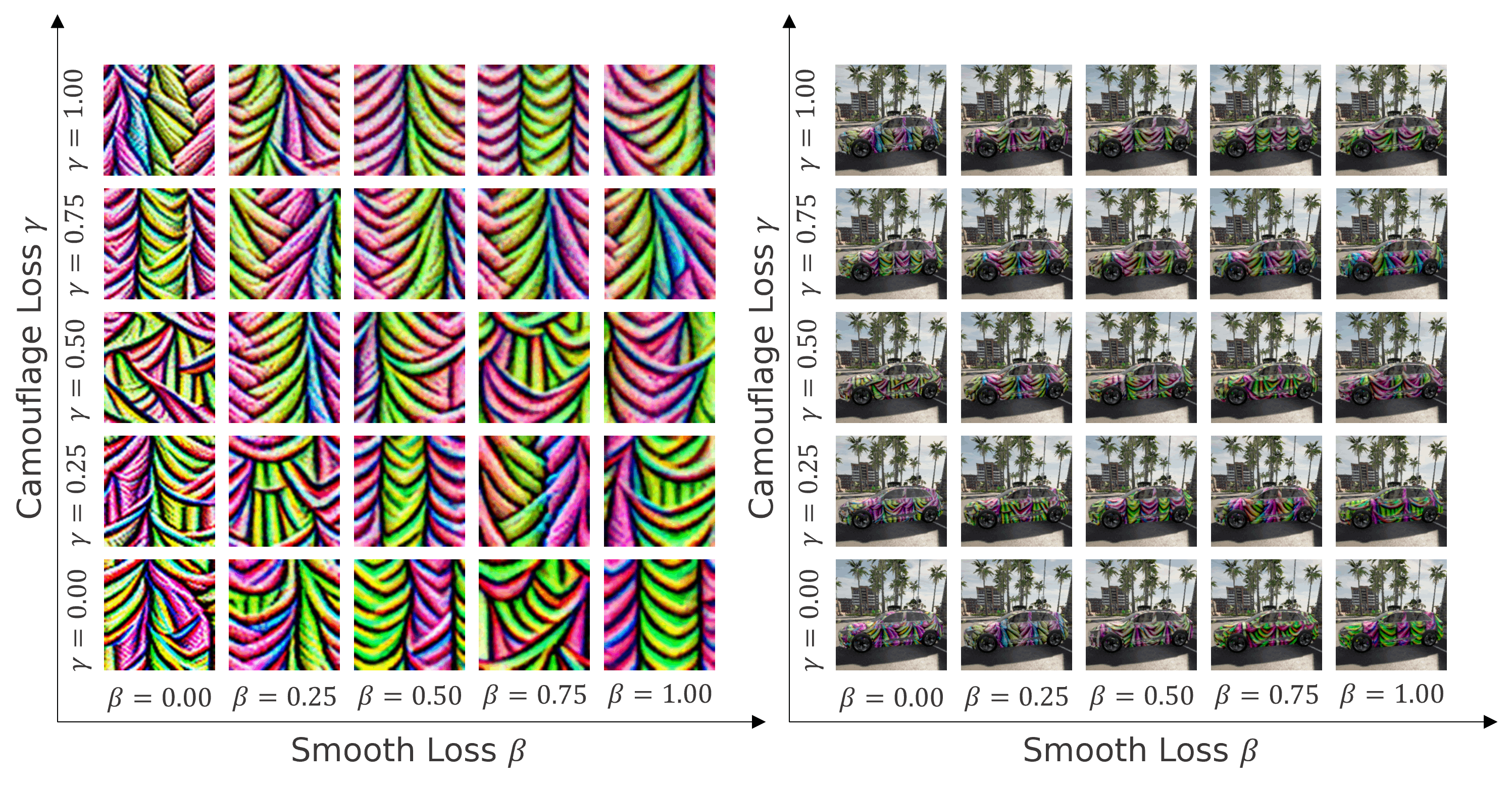}}
\caption{Our optimized adversarial camouflage texture with various smooth loss $\beta$ and camouflage loss $\gamma$: base texture (left), after rendered in UE4 (right).}
\label{fig:loss_to_naturallness}
\end{figure}

%HUONG: R2.1
\subsection{Effect of Different Target Models}
We conduct experiments using a state-of-the-art (SOTA) model to assess how different target models affect attack performance. The results of our experiments are presented in Table \ref{tab:diff-target-model}, indicating that when targeting YOLOv7, a more robust model, we obtain a corresponding increase in the robustness of the texture.
\begin{table}[H]
\caption{Universality evaluation on SOTA target model (YOLOv7)}
\label{tab:diff-target-model}
\resizebox{\columnwidth}{!}{
\fontsize{3pt}{3pt}\selectfont
\begin{tabular}{l|cccccc}
\hline
\rowcolor[HTML]{FFFC9E} 
\multicolumn{1}{c|}{\cellcolor[HTML]{FFFC9E}{\color[HTML]{000000} }} & \multicolumn{6}{c}{\cellcolor[HTML]{FFFC9E}{\color[HTML]{000000} \textbf{Evaluated Model - Car AP@0.5}}}               \\ \cline{2-7} 
\rowcolor[HTML]{FFFC9E} 
\multicolumn{1}{c|}{\multirow{-2}{*}{\cellcolor[HTML]{FFFC9E}{\color[HTML]{000000} \begin{tabular}[c]{@{}c@{}}\textbf{Methods}\\ (Target)\end{tabular}}}} &
  {\color[HTML]{000000} YOLOv3} &
  \multicolumn{1}{l|}{\cellcolor[HTML]{FFFC9E}{\color[HTML]{000000} YOLOv7}} &
  {\color[HTML]{000000} DRCN} &
  \multicolumn{1}{l|}{\cellcolor[HTML]{FFFC9E}{\color[HTML]{000000} SRCN}} &
  {\color[HTML]{000000} DDETR} &
  {\color[HTML]{000000} PVT} \\ \hline
Normal                                                               & 85.98 & \multicolumn{1}{l|}{92.54} & 83.20 & \multicolumn{1}{l|}{82.55} & 83.83 & 88.59 \\ \hline
Ours (YOLOv3)                                                               & \textbf{\underline{22.55}} & \multicolumn{1}{l|}{41.55} & 30.38 & \multicolumn{1}{l|}{42.00} & 14.69 & 51.54 \\ \hline
\textbf{Ours (YOLOv7)}                                                      & 33.13 & \multicolumn{1}{l|}{\textbf{\underline{20.98}}} & \textbf{15.33} & \multicolumn{1}{l|}{\textbf{17.98}} & \textbf{10.11} & \textbf{32.43} \\ \hline
\end{tabular}
}
%\vspace*{-4.5mm}
\end{table}

\end{document}